\documentclass[9.5pt,journal,twoside,twocolumn]{IEEEtran}
\usepackage{graphicx}
\usepackage{psfrag}
\usepackage{cite}
\usepackage{amsmath,amssymb}
\usepackage{math}
\usepackage{framed}
\usepackage{subfigure}

\usepackage{color}
\definecolor{REVcolor}{rgb}{1.0,0.1,0.0}

\begin{document}

\title{Rigid and Articulated Point Registration with Expectation
  Conditional Maximization}
\author{Radu Horaud, Florence Forbes, Manuel Yguel, Guillaume
  Dewaele, and Jian Zhang
\thanks{R. Horaud, F. Forbes, M. Yguel, and G. Dewaele are with
INRIA Grenoble Rh\^one-Alpes,
655, avenue de l'Europe
38330 Montbonnot, France}
\thanks{J. Zhang is with the University of Hong Kong, Pokfulam, Hong Kong}
\thanks{Corresponding author: Radu.Horaud@inria.fr}
}

\newlength{\imgwfour}
\setlength{\imgwfour}{0.234\columnwidth}

\maketitle

\begin{abstract}
This paper addresses the issue of matching rigid and articulated
shapes through probabilistic point registration. The problem is
recast into a missing data framework where unknown correspondences
are handled via mixture models. Adopting a maximum likelihood
principle, we introduce an innovative EM-like algorithm,
    namely the \textit{Expectation
Conditional Maximization for Point Registration} (ECMPR)
algorithm. The algorithm  allows the use of general covariance
matrices for  the mixture model components and improves over
    the isotropic covariance case.  We analyse in detail the associated
    consequences in terms of estimation of the registration parameters,
    and we propose an optimal method for estimating the rotational and
    translational
    parameters based on \textit{semi-definite
      positive relaxation}.
We extend rigid registration to articulated
    registration.  Robustness is ensured by detecting and rejecting outliers through the addition of a uniform component to the Gaussian mixture model at hand.
    We provide an
    in-depth analysis of
    our method and we compare it both theoretically and experimentally
    with other robust methods for point registration.

\end{abstract}


\section{Introduction, related work, and contributions}
\label{section:introduction}

In image analysis and computer vision there is a long tradition of
algorithms for finding an optimal alignment between two sets of points.
This is referred to as the \textit{point registration} (PR) problem, which is
twofold: (i)~Find point-to-point correspondences and (ii)~estimate
the transformation allowing the alignment of the two sets. Existing
PR methods
can be roughly divided into three
categories: The Iterative Closest Point (ICP) algorithm
\cite{BeslMcKay92,Zhang1994} and
its numerous extensions
\cite{RusinkiewiczLevoy2001,Fitzgibbon2003,ChetverikovStepanovKrsek2005,SharpLeeWehe2008,Demirdjian2004,MundermanCorazzaAndriacchi2007},
soft assignment methods  
\cite{RCB:IPMI97,CR:CVIU03,Liu2005,Liu2007},
and probabilistic methods
\cite{Wells97,ChuiRangarajan2000,GrangerPennec2002,MyronenkoSongCarreira-Perpinan2006,SofkaYangStewart2007,JianVemuri2005}
to cite just a few.

ICP alternates between binary point-to-point assignments and
optimal estimation of the transformation parameters. Efficient
versions of ICP use sampling processes, either deterministic or
based on heuristics \cite{RusinkiewiczLevoy2001}. Sampling
strategies can be cast into more elaborate outlier rejection
methods such as \cite{Fitzgibbon2003} which applies a robust loss
function to the Euclidean distance, thus yielding a non-linear
version of ICP called LM-ICP. Another standard robust method is to
select trimmed subsets of points through repeated random sampling,
such as the TriICP algorithm proposed in
\cite{ChetverikovStepanovKrsek2005}. In \cite{SharpLeeWehe2008} a
maximum-likelihood non-linear optimizer is bootstrapped by
combining ICP with a RANSAC-like trimming method \cite{Meer2004}.
Although ICP is attractive for its efficiency, it
 can be easily trapped in
local minima due to the strict selection of the best
point-to-point assignments. This makes ICP to be particularly
sensitive both to initialization and to the choice of a threshold
needed to accept or to reject a match.

The nearest-point strategy of ICP can be replaced by soft
assignments within a continuous optimization framework
\cite{RCB:IPMI97,CR:CVIU03}. Let $m_{ji}$ be the positive entries
of the \textit{assignment} matrix $\mm{M}$, subject to the
constraints $\sum_{j}m_{ji}=1$, $\sum_{i}m_{ji}=1$. When there is
an equal number of points in the two sets, $\mm{M}$ is a
\textit{doubly stochastic matrix}. This introduces nonconvex
constraints: Indeed, the PR problem is solved using Lagrange
parameters and a barrier function within a constrained
optimization approach \cite{RCB:IPMI97}. The RPM algorithm
\cite{CR:CVIU03} extends \cite{RCB:IPMI97} to deal with outliers.
This is done by adding one column and one row to matrix $\mm{M}$,
say $\tmm{M}$. Several data points are allowed to be assigned to
this extra column and, symmetrically, several model points may be
assigned to this extra row. Therefore, the resulting algorithm
must provide optimal entries for $\tmm{M}$ and satisfy the
constraints on $\mm{M}$, thus providing one-to-one assignments for
inliers, and many-to-one assignments for outliers, i.e., several
entries are allowed to be equal to 1 in both the extra row and the
extra column. As a consequence, $\tmm{M}$ is not doubly stochastic
anymore and hence the convergence properties as described in
\cite{Sinkhorn64} are not guaranteed in the presence of outliers.

Probabilistic point registration uses, in general, Gaussian
mixture models (GMM). Indeed, one may reasonably assume that
points from the first set (the data) are normally distributed
around points belonging to the second set (the model). Therefore,
the point-to-point assignment problem can be recast into that of
estimating the parameters of a mixture. This can be done within
the framework of \textit{maximum likelihood with missing data}
because one has to estimate the mixture parameters as well as the
point-to-cluster assignments, i.e., the missing data.  In this
case the algorithm of choice is the expectation-maximization (EM)
algorithm \cite{DempsterLairdRubin77}. Formally, the latter
replaces the maximization of the \textit{observed-data
log-likelihood} with the maximization of the \textit{expected
complete-data log-likelihood
  conditioned by the observations}. As
it will be explained in detail in this paper, there are intrinsic
difficulties when one wants to cast the PR problem in the EM
framework. The main topic and contribution of this paper is to
propose an elegant and efficient way to do that.

In the recent past, several interesting \textit{EM-like}
implementations for point registration have been proposed
\cite{Wells97,ChuiRangarajan2000,GrangerPennec2002,MyronenkoSongCarreira-Perpinan2006}.
In \cite{Wells97} the posterior marginal pose estimation (PMPE)
method estimates the \textit{marginalized joint posterior} of
alignment and correspondence over all possible correspondences.
This formulation does not lead to the standard M-step of EM and,
in particular, it does not allow the estimation of the covariances
of the Gaussian mixture components. The \textit{complete-data
posterior energy function} is used in \cite{ChuiRangarajan2000}.
This leads to an E-step which updates a set of continuous
assignment variables which are similar but not identical to the
standard posterior probabilities of assigning points to clusters
\cite{FraleyRaftery2002}. It also leads to an M-step which
involves optimization of a non-linear energy function which is
approximated for simplification. The algorithms proposed in
\cite{Wells97} and \cite{ChuiRangarajan2000} do not lead to the true
maximum-likelihood (ML) solution.

In \cite{ChuiRangarajan2000} as well as in
\cite{GrangerPennec2002} and
\cite{MyronenkoSongCarreira-Perpinan2006} a simplified GMM is
used, namely a mixture with a spherical (isotropic) covariance
common to all the components. This has two important consequences.
First, it significantly simplifies the estimation of the alignment
parameters because the Mahalanobis distance is replaced by the
Euclidean distance: this allows the use of a closed-form solution
to find the optimal rotation matrix
\cite{ArunHuangBlostein87,Horn87-ortho,Horn87-quat,Umeyama91}, as
opposed to an iterative numerical solution as proposed in
\cite{WilliamsBennamoun2000}. Second, it allows connections
between GMM, EM, and deterministic annealing
\cite{YuilleStolorzUtans94}: The common variance is interpreted as
a temperature and its value is decreased at each step of the
algorithm according to an annealing schedule
\cite{RCB:IPMI97,CR:CVIU03,ChuiRangarajan2000,GrangerPennec2002,MyronenkoSongCarreira-Perpinan2006}.
Nevertheless, the spherical-covariance assumption inherent to
annealing has a number of drawbacks: anisotropic noise in the data
is not properly handled, it does not use the full Gaussian model,
and it does not fully benefit from the convergence properties of
EM because it anneals the variance rather than considering it as a
parameter to be estimated.

Another approach is to model each one of the two point sets by two
probability distributions and to measure the dissimilarity between
the two distributions
\cite{LuoHancock2003,TsinKanade2004,JianVemuri2005,Wang2006}. For
example, in \cite{JianVemuri2005}, each point set is modelled by a
GMM  where the number of components is chosen to be equal to the
number of points. In the case of rigid registration, this is
equivalent to replace the quadratic loss function with a Gaussian
and to minimize the sum of these Gaussians over all possible point
pairs. The Gaussian acts as a robust loss function. However there
are two major drawbacks: The formulation leads to a non-linear
optimization problem which must be solved under the nonconvex
rigidity constraints, which require proper initialization. Second,
the outliers are not explicitly modeled.

This paper has the following
original contributions:

\begin{itemize}
\item We formally cast the PR problem into the framework of maximum likelihood
  with missing data. We
  derive a maximization criterion based on the expected complete-data
  log-likelihood. We show that, within this context, the PR
  problem can be solved by an instance of the the \textit{expectation conditional
    maximization} (ECM) algorithm. It has been proven that ECM is more broadly
applicable than EM while it shares its desirable convergence
properties  \cite{MengRubin93}. In ECM, each M-step is replaced by a sequence of
conditional maximization steps, or \textit{CM-steps}. As it will be
explained and detailed in this paper, ECM is
particularly well suited for point registration because the
maximization over the registration parameters cannot be carried out
independently of the other parameters of the model, namely the
covariances. For these reasons we propose the \textit{Expectation
  Conditional Maximization for Point Registration} algorithm (ECMPR).
\item The vast majority of existing rigid point registration methods
  use \textit{isotropic}
  covariances for reasons that we just explained. In the more general case of
  \textit{anisotropic} covariances, we show that the optimization problem
  associated with rigid alignment cannot be solved in closed-form. The
  iterative numerical solution proposed in
  \cite{WilliamsBennamoun2000} estimates the motion parameters without
  estimating the covariances.
We
  propose and devise a novel
  solution to this problem which consists in transforming the
  nonconvex problem into a convex one using \textit{semi-definite
    positive (SDP) relaxation} \cite{LemarechalOustry2001}. Hence,
  rigid alignment in the presence of
  anisotropic covariance matrices is amenable to a tractable
  optimization problem.

\begin{figure}[t]
\centering
\begin{tabular}{cc}
\includegraphics[width=0.45\columnwidth]{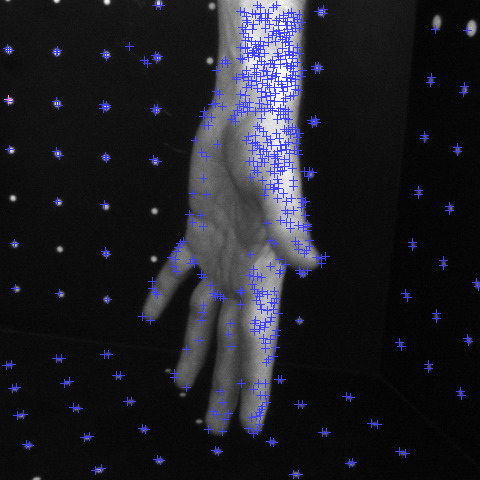}&
\includegraphics[width=0.45\columnwidth]{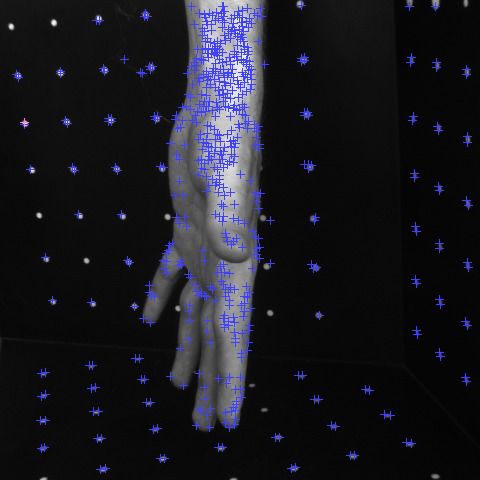}\\
\includegraphics[width=0.45\columnwidth]{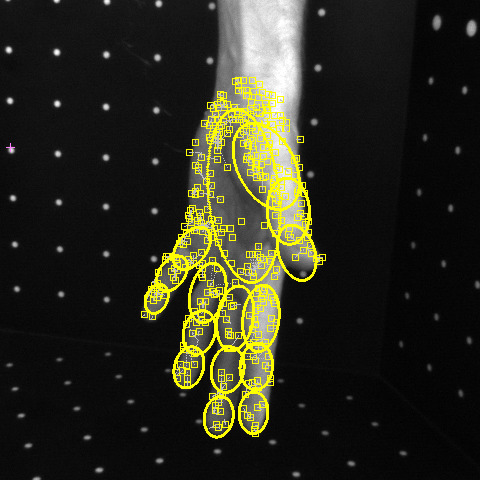}&
\frame{\includegraphics[width=0.45\columnwidth]{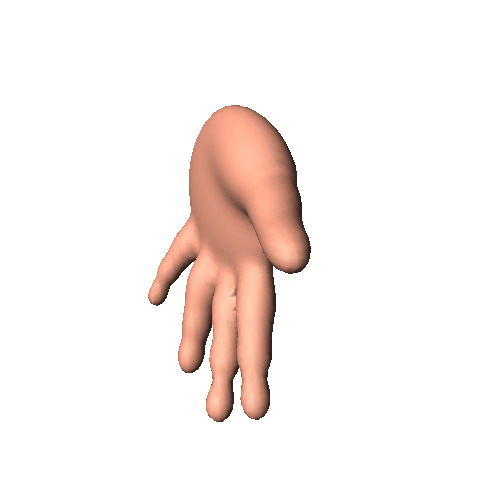}}
\end{tabular}
\caption{An illustration of the point registration method applied to
  the problem of aligning an articulated model of a hand to a set of
  3D points. The 3D data are obtained by stereo reconstruction from an
  image pair (top). The hand model consists of 3D points lying on
  16 hand parts (one root part, i.e., the palm, and 3 additional parts
  for each finger). The model contains 5 kinematic chains,
  each one is composed of the
  palm and one finger, i.e., 4 parts. Hence, the palm, or the root part, is common to
  all the kinematic chains. This articulated model has 27 degrees of freedom
  (3 translations and 3 rotations for the palm, 5 rotations for the
  thumb and 4 rotations for the index, middle, ring, and baby
  fingers). The result of the ECMPR-articulated
  algorithm is shown projected onto the left image (bottom-left) and
  as an implicit surface defined as a blending over the 16 hand parts
  (bottom-right). }
\label{fig:stereo-pair-and-result}
\end{figure}

\item We extend the rigid alignment solution just mentioned to
  articulated alignment. Based on the fact that the kinematic motion
  of an articulated object can be written as a chain of
  \textit{constrained} rigid motions, we devise an incremental
  solution which iteratively applies the rigid-alignment
  solution just mentioned to the rigid parts of the kinematic chain.
There are a few methods
for aligning articulated objects via point registration.
In
\cite{Demirdjian2004} ICP is first applied
independently to each rigid part of the articulated object and next,
the articulated constraints are enforced. The articulated ICP method
of \cite{MundermanCorazzaAndriacchi2007} alternates between
associating points from the two sets and estimating the articulated
pose. The latter is done by minimizing a non-linear least-square error
function which ensures that
the rigid body parts are in an optimal pose while the kinematic joint constraints
are only weakly satisfied. Our approach
has two advantages with respect to these methods. First, rather than ICP, we use ECM which has
proven convergence properties and which can handle inliers and
outliers in a principled way. Second, our incremental rigid alignment
formulation naturally enforces the kinematic constraints. As a
consequence, these constraints hold \textit{exactly} and there is no
need to enforce them a posteriori. Moreover, the articulated
registration method that we propose takes full advantage of the
rigid point registration algorithm, which is quite different from
data-point-to-object-part registration
\cite{KakadiarisMetaxas2000,PlankersFua2003,BMP04,KRH08,HNDB09}. We
note that our method is similar in spirit with
\cite{PellegriniSchindlerNardi2008}. However, the latter suffers from
the limitations of ICP.

\item One important property of any point registration method is its
  robustness to outliers. Our method has a built-in outlier model,
 namely a \textit{uniform component} that is added to the
  Gaussian mixture to account for non-Gaussian data, as suggested in
\cite{BanfieldRaftery93}. This adds an \textit{improper}
uniform-density component to the mixture. In theory
it is attractive to incorporate the estimation of the uniform
parameters into the EM algorithm. In
practice, this requires an in-depth analysis of ML for
the mixture
in the presence of several parametric component models,
which is an unsolved problem
\cite{Hennig2004,HennigCoretto2007}. We propose a treatment of
the uniform component on the basis of
considerations and properties that are specific to point
registration. This modifies the expressions of the posteriors without
adding any extra free parameters in the maximization step and
without
altering the general structure of the algorithm. Hence,
the convergence proofs of ECM
\cite{MengRubin93} and of EM
\cite{DempsterLairdRubin77,RednerWalker84,McLachlanKrishnan97} carry
over in this case.

Our approach to outlier rejection differs  from existing methods
currently used in point registration. Non-quadratic robust loss
functions are proposed in \cite{Fitzgibbon2003} and in
\cite{SofkaYangStewart2007} but the drawback is that the
optimization process can be trapped in local minima. This is not
the case with our method because of the embedding of outlier
rejection within EM. Other robust techniques such as RANSAC
\cite{Meer2004,SharpLeeWehe2008}, least median of squares (LMS)
\cite{Rousseeuw84}, or least trimmed squares (LTS)
\cite{RousseeuwVanAelst99,ChetverikovStepanovKrsek2005} must
consider a very large number of subsets sampled from the two sets
of points before a satisfactory solution can be found. Moreover,
there is a risk that the two trimmed subsets which are eventually
selected (a data subset and a model subset) contain outlying data
that lead to a good fit. These random sampling issues are even
more critical when one deals with articulated objects because
several subsets of trimmed data points must be available, i.e.,
one trimmed subset for each rigid part.

\item We perform extensive experiments with both the ECMPR-rigid and
  ECMPR-ariculated point registration algorithms. We thoroughly study the
  behaviour of the method with respect to (i)~the initial parameter
  values, (ii)~the amount of noise added to the observed data,
  (iii)~the presence of outliers, and (iv)~the use of anisotropic
  covariances instead of isotropic ones. We illustrate the
  effectiveness of the method in the case of tracking a complex
  articulated object -- a human hand composed of 5 kinematic chains,
  16 parts, and 27
  degrees of freedom, as shown in Fig.~\ref{fig:stereo-pair-and-result}.

\end{itemize}

The remainder of this paper is organized as follows. In section
\ref{section:formulation} the PR problem is cast into the framework of ML. In section
\ref{section:3D-point-registration} 
the expected complete-data
log-likelihood is derived. In section \ref{section:EM} the EM
algorithm for point registration, ECMPR, is formally derived.
The algorithm is applied to rigid point sets
(section~\ref{section:rigid}) and to articulated point sets
(section~\ref{section:articulated-pose-tracking}). Experimental
results obtained both with simulated and  real data are described
in section~\ref{section:Experimental results}.

\section{Problem formulation}
\label{section:formulation}

\subsection{Mathematical notations}
\label{subsection:definitions}

Throughout the paper, vectors will be in slanted bold style while matrices will
be in bold style. We will consider two sets of 3-D points.
We denote by $\mathcal{Y}=\{\vv{Y}_j\}_{1 \leq j
  \leq m}$ the 3-D coordinates of a set of \textit{observed} data points and by
$\mathcal{X}=\{\vv{X}_i\}_{1 \leq i \leq n}$ the 3-D coordinates of
a set of model
points.
The model points lie on the surface of either
a rigid or an articulated object. Hence, each model point may undergo either
a rigid or an articulated transformation which will be denoted by
$\vv{\mu}:\mathbb{R}^3\rightarrow\mathbb{R}^3$. The 3-D coordinates of
a transformed model point $\vv{\mu}(\vv{X}_i;\mathbf{\Theta})$ are
parameterized by $\mathbf{\Theta}$. In the case of rigid registration, the
parameterization will consist of a 3$\times$3 rotation matrix
$\mm{R}$ and a 3$\times$1
translation vector $\vv{t}$. Hence, in this case we have:
\begin{equation}
\label{eq:rigid-motion}
\vv{\mu}(\vv{X}_i;\mathbf{\Theta}) = \mm{R} \vv{X}_i + \vv{t},\;\;
\mathbf{\Theta}:=\{\mm{R},  \vv{t}\}.
\end{equation}
We will refer to the parameter vector $\mathbf{\Theta}$ as the
\textit{registration parameters}.
Section~\ref{section:articulated-pose-tracking} will make explicit the registration parameters in the case of articulated objects.

A parameter overscripted by $\ast$, e.g., $\mathbf{\Theta}^{\ast}$,
denotes the optimal value of that parameter. 
The overscript $\top$
denotes the transpose of a
vector or of a matrix. $\|\vv{X}-\vv{Y}\|^2$ is the squared
Euclidean
distance and $\|\vv{X}-\vv{Y}\|_\mathbf{\Sigma}^2$
is the squared Mahalanobis distance,
i.e. $(\vv{X}-\vv{Y})\tp\mm{\Sigma}\inverse(\vv{X}-\vv{Y})$ where
$\mm{\Sigma}$ is a 3$\times$3 symmetric positive definite matrix.

\subsection{Point registration,  maximum likelihood, and EM}
\label{subsection:PRP-MLE}

In this paper we will formulate point registration as the estimation
of a mixture of densities:
A Gaussian mixture model (GMM) is fitted to the
data set $\mathcal{Y}$ such that the centers of the Gaussian densities are constrained to
coincide with the transformed model points
$\vv{\mu}(\vv{X}_i;\mathbf{\Theta}),\; \vv{X}_i\in\mathcal{X}$. Therefore,
each density in the mixture is characterized by a mean vector $\vv{\mu}_i$
and a
covariance matrix $\mathbf{\Sigma}_i$.

In the standard mixture model approach both the means and the
covariances are the free parameters. Here the means are parameterized by the
registration parameters which enforce prior knowledge about the
transformation that exists between the two sets of points. Therefore,
the \textit{observed-data log-likelihood} is a function of both the
registration parameters and of the covariance matrices:
\begin{equation}
\label{eq:obs-data-logl}
\mathcal{L}(\mathbf{\Theta},\mathbf{\Sigma}_1,\ldots,\mathbf{\Sigma}_n|\mathcal{Y})
= \log P
(\mathcal{Y};\mathbf{\Theta},\mathbf{\Sigma}_1,\ldots,\mathbf{\Sigma}_n)
\end{equation}
The direct maximization of $\mathcal{L}$ over these parameters
is intractable
due to the presence of missing data, namely the unknown assignment of each
observed data point
$\vv{Y}_j$ to one of the mixture's components.
Let $\mathcal{Z}=\{Z_j\},1\leq j \leq m$
be these missing data which will be treated as a set of \textit{hidden
  random variables}. Each variable $Z_j$ assigns
an observed data point  $\vv{Y}_j$ to a model point $\vv{X}_i,
1\leq j\leq n$, or to an \textit{outlier class} indexed by $n+1$.

Dempster, Laird \& Rubin
\cite{DempsterLairdRubin77} proposed to replace $\mathcal{L}$ with the  \textit{expected complete-data log-likelihood conditioned by
the observed data}, where the term complete-data refers to both the
observed data $\mathcal{Y}$ and the missing data $\mathcal{Z}$, and where the
expectation is taken over the missing data (or the hidden variables):
\begin{equation}
\label{eq:complete-data-logl}
\mathcal{E}(\mathbf{\Theta},\mathbf{\Sigma}_1,\ldots,\mathbf{\Sigma}_n|\mathcal{Y},\mathcal{Z}) = E_{\mathcal{Z}}[\log
P(\mathcal{Y},\mathcal{Z};\mathbf{\Theta},\mathbf{\Sigma}_1,\ldots,\mathbf{\Sigma}_n)|\mathcal{Y}]
\end{equation}


Expectation maximization (EM)
\cite{DempsterLairdRubin77}, is an iterative method for finding maximum
likelihood estimates in
incomplete-data problems like the one just stated.
 It has been proven that the EM algorithm converges to a local maximum of
 the expected complete-data log-likelihood ($\mathcal{E}$)
 and that the maximization of $\mathcal{E}$
 also maximizes the observed-data
log-likelihood $\mathcal{L}$
\cite{RednerWalker84,McLachlanKrishnan97}.

\subsection{The proposed method}
\label{subsection:ourmethod}
As it will be explained in detail below, in the case of point
registration, EM must be replaced by ECM. This will yield the following method:
\begin{enumerate}
\item Provide initial values for the model parameters;
\item \textit{E-step}. Compute the posterior probabilities 
  given the current estimates of the registration parameters
  ($\mathbf{\Theta}^{q}$) and
  of the covariance matrices $\mathbf{\Sigma}^{q}=(\mathbf{\Sigma}_1^{q},\ldots,\mathbf{\Sigma}_n^{q})$:
\[ \alpha_{ji}^{q} =
P(Z_j=i|\vv{X}_j;\mathbf{\Theta}^{q},\mathbf{\Sigma}^{q})
\]
\item \textit{CM-steps}. Maximize the expectation in (\ref{eq:complete-data-logl}) with respect to:
\begin{enumerate}
\item The registration parameters, \textit{conditioned} by the current
  covariance matrices:
\[
\mathbf{\Theta}^{q+1} = \arg \max_{\mathbf{\Theta}}
E_{\mathcal{Z}}[\log
P(\mathcal{Y},\mathcal{Z};\mathbf{\Theta},\mathbf{\Sigma}^{q})|\mathcal{Y}]
\]
\item The covariance matrices \textit{conditioned} by the newly
  estimated registration parameters:
\[ \mathbf{\Sigma}^{q+1} = \arg \max_{\mathbf{\Sigma}}
E_{\mathcal{Z}}[\log
P(\mathcal{Y},\mathcal{Z};\mathbf{\Theta}^{q+1} ,\mathbf{\Sigma}
)|\mathcal{Y}]
\]
\end{enumerate}
\item Check for convergence.
\end{enumerate}

\section{Point registration and Gaussian mixtures}
\label{section:3D-point-registration}

In order to estimate the registration parameters, one needs to find
correspondences between the observed data points and the model points. These
correspondences are the \textit{missing data} and will be treated as {\em hidden} variables within the
framework of
\textit{maximum likelihood}. Hence, there is a strong analogy with
clustering. An observed data point $\vv{Y}_j$ could be assigned either
to a Gaussian cluster centered at
$\vv{\mu}(\vv{X}_i;\mathbf{\Theta})$, or to a \textit{uniform class}
defined in detail below. In
section~\ref{subsection:PRP-MLE} we already briefly introduced the hidden variables
$\mathcal{Z}=\{Z_j\}_{1 \leq j \leq m}$ which describe the
assignments of the observations to clusters, or equivalently, the
data-point-to-model-point correspondences.
More specifically, the notation $Z_j=i$ (or $Z:j\rightarrow
i$) means that the observation $\vv{Y}_j$
matches the model point $\vv{X}_i$ while $Z_j=n+1$ means
that the observation $\vv{Y}_j$ is an outlier.

We also denote by $p_i=P(Z_j=i)$ the prior probability that
observation $\vv{Y}_j$  belongs to cluster $i$ with center
$\vv{\mu}(\vv{X}_i;\mathbf{\Theta})$ and by $p_{n+1}= P(Z_j=n+1)$
the prior probability of observation $j$ to be an outlier. We also
denote by $P(\vv{Y}_j|Z_j=i), \forall j\in\{1,\ldots,m\}, \forall
i\in\{1,\ldots,n+1\}$ the conditional likelihood of $\vv{Y}_j$,
namely the probability of $\vv{Y}_j$ given its cluster assignment.

The likelihood of an observation $j$ given its assignment to cluster $i$
is drawn from a Gaussian distribution with mean
$\vv{\mu}(\vv{X}_i;\mathbf{\Theta})$ and covariance $\mathbf{\Sigma}_i$:
\begin{equation}
\label{equ:dist1}
P(\vv{Y}_j|Z_j=i) = \mathcal{N} (\vv{Y}_j| \vv{\mu}(\vv{X}_i;\mathbf{\Theta}),
\mathbf{\Sigma}_i),\; \forall i,\; 1\leq i\leq n
\end{equation}

Similarly, the likelihood of an observation given its assignment to the outlier
cluster is a uniform distribution over the volume $V$ of the 3-D
working space:
\begin{equation}
\label{equ:dist2}
P(\vv{Y}_j|Z_j=n+1) = \mathcal{U} (\vv{Y}_j|V,0) = \frac{1}{V}
\end{equation}

Since $\{Z_j=1,\hdots,Z_j=n,Z_j=n+1\}$ is a partition of the event
space of $Z_j$, the marginal distribution of an observation is:
\begin{equation}
\label{eq:likelihood-Yi-1}
P
(\vv{Y}_j) =  \sum_{i=1}^{n+1} p_i P(\vv{Y}_j|Z_j=i)
\end{equation}
By assuming that the observations are independent and identically
distributed, the
\textit{observed-data}
\textit{log-likelihood} $\mathcal{L}$, i.e., eq.~(\ref{eq:obs-data-logl}) writes:
\begin{equation}
\label{eq:joint-log-likelihood}
\log P
(\mathcal{Y}) =
\sum_{j=1}^{m} \log \left(
\sum_{i=1}^{n} p_i  \mathcal{N} (\vv{Y}_j| \vv{\mu}(\vv{X}_i;\mathbf{\Theta}),\mathbf{\Sigma}_i)
+ \frac{p_{n+1}}{V}
\right)
\end{equation}
The observed-data log-likelihood is conditioned by the registration
parameters $\mathbf{\Theta}$ (which constrain the centers of the
Gaussian clusters), by $n$ covariance matrices $\mathbf{\Sigma}_i$, by
$n+1$ cluster priors $p_i$ subject to the constraint
$\sum_{i=1}^{n+1}p_i=1$, and by the uniform-distribution parameter  $V$.
In the next section we will discuss the choice of the priors and the
parameterization of the uniform distribution in the specific context
of point registration.

It will be convenient to denote the parameter set by:
\begin{equation}
\label{eq:parameter-vector}
\vv{\Psi} = \{
\mathbf{\Theta},\mathbf{\Sigma}_1,\ldots,\mathbf{\Sigma}_n\}
\end{equation}
A powerful method for finding ML solutions in the presence of hidden
variables is to replace the observed-data log-likelihood
with the complete-data log-likelihood and to maximize the
\textit{expected} complete-data
  log-likelihood \textit{conditioned by the observed data}.
The criterion to be maximized (i.e.,
eq.~(\ref{eq:complete-data-logl})) becomes \cite{Bishop2006}:
\begin{equation}
\label{eq:MLE-expectation}
\mathcal{E}(\vv{\Psi}|\mathcal{Y},\mathcal{Z})= \sum_{\mathcal{Z}}
P(\mathcal{Z}|\mathcal{Y},\vv{\Psi} )\;
\log P (\mathcal{Y},\mathcal{Z};\vv{\Psi})
\end{equation}
\section{EM for point registration}
\label{section:EM}
In this section
we formally derive the EM algorithm for robust point registration.
We start by making explicit the posterior probabilities of the assignments
conditioned by the observations when both the observed data and the
model data are described by 3-D points; Using Bayes' rule we have:
\begin{equation}
\label{eq:alpha-ij}
\alpha_{ji} = P(Z_j=i | \vv{Y}_j) = \frac{P(\vv{Y}_j| Z_j=i)P(Z_j=i)}{P(\vv{Y}_j)}
\end{equation}
In general, EM treats the priors $p_i=P(Z_j=i)$ as parameters. In the
case of point registration we propose to specialize the priors as
follows:
\begin{equation}
\label{equ:pi1}
p_i =
\left\{
\begin{array}{clc}
p_{in} &=  \frac{v}{V} & \text{if }  1\leq i \leq n \\
p_{out}& = \frac{V-nv}{V} & \text{if }  i=n+1
\end{array}
\right.
\end{equation}
where $v=4\pi  r^3/3$ is the volume of a small sphere with radius
$r$ centered at a model point $\vv{X}_i$. We assume $nv \ll V$. By
combining eqs.~(\ref{equ:dist1}), (\ref{equ:dist2}),
(\ref{eq:likelihood-Yi-1}), and (\ref{equ:pi1}) we obtain, for all
$i=1 \ldots n$:
\begin{equation}
\label{eq:alpha-Estep}
\alpha_{ji} = \frac
{ |\mathbf{\Sigma}_i|^{-\frac{1}{2}}\exp
\left( - \frac{1}{2}
  \|\vv{Y}_j-\vv{\mu}(\vv{X}_i;\mathbf{\Theta})\|^2_{\mathbf{\Sigma}_i}\right) }
{\sum_{k=1}^{n} |\mathbf{\Sigma}_k|^{-\frac{1}{2}} \exp \left( - \frac{1}{2}
    \|\vv{Y}_j-\vv{\mu}(\vv{X}_k,\mathbf{\Theta})\|^2_{\mathbf{\Sigma}_k} \right) +   \emptyset_{3D}
  }
\end{equation}
where $\emptyset_{3D}$ corresponds to the \textit{outlier
component} in the case of 3-D point registration:
\begin{equation}
\label{eq:outlier-component}
\emptyset_{3D} =  1.5\sqrt{2\pi}r^{-3}
\end{equation}
Note that there is a similar expression in the case of 2-D point
registration, namely $\emptyset_{2D} = 2r^{-2}$ and that this can
be generalized to any dimension. The posterior probability of an
\textit{outlier} is given by:
\begin{equation}
\label{eq:alpha-ij-outlier}
\alpha_{j\;n+1} = 1 - \sum_{i=1}^{n} \alpha_{ji}
\end{equation}
Next we derive an explicit formula for $\mathcal{E}$ in
eqs.~(\ref{eq:complete-data-logl}) and
(\ref{eq:MLE-expectation}). For that purpose we expand the
complete-data log-likelihood:
\begin{eqnarray*}
\log P (\mathcal{Y}, \mathcal{Z};\mathbf{\Psi}) &=&
\log \prod_{j=1}^{m} P(\vv{Y}_j,Z_j;\mathbf{\Psi}) \\
&=& \log \prod_{j=1}^{m}
P(\vv{Y}_j|Z_j;\mathbf{\Psi})P(Z_j) \\
&=& \log \prod_{j=1}^{m} \prod_{i=1}^{n+1} \left\{p_i
  P(\vv{Y}_j|Z_j=i;\mathbf{\Psi})
\right\}^{\delta_{i Z_{j}}}
\end{eqnarray*}
where $\delta_{i z_{j}}$ is the Kronecker symbol defined by:
\begin{equation}
\label{eq:indicator-variable} \delta_{i Z_{j}} = \left\{
\begin{array}{cl}
1 & \text{if} \; Z_j=i \\
0 & \text{otherwise}
\end{array}
\right.
\end{equation}
Therefore, eq.~(\ref{eq:complete-data-logl}), i.e.,
$\mathcal{E}(\vv{\Psi}|\mathcal{Y},\mathcal{Z})$ can be written as:
\begin{eqnarray}
E_{\mathcal{Z}} \left[ \sum_{j=1}^{m} \sum_{i=1}^{n+1} \delta_{i
Z_{j}} \big( \log p_i + \log P(\vv{Y}_j|Z_j=i,\mathbf{\Psi})
\big) \mid \mathcal{Y}\right] \nonumber \\
\label{eq:E-loglikelihood} = \sum_{j=1}^{m} \sum_{i=1}^{n+1}
E_{\mathcal{Z}} [\delta_{i Z_{j}}|\mathcal{Y}] \big( \log p_i +
\log P(\vv{Y}_j|Z_j=i,\mathbf{\Psi}) \big)
\end{eqnarray}
where the conditional expectation of
$\delta_{i Z_{j}}$ writes
\begin{equation}
E_{\mathcal{Z}} [\delta_{i Z_{j}}|\mathcal{Y}] = \sum_{k=1}^{n+1}
\delta_{i k} P(Z_j=k|\vv{Y}_j) = \alpha_{ji}
\end{equation}
By replacing the conditional probabilities with the normal and
uniform distributions, i.e., (\ref{equ:dist1}) and
(\ref{equ:dist2}), and by neglecting constant terms, i.e., terms
that do not depend on $\mathbf{\Psi}$,
eq.~(\ref{eq:E-loglikelihood}) can be written as:
\begin{equation}
\label{eq:EM-min}
\mathcal{E}(\mathbf{\Psi})
= -\frac{1}{2}  \sum_{j=1}^{m}
\sum_{i=1}^{n}  \alpha_{ji} \left(
\|\vv{Y}_j-\vv{\mu}(\vv{X}_i;\mathbf{\Theta})\|^2_{\mathbf{\Sigma}_i}
+ \log |\mathbf{\Sigma}_i|
\right)
\end{equation}
It was proven that the maximizer of (\ref{eq:EM-min}) also
maximizes the observed-data log-likelihood
(\ref{eq:joint-log-likelihood}) and that this maximization may be
carried out by the EM algorithm
\cite{RednerWalker84,McLachlanKrishnan97}. Nevertheless, there is
an additional difficulty in the case of point registration. In the
standard EM, the free parameters are the means and the covariances
of the Gaussian mixture and the estimation of these parameters is
quite straightforward. In the case of point registration, the
means are constrained by the registration parameters and,
moreover, the functions $\vv{\mu}_i(\vv{X}_i;\vv{\Theta})$ are
complicated by the presence of the rotation matrices, as detailed
in section~\ref{section:rigid}. In practice, the estimation of
$\mathbf{\Theta}$ is conditioned by the covariances. The
simultaneous estimation of all the model parameters within the
M-step would lead to a difficult non-linear minimization problem.
Instead, we propose to minimize (\ref{eq:EM-min}) over
$\vv{\Theta}$ while keeping the covariance matrices constant,
which leads to (\ref{eq:minimum-theta}) below, and next we
estimate the empirical covariances $\mm{\Sigma}_i$ using the newly
estimated registration parameters. This amounts to replace EM by
ECM \cite{MengRubin93}. In practice we obtain two conditional
minimization steps, using $\alpha_{ij}^{(q)}$ given by
(\ref{eq:alpha-Estep}) and (\ref{eq:alpha-ij-outlier}):
\begin{equation}
\label{eq:minimum-theta}
\mathbf{\Theta}^{q+1} = \arg \min_{\mathbf{\Theta}}
\frac{1}{2}  \sum_{j=1}^{m}
\sum_{i=1}^{n}  \alpha_{ji}^q
  \|\vv{Y}_j-\vv{\mu}(\vv{X}_i;\mathbf{\Theta})\|^2_{\mathbf{\Sigma}_i^q}
\end{equation}
and for all $i=1 \ldots n$,
\begin{equation}
\label{eq:empirical-covariance}
\mm{\Sigma}_i^{q+1} = \frac{
\sum_{j=1}^m \alpha_{ji}^q
(\vv{Y}_j-\vv{\mu}(\vv{X}_i;\mathbf{\Theta}^{q+1}))
(\vv{Y}_j-\vv{\mu}(\vv{X}_i;\mathbf{\Theta}^{q+1}))\tp}{\sum_{j=1}^m \alpha_{ji}^q}
\end{equation}
It is well known  (e.g. \cite{IngrassiaRocci07,Bishop2006})  that when the mean $\vv{\mu}_i$ of one of the Gaussian components collapses
onto a specific data point while the other data points are
``infinitely'' away from $\vv{\mu}_i$, the entries of the corresponding covariance matrix
$\mm{\Sigma}_i$ tend to zero. Since \cite{Hathaway85}, the phenomenon has been well studied for Gaussian mixtures. Under suitable conditions, constrained global maximum likelihood
formulations have been proposed, which present no singularities and a smaller number
of spurious maxima (see \cite{IngrassiaRocci07} and the references
therein). However, in practice these studies do not always lead to efficient EM implementations.
Thus, in order to avoid such degeneracies, the
covariance is artificically fattened as follows. Let
$\mm{Q}\mm{D}\mm{Q}\tp$ be the eigendecomposition of $\mm{\Sigma}_i$
and let's replace the diagonal matrix $\mm{D}$ with $\mm{D}+\varepsilon
\mm{I}$. We obtain $\mm{\Sigma}_i^{\varepsilon}=\mm{Q}(\mm{D}+\varepsilon
\mm{I})\mm{Q}\tp=\mm{\Sigma}_i+\varepsilon \mm{I}$. Hence, adding
$\varepsilon\mm{I}$, where $\varepsilon$ is a small positive number
slightly
fattens the covariance matrix without 
affecting its characteristics (eccentricity and orientation of the
associated ellipsoid).  A more theoretical analysis and other similar transformations of problematic covariance matrices
are proposed in \cite{IngrassiaRocci07} but the straightforward choice above provided satisfying 
results.

\begin{figure*}[htb]
\centering
\subfigure[2-nd iteration]{
\frame{\includegraphics[width=0.600\columnwidth,viewport=60 30 300 280,clip]{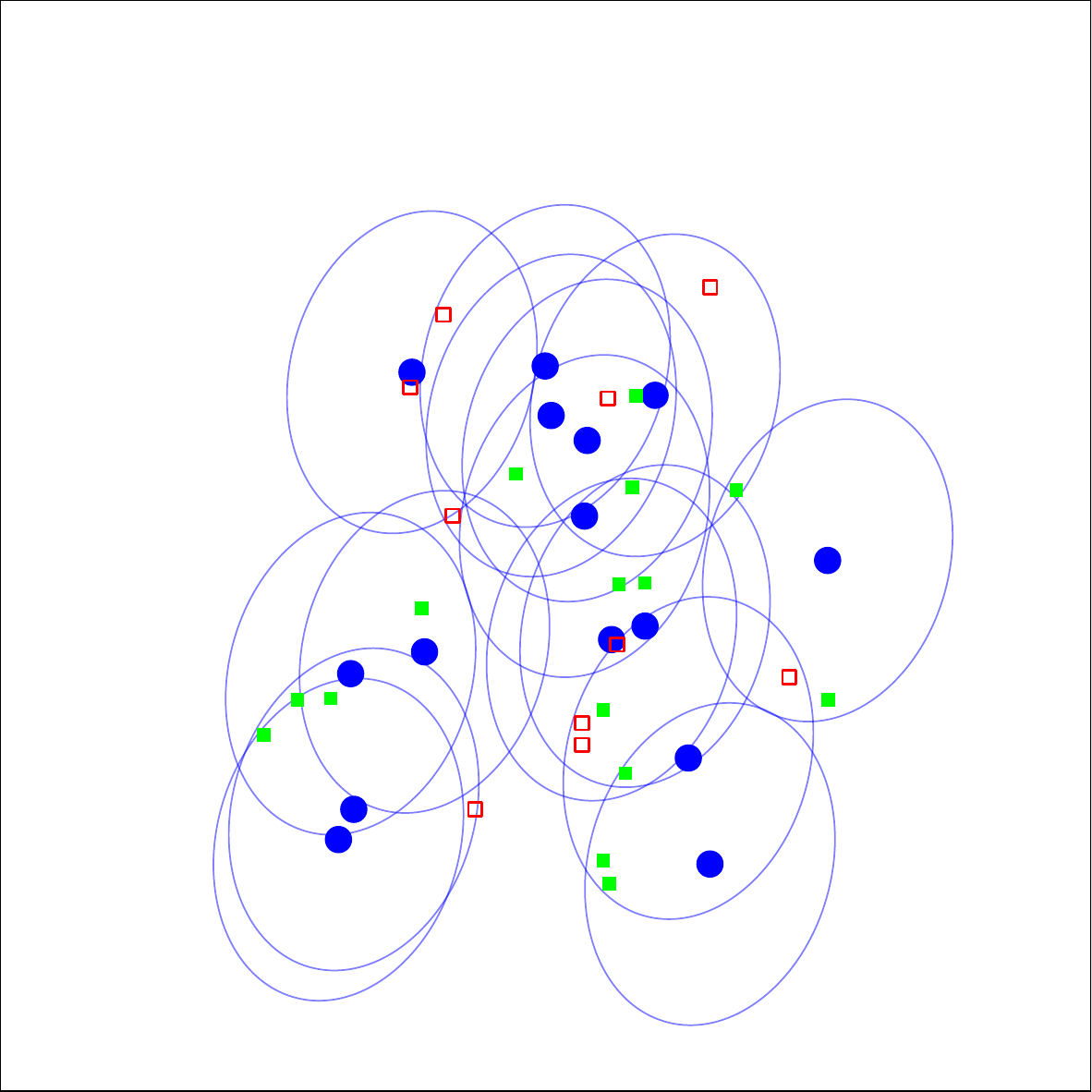}}}
\hfil
\subfigure[6-th iteration]{
\frame{\includegraphics[width=0.600\columnwidth,viewport=60 30 300 280,clip]{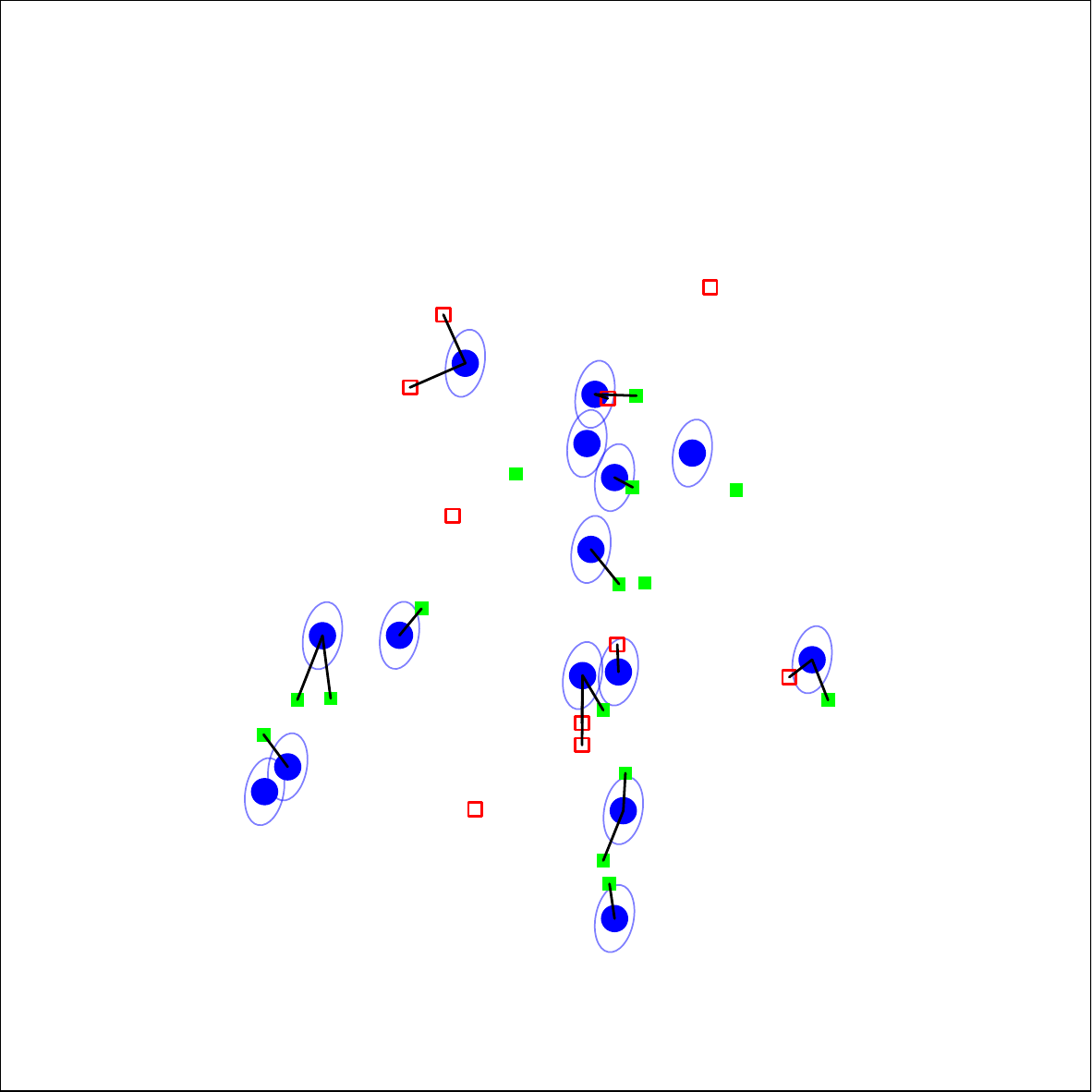}}}
\hfil
\subfigure[35-th iteration]{
\frame{\includegraphics[width=0.600\columnwidth,viewport=60 30 300 280,clip]{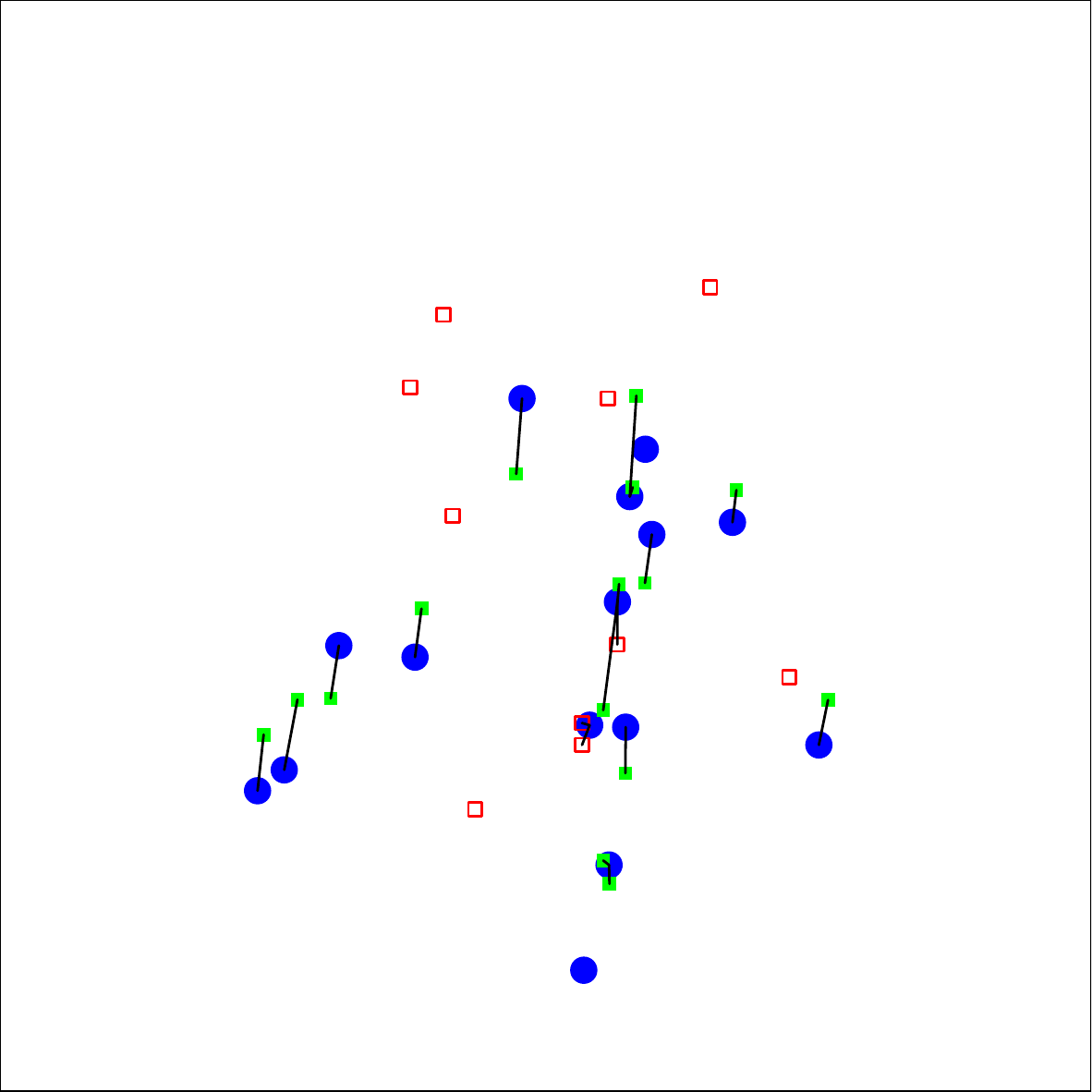}}}
\caption{Illustration of the ECMPR-rigid algorithm. There are 15 model
  points (filled blue circles) and 25 data points, 15 inliers
  (filled green squares)
  that correspond to model points that were rotated, translated and
  corrupted by anisotropic Gaussian noise, and 10 outliers
  (empty red squares)
  drawn from a uniform distribution. In this example we modeled all
  the components of the mixture model with a common covariance
  matrix (shown with ellipses), as in (\ref{eq:empirical-covariance-one}). The lines correspond to current data-to-model assignments.
The algorithm converged at the 35-th iteration. There are 12
  data-point-to-model-point assignments and 7 data-point-to-outlier-class assignments
  corresponding to the ground truth. This example corresponds to the
  second row in
  Table~\ref{table:rigid-registration}. See
  Section~\ref{section:Experimental results} for more details.}
\label{fig:ecmpr-anisotropicnoise}
\end{figure*}

Alternatively,
one may 
model all the components of the
mixture with a common covariance matrix:
\begin{equation}
\label{eq:empirical-covariance-one}
\mm{\Sigma}^{q+1} = \frac{
\sum\limits_{j=1}^m \sum\limits_{i=1}^n \alpha_{ji}^q
(\vv{Y}_j-\vv{\mu}(\vv{X}_i;\mathbf{\Theta}^{q+1}))
(\vv{Y}_j-\vv{\mu}(\vv{X}_i;\mathbf{\Theta}^{q+1}))\tp}{\sum_{j=1}^m
\sum_{i=1}^n \alpha_{ji}^q}
\end{equation}
When the number of data points is small, it is preferable to use
(\ref{eq:empirical-covariance-one}), e.g.,
Fig.~\ref{fig:ecmpr-anisotropicnoise}.

Notice that (\ref{eq:minimum-theta}) can be further simplified by introducing the \textit{virtual
  observation} $\vv{W}_i$ and its weight $\lambda_i$ that are assigned
to a model point $\vv{X}_i$:
\begin{eqnarray}
\label{eq:mean-observation}
\vv{W}_i& =& \frac{1}{\lambda_i} \sum_{j=1}^{m} \alpha_{ji}
\vv{Y}_j\\
\label{eq:lambda}
\lambda_i &=& \sum_{j=1}^{m} \alpha_{ji}
\end{eqnarray}
By expanding (\ref{eq:minimum-theta}), substituting the
corresponding terms with
(\ref{eq:mean-observation}) and (\ref{eq:lambda}), and by neglecting
constant terms the minimizer yields a simpler
expression:
\begin{equation}
\label{eq:minimize-E}
\mathbf{\Theta}^{q+1} = \arg \min_{\mathbf{\Theta}} \frac{1}{2} \sum_{i=1}^{n}  \lambda_{i}^{q}
  \| \vv{W}_{i}^q-\vv{\mu}(\vv{X}_i;\mathbf{\Theta})\|^2_{\mathbf{\Sigma}_i^{q}}
\end{equation}
It is worth noticing that the many-to-one assignment
model developed here
has a one-to-one (data-point-to-model-point) structure: The virtual
observation $\vv{W}_i$ (corresponding to a normalized sum over
\textit{all} the observations
weigthed by their
posteriors $\{\vv{Y}_j,\alpha_{ji}\}, 1\leq j \leq m$) is
assigned to the model
point $\vv{X}_i$.
Eq.~(\ref{eq:minimize-E}) will facilitate the development of an
optimization method for the rigid and articulated point registration
problems as outlined in the next sections.
Moreover, the minimization of (\ref{eq:minimize-E})
 is computationally more efficient than the
minimization of eq.~(\ref{eq:minimum-theta}) because it involves fewer terms.

\section{Rigid point registration}
\label{section:rigid}

In this section we assume that the model points lie on a rigid
object. Therefore:
\begin{equation}
\label{eq:theta-rigid} \mathbf{\Theta}:= (\mm{R},\vv{t})\tp\;.
\end{equation}
Eq.~(\ref{eq:rigid-motion}) holds in this case and
(\ref{eq:minimize-E}) becomes:
\begin{equation}
\label{eq:rigid-case}
\mathbf{\Theta}^{\ast} =
\arg \min_{\mm{R},\vv{t}} \frac{1}{2} \sum_{i=1}^{n}  \lambda_{i}
\|\vv{W}_i - \mm{R} \vv{X}_i - \vv{t}\|^2_{ \mathbf{\Sigma}_i}
\end{equation}
Minimization with respect to the translation parameters is easily obtained by taking the
derivatives of (\ref{eq:rigid-case}) with respect to the 3-D vector
$\vv{t}$ and setting these derivatives to zero. We
obtain:
\begin{equation}
\label{eq:optimal-t}
\vv{t}^{\ast} =\left(\sum_{i=1}^{n}\lambda_i\mathbf{\Sigma}_i\inverse\right)\inverse
\sum_{i=1}^{n}\lambda_i\mathbf{\Sigma}_i\inverse(\vv{W}_i-\mm{R}\vv{X}_i)
\end{equation}
By substituting this expression
in (\ref{eq:rigid-case}), we obtain:
\begin{eqnarray}
\label{eq:minimize-rigid-1}
\mm{R}^{\ast} &=& \arg \min_{\mm{R}} \frac{1}{2}
\sum_{i=1}^{n}  \lambda_{i} \left(
\vv{X}_i\tp\mm{R}\tp\mathbf{\Sigma}_i\inverse \mm{R} \vv{X}_i +
2\vv{X}_i\tp\mm{R}\tp\mathbf{\Sigma}_i\inverse\vv{t}^{\ast}
\right. \nonumber \\
&&\left.
- 2\vv{X}_i\tp\mm{R}\tp\mathbf{\Sigma}_i\inverse\vv{W}_i
-2{\vv{t}^{\ast}}\tp\mathbf{\Sigma}_i\inverse\vv{W}_i +
{\vv{t}^{\ast}}\tp\mathbf{\Sigma}_i\inverse\vv{t}^{\ast}
\right)
\end{eqnarray}
The minimization of (\ref{eq:minimize-rigid-1}) must be carried out in
the presence of the orthonormality constraints associated with the
rotation matrix, i.e., $\mm{R}\mm{R}\tp=\mm{I}$ and $|\mm{R}|=+1$.

\subsection{Isotropic covariance model}
Eq. (\ref{eq:minimize-rigid-1}) significantly simplifies when
isotropic covariance
matrices are being used, namely $\mathbf{\Sigma}_i=\sigma^2_i\mm{I}_3$. In this case,
the criterion above has a
much simpler form because the Mahalanobis distance reduces to the Euclidean distance. We obtain:
\begin{equation}
\label{eq:optimal-t-isotropic}
\vv{t}^{\ast} = \frac{\sum_{i=1}^{n}\lambda_i\sigma_i^{-2}(\vv{W}_i-\mm{R}\vv{X}_i)}{\left(\sum_{i=1}^{n}\lambda_i \sigma_i^{-2}\right)}
\end{equation}
and:
\begin{eqnarray}
\mm{R}^{\ast} &=& \arg \min_{\mm{R}} \frac{1}{2}
\sum_{i=1}^{n}  \lambda_{i}\sigma_i^{-2} \left(
2\vv{X}_i\tp\mm{R}\tp\vv{t}^{\ast} -
2\vv{X}_i\tp\mm{R}\tp\vv{W}_i
\right. \nonumber \\
\label{eq:minimize-rigid-isotropic}
&& \left.
-2{\vv{t}^{\ast}}\tp\vv{W}_i +
{\vv{t}^{\ast}}\tp\vv{t}
\right)
\end{eqnarray}
The vast majority of existing point registration methods use an
isotropic covariance. The minimizer of
(\ref{eq:minimize-rigid-isotropic}) can be estimated in
closed-form using one of the methods proposed in \cite{ArunHuangBlostein87,Horn87-ortho,Umeyama91}.

\subsection{Anisotropic covariance model}
\label{subsection:anisotropic-covariance}

In this section we provide a solution for
(\ref{eq:minimize-rigid-1}) in the general case i.e. when the
covariances are anisotropic. Our formulation relies on
transforming (\ref{eq:minimize-rigid-1}) into a constrained
quadratic optimization problem and on using semi-definite positive
(SDP) relaxation to solve it, as detailed below. We denote by
$\vv{r}$ the $9\times 2$ vector containing the entries of the
3$\times$3 matrix $\mm{R}$, namely $\vv{r}:=\text{vec}(\mm{R})$.
We also denote by $\vv{\rho}$ the following rank-one positive
symmetric matrix:
\begin{equation}
\label{eq:ro-matrix}
\vv{\rho} := \vv{r} \vv{r}\tp
\end{equation}
By developing and regrouping terms, (\ref{eq:minimize-rigid-1})
can be written as the following quadratic minimization criterion
subject to orthogonality constraints:
\begin{equation}
\label{eq:minimize-rigid-2}
\left\{
\begin{array}{l}
\vv{r}^{\ast} = \arg \min 
\frac{1}{2}
\left(
\vv{r}\tp\mm{A}\vv{r} + 2 \vv{b}\tp \vv{r}
\right) \\
\vv{r}\tp\mm{\Delta}_{kl}\vv{r} = \delta_{kl},\;\;
k=1,2,3;\; l=1,2,3.
\end{array}
\right.
\end{equation}
The entries of the 9$\times$9 real symmetric matrix $\mm{A}$ and
that of the 9$\times$1 vector $\vv{b}$ are easily obtained by
identification with the corresponding terms in
(\ref{eq:minimize-rigid-1}); The entries of $\mm{A}$ and of
$\vv{b}$ are derived in the Appendix.
The entries of the six 9$\times$9 matrices $\mm{\Delta}_{kl}$ are
easily obtained from the constraint $\mm{R}\mm{R}\tp=\mm{I}$.

As already outlined, one fundamental tool for solving such a constrained quadratic
optimization problem is \textit{SDP
  relaxation}
\cite{LemarechalOustry99,LemarechalOustry2001}. Indeed, a
quadratic form such as $\vv{r}\tp\mm{A}\vv{r}$ can equivalently be
written as the matrix dot-product $\langle \mm{A}, \vv{r}\vv{r}\tp
\rangle$\footnote{The dot-product of two $n\times n$ matrices
$\mm{A}=[A_{ij}]$ and $\mm{B}=[B_{ij}]$ is
  defined as $\langle\mm{A},\mm{B}\rangle :=
  \sum\limits_{i=1}^{n} \sum\limits_{j=1}^n A_{ij}B_{ij}$.}.
Using the notation (\ref{eq:ro-matrix}) one can rewrite
(\ref{eq:minimize-rigid-2}):
\begin{equation}
\label{eq:minimize-rigid-3}
\left\{
\begin{array}{l}
(\vv{\rho}^{\ast},\, \vv{r}^{\ast}) = \arg
\min\limits_{(\vv{\rho},\vv{r})} \frac{1}{2} \left(\langle \mm{A},
\vv{\rho} \rangle + 2 \vv{b}\tp \vv{r}
\right) \\
\langle \mm{\Delta}_{kl}, \vv{\rho} \rangle = \delta_{kl},\;\;
k=1,2,3;\; l=1,2,3.\\
\vv{\rho}=\vv{r}\vv{r}\tp
\end{array}
\right.
\end{equation}
In (\ref{eq:minimize-rigid-3}) everything is linear except the last
constraint which is nonconvex. As already noticed, matrix
$\vv{r}\vv{r}\tp$ is a rank-one positive symmetric
matrix. Relaxing the positivity constraint to \textit{semi-definite
positivity} amounts to taking the convex hull of the rank-one positive
symmetric matrices. Within this context, (\ref{eq:minimize-rigid-3})
\textit{relaxes} to:
\begin{equation}
\label{eq:minimize-rigid-4}
\left\{
\begin{array}{l}
(\vv{\rho}^{\ast},\, \vv{r}^{\ast}) = \arg
\min\limits_{(\vv{\rho},\vv{r})} \frac{1}{2} \left(\langle \mm{A},
\vv{\rho} \rangle + 2 \vv{b}\tp \vv{r}
\right) \\
\langle \mm{\Delta}_{kl}, \vv{\rho} \rangle = \delta_{kl},\;\;
k=1,2,3;\; l=1,2,3.\\
\vv{\rho} \succeq \vv{r}\vv{r}\tp
\end{array}
\right.
\end{equation}
To summarize, rigid point registration with anisotropic
covariances, i.e. (\ref{eq:minimize-rigid-1}), can be formulated
as the convex optimization problem (\ref{eq:minimize-rigid-4}). It
is well known that this generally provides a very good initial
solution to a standard non-linear optimizer such as the one
proposed in \cite{WilliamsBennamoun2000}. Finally, this yields the
following algorithm illustrated in
Fig.~\ref{fig:ecmpr-anisotropicnoise}:
\begin{description} \item[\textbf{The ECMPR-rigid algorithm:}] \end{description}
\begin{enumerate}
\item \textit{Initialization}:
Set $\mm{R}^q=\mm{I}$, $\vv{t}^q=\vv{0}$. Choose the initial
covariance matrices $\mm{\Sigma}_i^q$, $i=1 \ldots n$.
\item \textit{E-step}: Evaluate the posteriors $\alpha_{ji}^q$ from
(\ref{eq:alpha-Estep}) and (\ref{eq:alpha-ij-outlier}),
$\vv{W}_i^q$ from (\ref{eq:mean-observation}), and $\lambda_i^q$
from (\ref{eq:lambda}), using  the current parameters $\mm{R}^q$,
$\vv{t}^q$, and $\mm{\Sigma}_i^q$.
\item \textit{CM-steps}:
\begin{enumerate}
\item Use SDP relaxation to estimate the new rotation matrix $\mm{R}^{q+1}$ by
  minimization of (\ref{eq:minimize-rigid-1}) with the current posteriors $\alpha_{ji}^q$ and the
current covariances $\mm{\Sigma}_i^q$;
\item Estimate the new translation vector $\vv{t}^{q+1}$ from (\ref{eq:optimal-t})
  using the new rotation $\mm{R}^{q+1}$,  the current
  posteriors, and the covariance matrices;
\item Estimate the new covariances from
  (\ref{eq:empirical-covariance}) or from
  (\ref{eq:empirical-covariance-one}) with the
  current posteriors, the new rotation matrix, and the new
  translation vector.
\end{enumerate}
\item \textit{Convergence}: Compare the new and  current
  rotations. If $\|\mm{R}^{q+1}-\mm{R}^q\|^2<\varepsilon$ then
  go to the \textit{Classification} step. Else, set the current parameter values to their new
  values and return to the \textit{E-step}.
\item \textit{Classification}: Assign each observation to a model
  point (inlier) or to the uniform class (outlier) based on the maximum a posteriori (MAP) principle:\\
  $z_j=\arg\max\limits_i \alpha_{ji}^q$.
\end{enumerate}

\section{Articulated point registration}
\label{section:articulated-pose-tracking}

\subsection{The kinematic model}
In this section we will develop a solution for the articulated
point registration problem. We will consider the case of an open
kinematic chain. Such a chain is generally composed of rigid
parts. Two adjacent parts are mechanically linked. Each link has
one, two or three rotational degrees of freedom. i.e., spherical
motions. In addition we assume that the \textit{root part} of such
an open chain may undergo a free motion with six degrees of
freedom. Consequently the articulated object motions considered
here are combinations of \textit{free} and \textit{constrained}
motions. This is more general than traditional open or closed
kinematic chains considered in standard robotics
\cite{McCarthy90}.

\begin{figure*}[htb]
\centering
\begin{tabular}{ccc}
\frame{\includegraphics[width=0.620\columnwidth,viewport=225 0 408 106,clip]{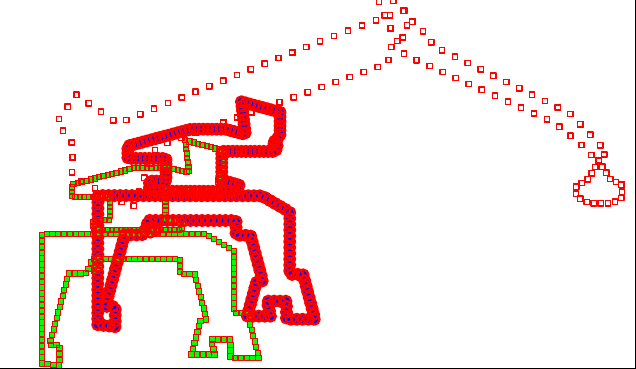}} &
\frame{\includegraphics[width=0.620\columnwidth,viewport=225 0 408 106,clip]{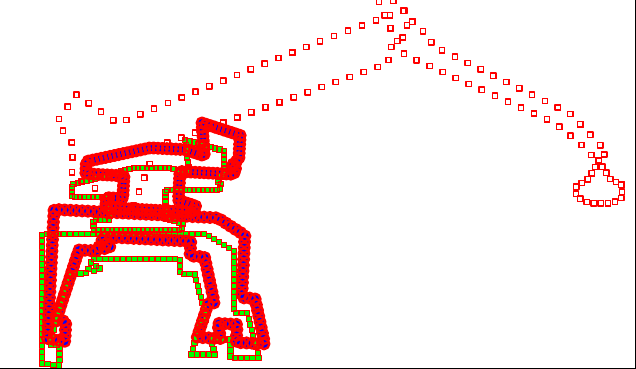}} &
\frame{\includegraphics[width=0.620\columnwidth,viewport=225 0 408 106,clip]{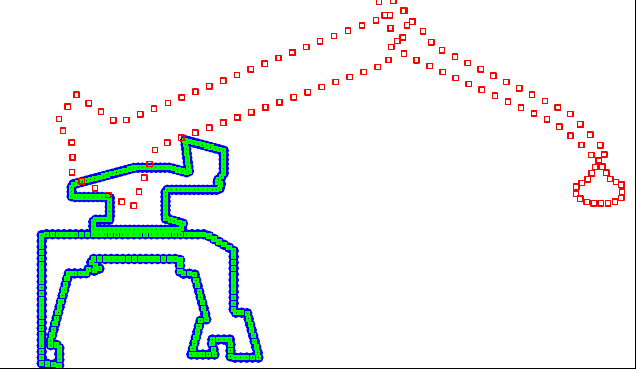}}\\
(a) & (b) & (c) \\
\end{tabular}
\begin{tabular}{ccc}
\frame{\includegraphics[width=0.620\columnwidth]{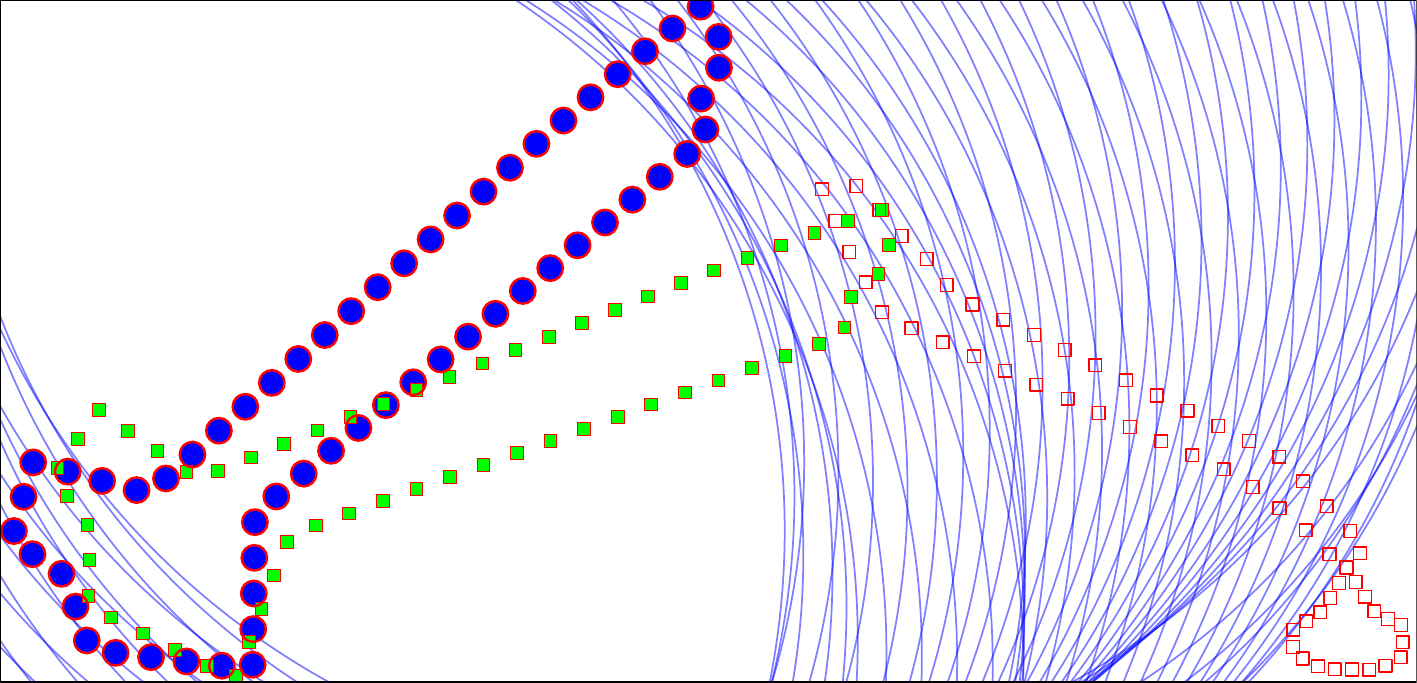}}&
\frame{\includegraphics[width=0.620\columnwidth]{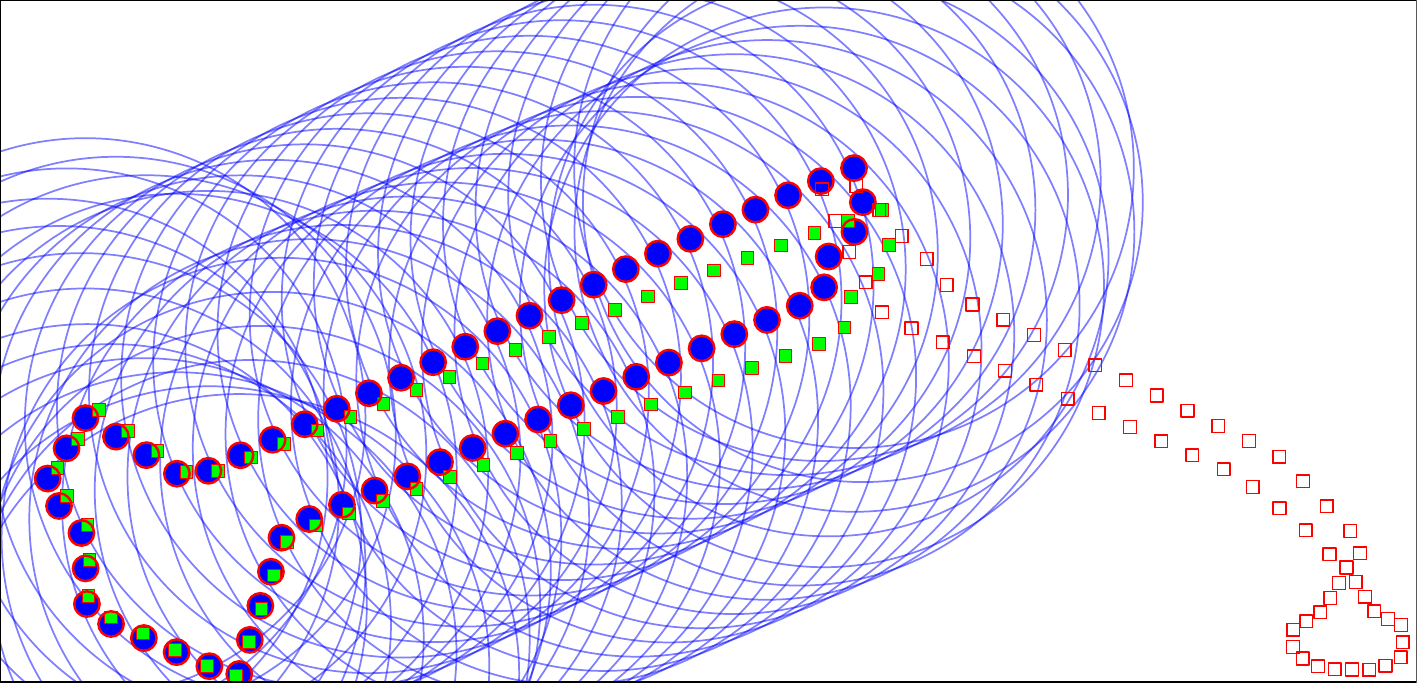}}&
\frame{\includegraphics[width=0.620\columnwidth]{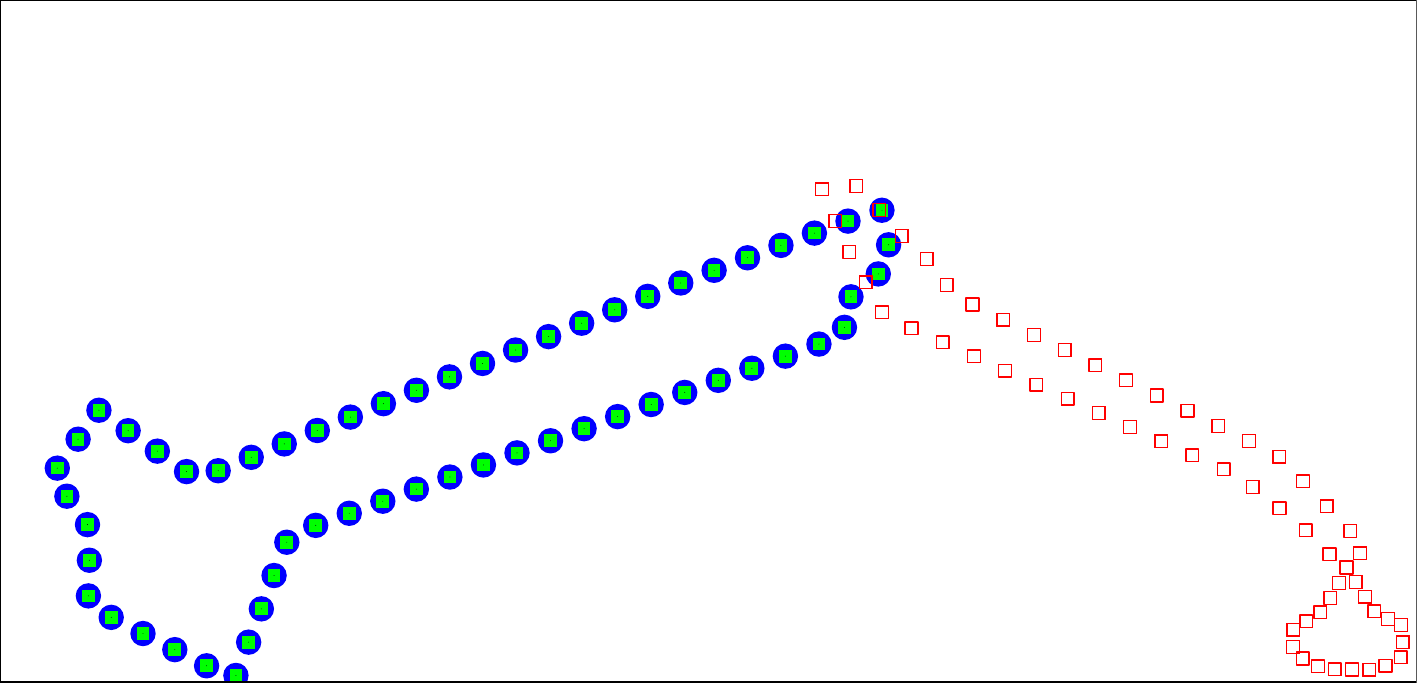}}\\
(d) & (e) & (f) \\
\end{tabular}
\begin{tabular}{ccccc}
\frame{\includegraphics[height=0.320\columnwidth]{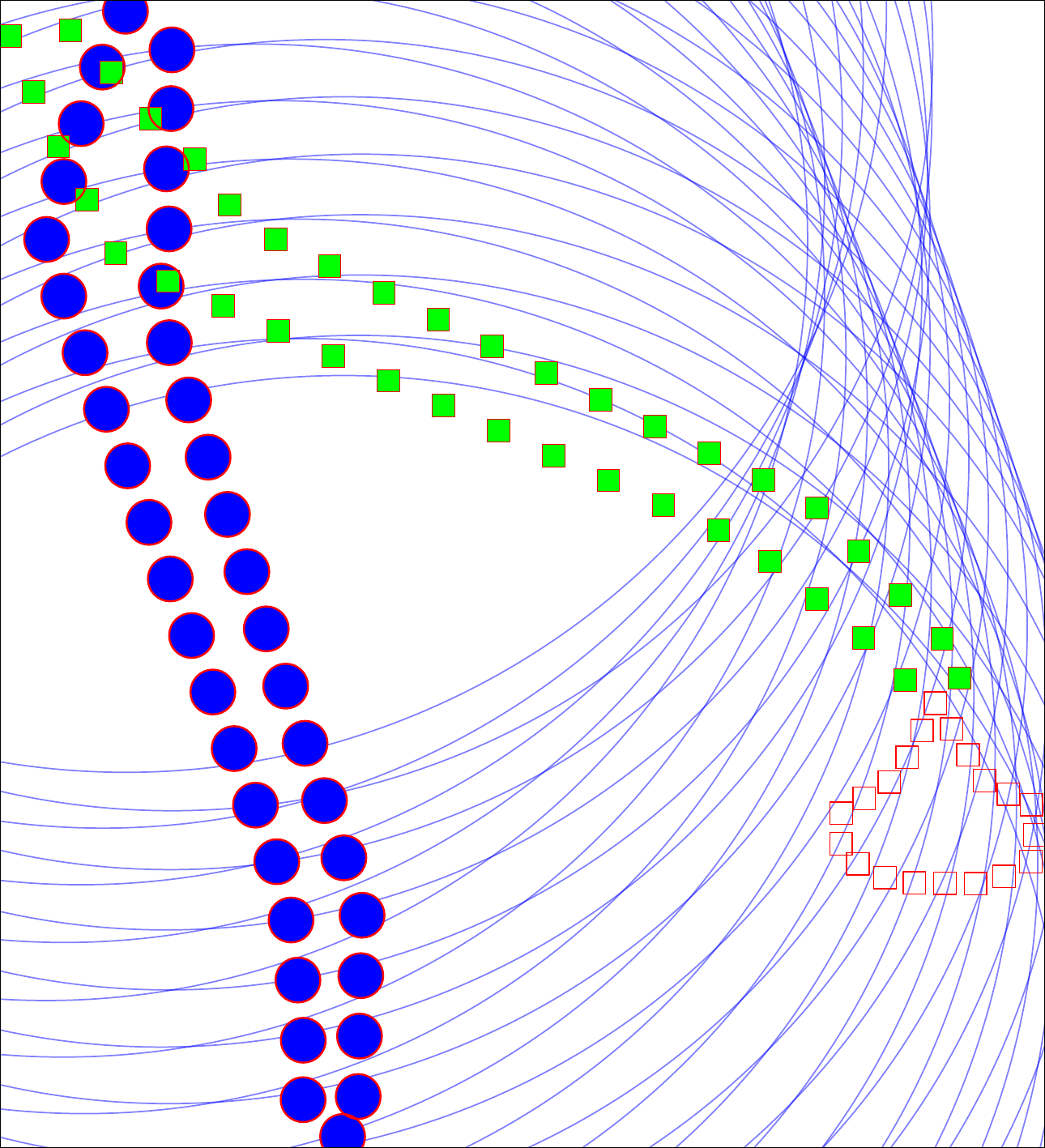}}&
\frame{\includegraphics[height=0.320\columnwidth]{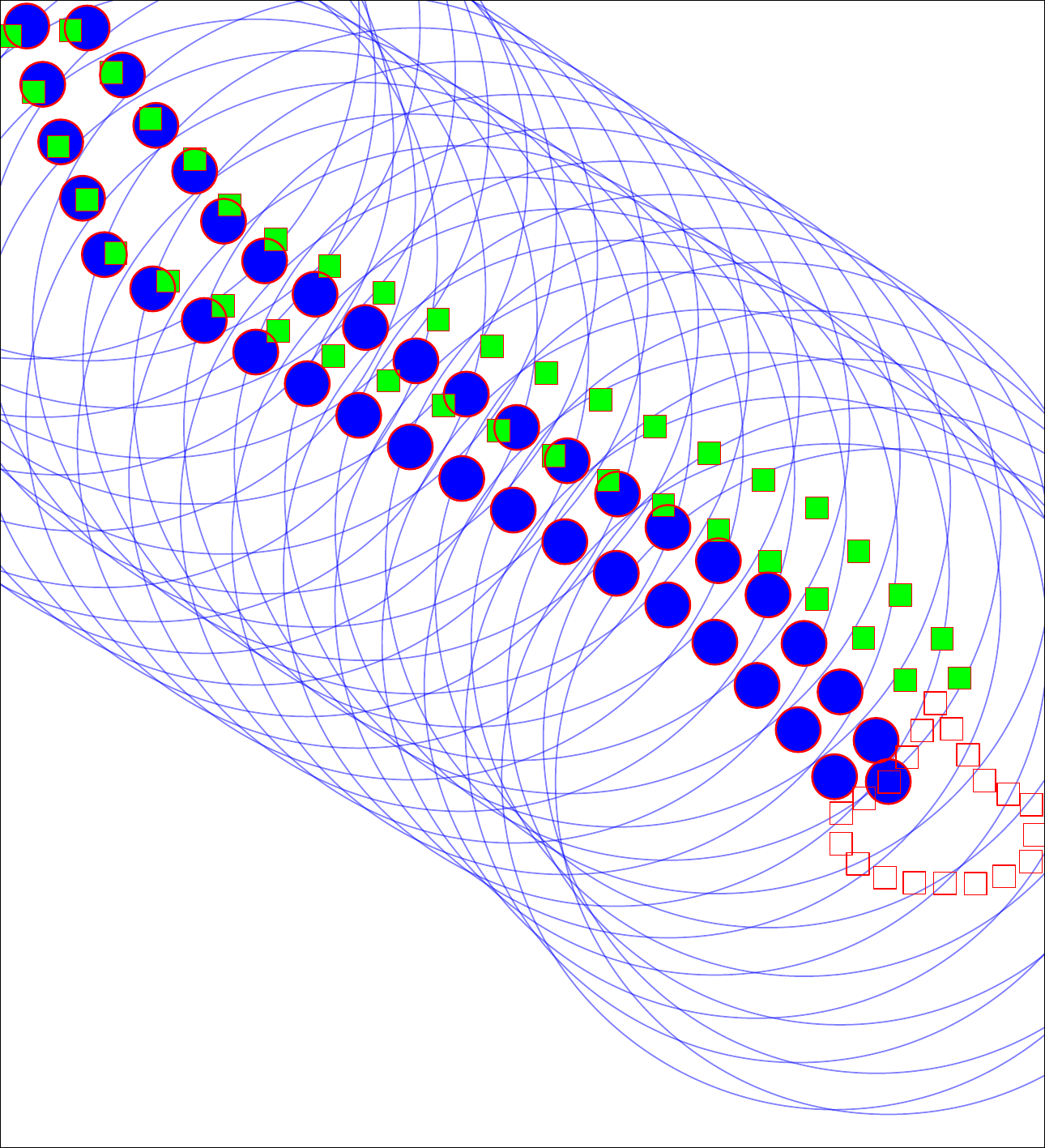}}&
\frame{\includegraphics[height=0.320\columnwidth]{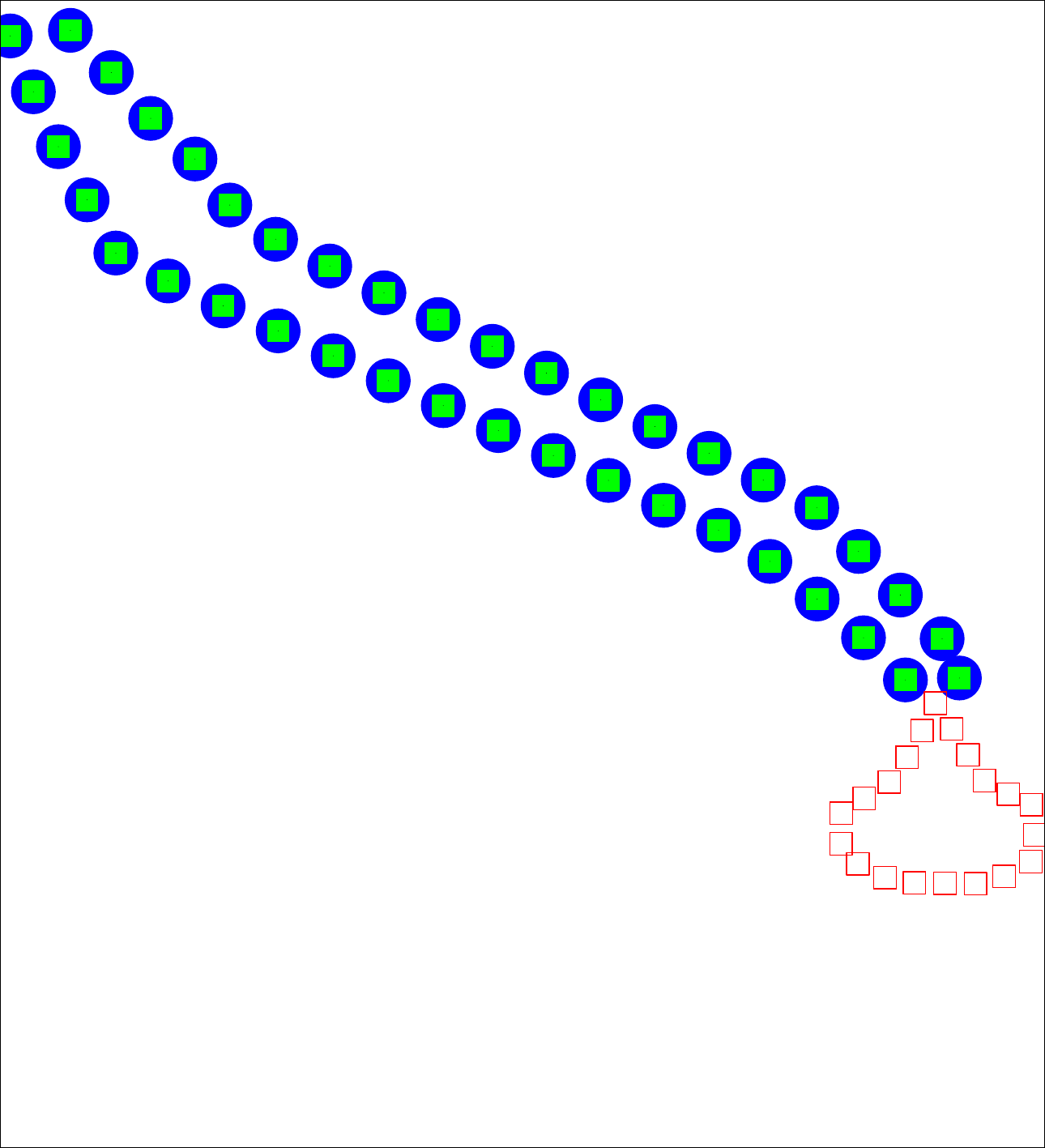}}&
\frame{\includegraphics[height=0.320\columnwidth]{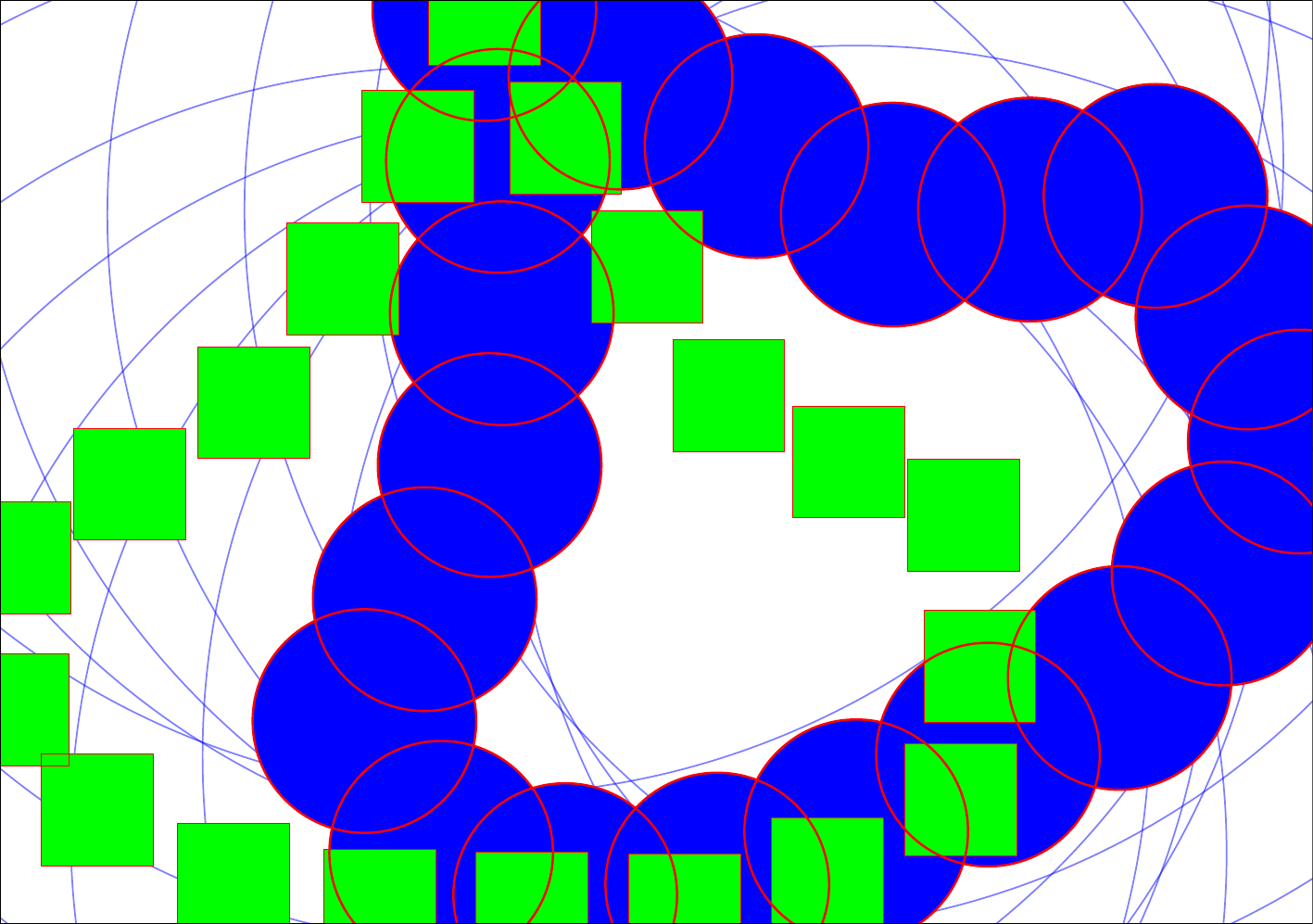}}&
\frame{\includegraphics[height=0.320\columnwidth]{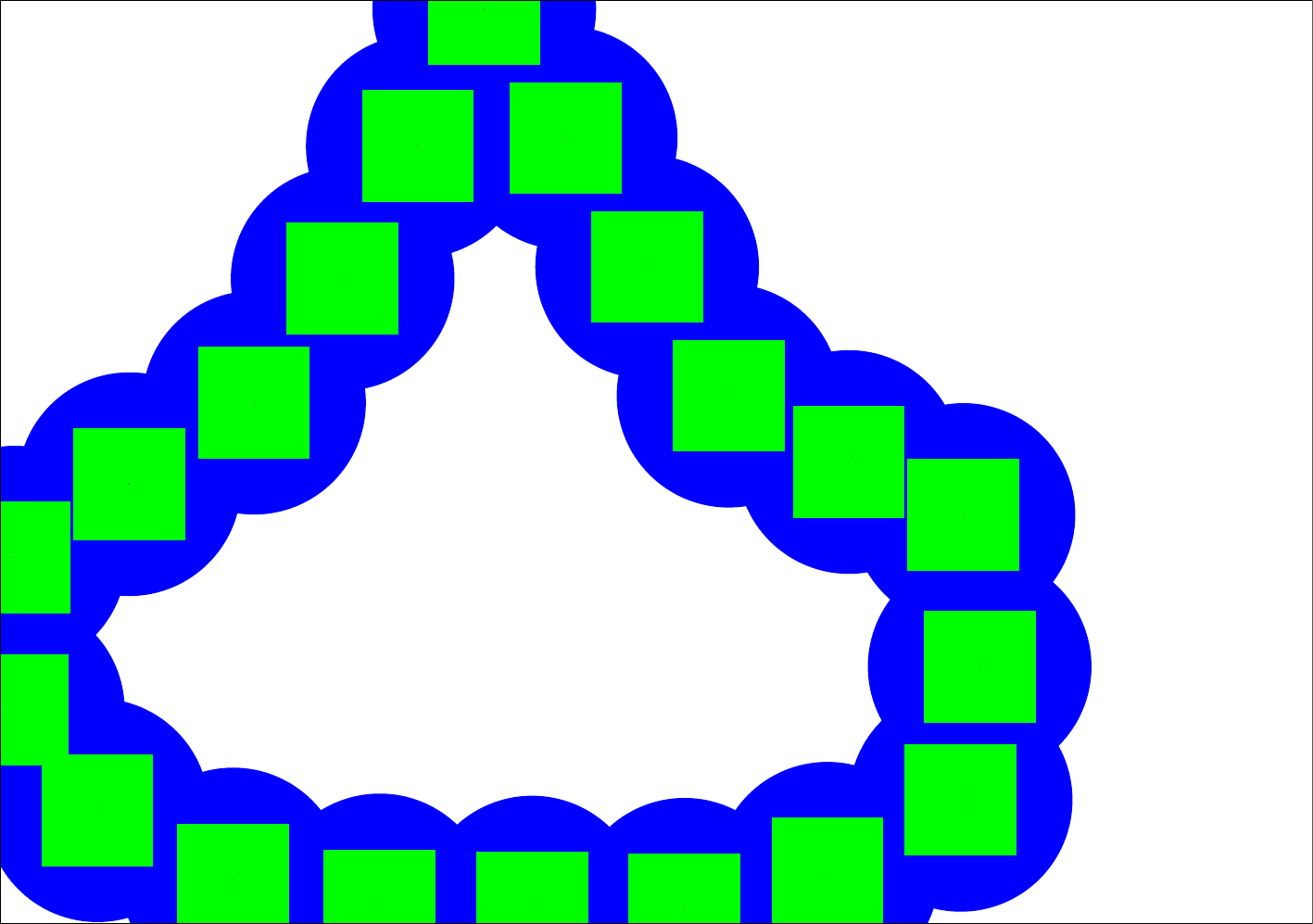}}\\
(g) & (h) & (i) & (j) & (k) \\
\end{tabular}
\caption{Illustration of the ECMPR-articulated algorithm. The
  articulated object consists of 4 rigid parts. (a)--(c): 1st, 4th and 25th iterations of ECMPR-rigid
  used to register the root part, i.e., $p=0$; the algorithm converged
  in 25 iterations. (d)--(f): The unmatched data (the outliers), are
  used to register the second part, $p=1$, in 13 iterations.
  (g)--(i): The
  third part, $p=2$, is aligned with the remaining data after 13
  iterations. (j)--(k): The fourth part, $p=3$,
  is registered with the remaining data in 4 iterations.}
\label{fig:grue-example1}
\end{figure*}

More precisely, any
rigid part $p, 1\leq p \leq P$, moves with respect to the
{\em root} part $p=0$ through a chain of {\em constrained motions}. The root part itself undergoes
a {\em free motion} with up
to six degrees of freedom, three rotations and three translations.
We assume that a partition of the set of model points is provided,
$\mathcal{X}=\{\mathcal{X}_0,\hdots \mathcal{X}_p,\hdots
\mathcal{X}_P\}$; Each subset of model points $\mathcal{X}_p =\{\vv{X}_i^{(p)}\}, 1\leq i \leq n_p$
is attached to the $p^{\text{th}}$ rigid part of the
articulated object. It is worthwhile to point out that
a partitioning of the set of
observations is not required in advance and is merely an output of
our method.
The
model point $\vv{X}_i^{(p)}$ belonging to part $p$ is transformed with:
\begin{equation}
\label{eq:kinematic-chain}
\vv{\mu}(\vv{X}_i^{(p)};\mathbf{\Theta}) =
\mathbf{R}(\mathbf{\Theta})
\vv{X}_i^{(p)} + \vv{t}(\mathbf{\Theta}),\;
\mathbf{\Theta}:=\{\vv{\Theta}_0,\hdots \vv{\Theta}_p\}
\end{equation}
The main difference between rigid and articulated motion is that
in the former case, (i.e., eq.~(\ref{eq:rigid-motion})) the
rotation matrix and translation vector are the free parameters
while in the latter case, (i.e., eq.~(\ref{eq:kinematic-chain}))
the motion of any part is constrained by both the kinematic
parameters $\vv{\Theta}_1,\hdots \vv{\Theta}_p$   and by the
motion of the root part $\vv{\Theta}_0$. For convenience we will
adopt the homogeneous representation of the Euclidean group of 3-D
rigid displacements. Hence the rotation matrix and the translation
vector can be embedded into a 4$\times$4 displacement matrix
$\mm{T}_p(\vv{\Theta})$.
The latter may well be written as a chain of homogeneous transformations:
\begin{equation}
\label{eq:articulated-model}
\mathbf{T}_p(\vv{\Theta}) = \mathbf{Q}_0(\vv{\Theta}_0) \mathbf{Q}_1(\vv{\Theta}_1)
\hdots\mathbf{Q}_p(\vv{\Theta}_p)
\end{equation}
\begin{itemize}
\item
$\mathbf{Q}_0$ describes
the free motion of the root part parameterized by
$\vv{\Theta}_0=\{\text{vec}(\mm{R}_0),\vv{t}_0\}$.
\item
Each transformation $\mathbf{Q}_p, 1\leq p \leq P$ has
two components: A \textit{fixed} component that describes a change of
coordinates, and a \textit{constrained motion} component
parameterized by one, two or three angles \cite{KRH08}.
\end{itemize}
Therefore, the estimation of the
parameter vector
$\mathbf{\Theta}$ amounts to solving a difficult inverse kinematic
problem, namely a set of non-linear equations
that are generally solved using iterative optimization methods requiring
proper initialization.

Rather than estimating the problem parameters simultaneously, in
this section we devise a closed-form solution which is based on
the formulation developed in section~\ref{section:rigid}.  We
propose the ECMPR-articulated algorithm which is built on top of
the ECMPR-rigid algorithm and which solves for the free and
kinematic parameters incrementally by considering a single rigid
part at each iteration. This contrasts with methods that attempt
to estimate all the kinematic parameters simultaneously from
data-point-to-object-part associations, as done in previous
approaches
\cite{KakadiarisMetaxas2000,PlankersFua2003,BMP04,KRH08,HNDB09}.
The motion of the root part of the articulated object is
parameterized by a rotation and a translation, while the motion of
each one of the other parts is parameterized by a rotation, hence
$\vv{\Theta}_0=\{\text{vec}(\mm{R}_0),\vv{t}_0\}$ and
$\vv{\Theta}_p=\{\text{vec}(\mm{R}_p)\}$, for all $p=1 \ldots  P$,
and:
\begin{equation}
\label{eq:rigid-Qp}
\mathbf{Q}_0 = \left[
\begin{array}{cc}
\mathbf{R}_0 & \vv{t}_0 \\
\zerovect & 1
\end{array}
\right],
\;\;\;
\mathbf{Q}_p = \left[
\begin{array}{cc}
\mathbf{R}_p & \zerovect \\
\zerovect & 1
\end{array}
\right] \;.
\end{equation}
Moreover, eq.~(\ref{eq:articulated-model}) can be written as
$\mathbf{T}_0 = \mathbf{Q}_0$, $\mathbf{T}_1 =
\mathbf{T}_0\mathbf{Q}_1 $, or more generally
$\forall p, 1\leq p \leq
P$:
\begin{equation}
\label{eq:incremental-kinematics}
\mathbf{T}_{p} = \mathbf{T}_{p-1} \mathbf{Q}_{p}
\end{equation}
which can be expanded as:
\begin{equation}
\label{eq:pose-of-part}
\mm{T}_{p}= \left[
\begin{array}{cc}
\mm{R}_{0,p-1}\mm{R}_{p} &
\tvect_{0,p-1} \\
\zerovect\tp & 1
\end{array}
\right]
\end{equation}
where the rotation matrix $\mm{R}_{0,p-1}$ and the translation
vector $\tvect_{0,p-1}$ are associated with  $\mm{T}_{p-1}$
describing the \textit{articulated pose} of body part $p-1$.

\subsection{The pose of an articulated shape}
\label{subsection:articulated}

As already mentioned, there are $n_p$ model points $\vv{X}_i^{(p)}$ associated with the
$p^{\text{th}}$ body part. Using the set of available observations
together with current estimates of their
posterior probabilities one can easily compute the set of $n_p$ virtual
observations $\vv{W}_i$ and their weights $\lambda_i$.
Therefore, the
criterion (\ref{eq:rigid-case}) allows rigid registration of the root
part as well as registration of the $p^{\text{th}}$ body
part conditioned by the articulated pose of the $(p-1)^{\text{th}}$ body part:
\begin{eqnarray}
\label{eq:root-registration}
\vv{\Theta}_{0}^{\ast} =
\arg \min_{\mm{R}_{0},\tvect_{0}} \frac{1}{2} \sum_{i=1}^{n_0} \lambda_i \|
\vv{W}_i - \mm{R}_{0}
\vv{X}_i^{(0)} - \tvect_{0} \|^2_{\mm{\Sigma}_i} \\
\label{eq:minimization-body-p}
\mm{R}_{p}^{\ast} =
\arg \min_{\mm{R}_{p}} \frac{1}{2} \sum_{i=1}^{n_p} \lambda_i \|
\vv{W}_i - \mm{R}_{0,p-1}\mm{R}_{p}
\vv{X}_i^{(p)}
-\tvect_{0,p-1}\|^2_{\mm{\Sigma}_i}
\end{eqnarray}
By introducing the following substitutions:
\begin{eqnarray}
\label{eq:bodypose-1}
\mm{U}_{p}&=&\mm{R}_{0,p-1}\mm{R}_{p}\mm{R}_{0,p-1}\tp \\
\label{eq:bodypose-3}
\vv{V}_i^{(p)}&=&\mm{R}_{0,p-1}\vv{X}_i^{(p)}
\end{eqnarray}
the minimization of (\ref{eq:minimization-body-p}) becomes:
\begin{equation}
\label{eq:theta-articulated}
\mm{U}_{p}^{\ast} = \arg \min_{\mm{U}_{p}} \frac{1}{2}\sum_{i=1}^{n_p} \lambda_i \|
\vv{W}_i - \mm{U}_{p}
\vv{V}_i^{(p)} - \tvect_{0,p-1} \|^2_{\mm{\Sigma}_i}
\end{equation}
Therefore, if the transformation $\mm{T}_{p-1}$ is known, the
parameters of the transformation $\mm{Q}_p$ can be obtained by
minimization of (\ref{eq:theta-articulated}) and from
(\ref{eq:bodypose-1}) and (\ref{eq:bodypose-3}), which is strictly
equivalent to the minimization of (\ref{eq:rigid-case}). To
summarize, we obtain the following algorithm illustrated in
Fig.~\ref{fig:grue-example1}:
\begin{description} \item[\textbf{The ECMPR-articulated algorithm:}] \end{description}
\begin{enumerate}
\item \textit{Rigid registration of the root part}: Initialize the \textit{current} set of data points $\mathcal{Y}^{(0)}$ with the whole data
set. Apply the
  ECMPR-rigid algorithm to the data set $\mathcal{Y}^{(0)}$ and
  to the set of model points associated with the root part,
  $\mathcal{X}_0$, in order to estimate the pose of the root part. Compute
  $\mm{T}_0$ using (\ref{eq:rigid-Qp}).
  Classify the data points into inliers and outliers. Remove the
  inliers from $\mathcal{Y}^{(0)}$ to generate a \textit{new} data set $\mathcal{Y}^{(1)}$.
\item \textit{For each $p = 1\ldots P$},
 \textit{rigid registration of the $p^{th}$ part}: Apply the
  ECMPR-rigid algorithm to the current set of data points
  $\mathcal{Y}^{(q)}$ and to the set $\mathcal{X}_p$.
Estimate $\mm{R}_p$ from (\ref{eq:bodypose-1}) and (\ref{eq:theta-articulated}).
Compute
  $\mm{Q}_p$ and then $\mm{T}_p$ using (\ref{eq:rigid-Qp}) and
  (\ref{eq:incremental-kinematics}). Classify the data points into
  inliers and outliers. Remove the inliers from  $\mathcal{Y}^{(q)}$ to generate  $\mathcal{Y}^{(q+1)}$.
\end{enumerate}

\section{Experimental results}
\label{section:Experimental results}

We carried out a large number of experiments with both algorithms.
ECMPR-rigid was applied to simulated data to assess the
performance of the method with respect to (i) the initialization
parameters, (ii) the amplitude of Gaussian noise added to the data,
 and (iii) the percentage of outliers drawn from
a uniform distribution. ECMPR-rigid was also applied to a real data
set and compared with TriICP.
ECMPR-articulated was applied to a
simulated data set to illustrate the method,
Fig.~\ref{fig:grue-example1} as well as to the problem of hand
tracking with both real and simulated data.

In all the experiments described in this section, ECMPR-rigid's parameters were
  initialized the same way: the rotation is initialized with
  the identity matrix and the translation with the zero vector. Notice
  that ECMPR-rigid resides in the inner loop of ECMPR-articulated,
  i.e., section~\ref{subsection:articulated}.

\subsection{Experiments with ECMPR-rigid}

\begin{figure}[t!]
\centering
\subfigure[Correct matches]{\includegraphics[height=0.500\columnwidth,viewport=20 0 776 418,clip]{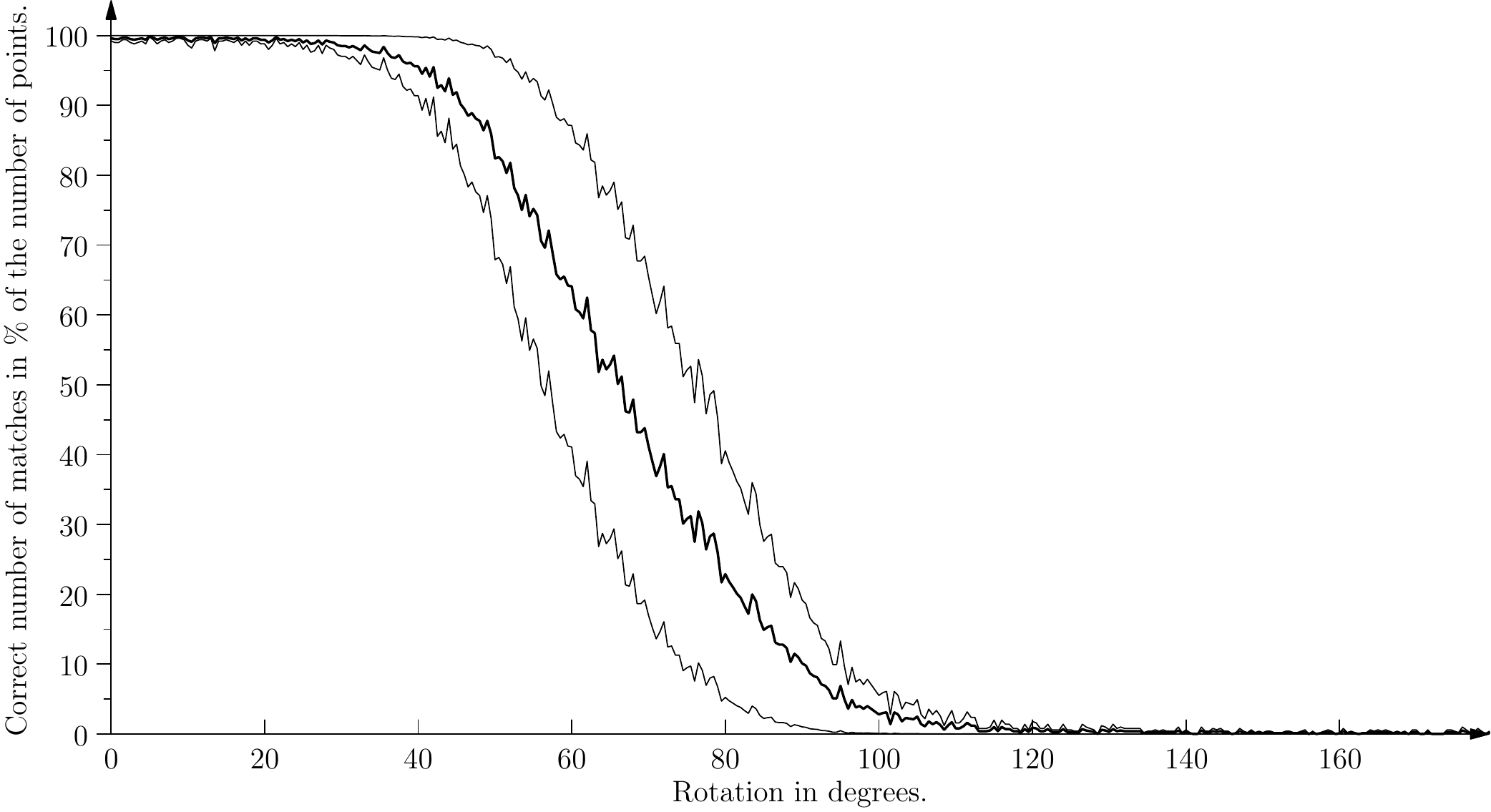}}
\subfigure[Error in rotation]{\includegraphics[height=0.500\columnwidth,viewport=20 0 776 418,clip]{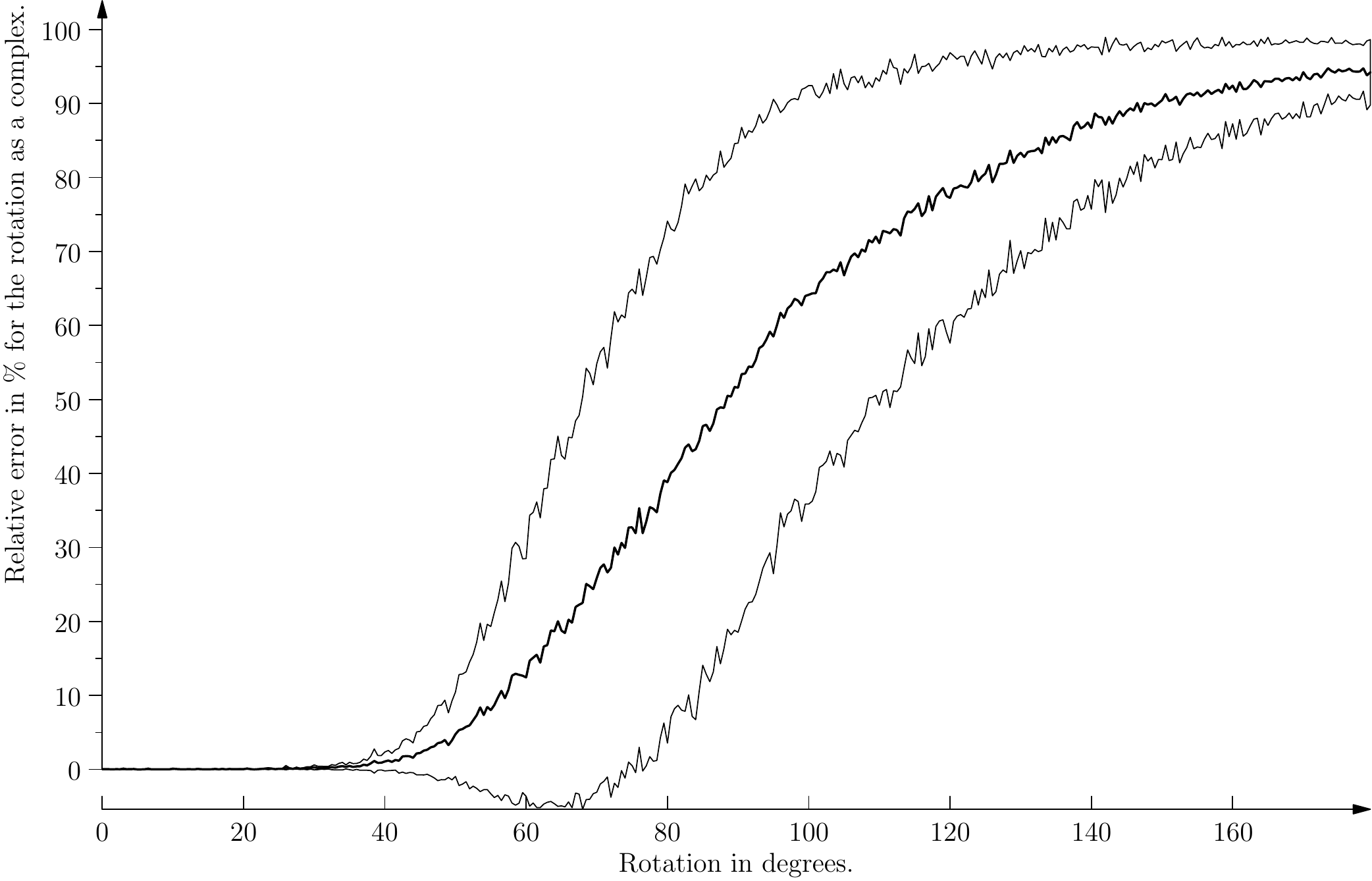}}
\subfigure[Error in translation]{\includegraphics[height=0.500\columnwidth,viewport=20 0 776 418,clip]{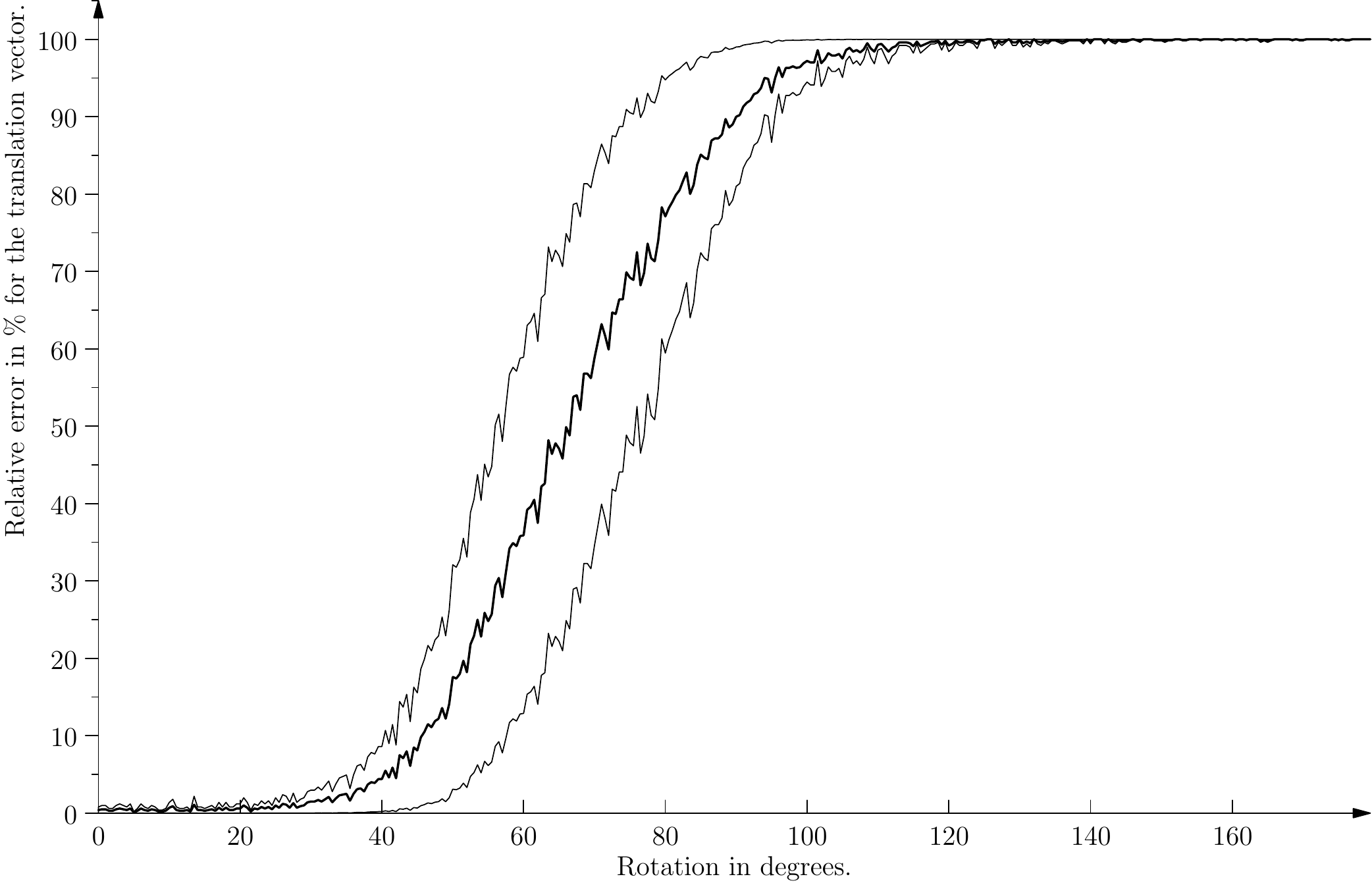}}
\caption{Statistics obtained with ECMPR-rigid over a large number of
  trials. The percentage of correct matches (a), and relative errors in
  rotation (b) and translation (c) are shown as a function of the
  ground-truth rotation
  angle between the set of data points and the the set of model points, in the presence of
  outliers. All the runs of the algorithm where initialized with a
  zero rotation angle. The three plots correspond to the means (central curves) and
  to the means +/- the standard deviation (upper and lower curves) computed over 1,000 trials.}
\label{fig:statistics}
\end{figure}

We carried out several experiments with ECMPR-rigid and with the
Trimmed Iterative Closest Point algorithm (TriICP)
\cite{ChetverikovStepanovKrsek2005}, which is a robust
implementation of ICP using random sampling. These experiments are summarized in
Table~\ref{table:rigid-registration} and on
Fig.~\ref{fig:ecmpr-anisotropicnoise} and Fig.~\ref{fig:statistics}.

In all these experiments we considered 15
model points corresponding to the clusters' centers in the mixture
model, as well as 25
observations: 15 inliers and 10 outliers. The inliers are generated
from
the model points: they are rotated, translated, and corrupted by
noise. All the outliers in all the experiments are drawn from a
uniform distribution spanning the bounding box of the set of
observations.

In the examples shown in Table~\ref{table:rigid-registration} and
on Fig.~\ref{fig:ecmpr-anisotropicnoise} the inliers are rotated
by $25^0$ and then translated using a randomized vector. The first
example (first row in Table~\ref{table:rigid-registration}) is
noise free. We simulated anisotropic Gaussian noise that was added
to the inliers in the second and third examples. This noise is
centered at each inlier location and is drawn from two
one-dimensional Gaussian probability distributions with two
different variances along each dimension. The variances were
allowed to vary between 10\% and 100\% of box bounding the set of
observations. In all the reported experiments, ECMPR-rigid was
initialized with a null rotation angle (the identity matrix), a
null translation vector, and with large variances. We used the
same data with TriICP. Unlike our method, ICP methods require
proper initialization, in particular in the presence of outliers.
TriICP embeds multiple initializations using a random sampling
strategy, which explains the large number of iterations of this
method \cite{ChetverikovStepanovKrsek2005}.

Additionally, we performed a large number of trials with
ECMPR-rigid in the anisotropic covariance case (second row in
Table~\ref{table:rigid-registration}). The inliers are rotated
with an angle that varies between $0^{0}$ and $180^{0}$. For each
angle we performed 1,000 trials. Fig.~\ref{fig:statistics} shows
the percentage of correct matches (a), the relative error in
rotation (b), and the relative error in translation (c) as a
function of the ground-truth rotation angle between the sets of data and model
points. The plotted curves correspond to the mean values and to
the variances computed over 1,000 trials for each rotation.


ECMPR-rigid behaves very well in the presence of both
high-amplitude anisotropic Gaussian noise and outliers. The
anisotropic covariance model advocated in this paper yields better
results than the isotropic model both in terms of parameter
estimation and number of correct assignments. The errors in
rotation and translation are consistent with the level of noise
added to the inliers; Overall the performance of ECMPR-rigid is
very robust in the presence of outliers. This is a crucial feature
of the algorithm that directly conditions the robustness of
ECMPR-articulated, since the former resides in the inner loop of
the latter.
\begin{table*}[t!]
\caption{Summary of experiments with simulated data using ECMPR-rigid
  and TriICP.}
\begin{center}
\begin{tabular}{|p{0.13\textwidth}|p{0.09\textwidth}|p{0.09\textwidth}|p{0.09\textwidth}|p{0.10\textwidth}|p{0.12\textwidth}|p{0.095\textwidth}|p{0.08\textwidth}|}
\hline
Algorithm & Simulated \newline noise & Covariance \newline model & Number of
\newline iterations
& Error in \newline rotation (\%) & Error in \newline translation
(\%) & Correct
\newline matches (\%) & Process. \newline time (ms) \\
\hline
ECMPR
& -- & anisotropic & 20 &0.0 &
0.0 & 100 & 16.8\\ \hline
ECMPR (Fig.~\ref{fig:ecmpr-anisotropicnoise}) & anisotropic & anisotropic & 36
& 1.5 & 5.6  & 76 & 31.4 \\ \hline
ECMPR
& anisotropic & isotropic
& 35 & 8.1 &
 26.3 & 52 & 16.8 \\ \hline
TriICP & --          & -- & 217 & 0.0 & 0.0 & 100 & 67.2\\ \hline
TriICP & anisotropic & -- & 215  & 10.3 & 6.5 & 28 & 63.2\\ \hline
\end{tabular}
\end{center}
\label{table:rigid-registration}
\end{table*}

To farther assess the algorithms' performance, we computed the
percentage of correct matches (see
Table~\ref{table:rigid-registration}), namely the number of
observations that were correctly classified over the total number
of observations. In case of ECMPR-rigid, this classification is
based on the maximum a posteriori (MAP) principle: each
observation $j$ is assigned to the cluster $k$ (either a Gaussian
cluster for a model point or a uniform class for an outlier) such that
$k=\arg\max_i(\alpha_{ji})$.
This
implies that each data point, which is not an outlier, is assigned
to one model point but there may be several data points assigned
to the same model point. ICP algorithms use a different assignment
strategy, namely they retain the closest data point for each model
point and they apply a threshold to this point-to-point distance
to decide whether the assignment should be validated or not. For
these reasons, the counting of matches has a different meaning
with ECMPR and with ICP. For example, in the case of an
anisotropic covariance model (Table~\ref{table:rigid-registration}
second row and Fig.~ \ref{fig:ecmpr-anisotropicnoise}), ECMPR
assigned 3 outliers to 3 model points while 3 inliers were
incorrectly assigned. In the case of an isotropic covariance model
(Table~\ref{table:rigid-registration}, third row), 4 outliers were
assigned to 4 model points while 8 inliers were incorrectly
assigned. In the presence of both anisotropic noise and outliers,
TriICP rejected 18 data points, namely 10 outliers and 8 inliers.
Comparing  correct matches then may not be straightforward. A more
meaningful comparison can be made by looking at the transformation
estimation. It appears that ECMPR has superior performance with
smaller rotation and translation errors.

As we already mentioned and as observed by others, the
initialization of TriICP (and more generally of ICP algorithms) is
crucial to obtain a good match. Starting from any initial guess,
ICP converges very fast (4 to 5 iterations on average).
However, ICP is easily trapped in a local minimum. To overcome
this problem,  TriICP combines ICP with a random sampling method:
The space of rotational parameters is uniformly discretized and an
initial solution is randomly drawn from this space.

\begin{figure}[b!]
\centering
\subfigure[First image pair]{\includegraphics*[width=0.49\columnwidth,viewport=300 50 730 540,clip]{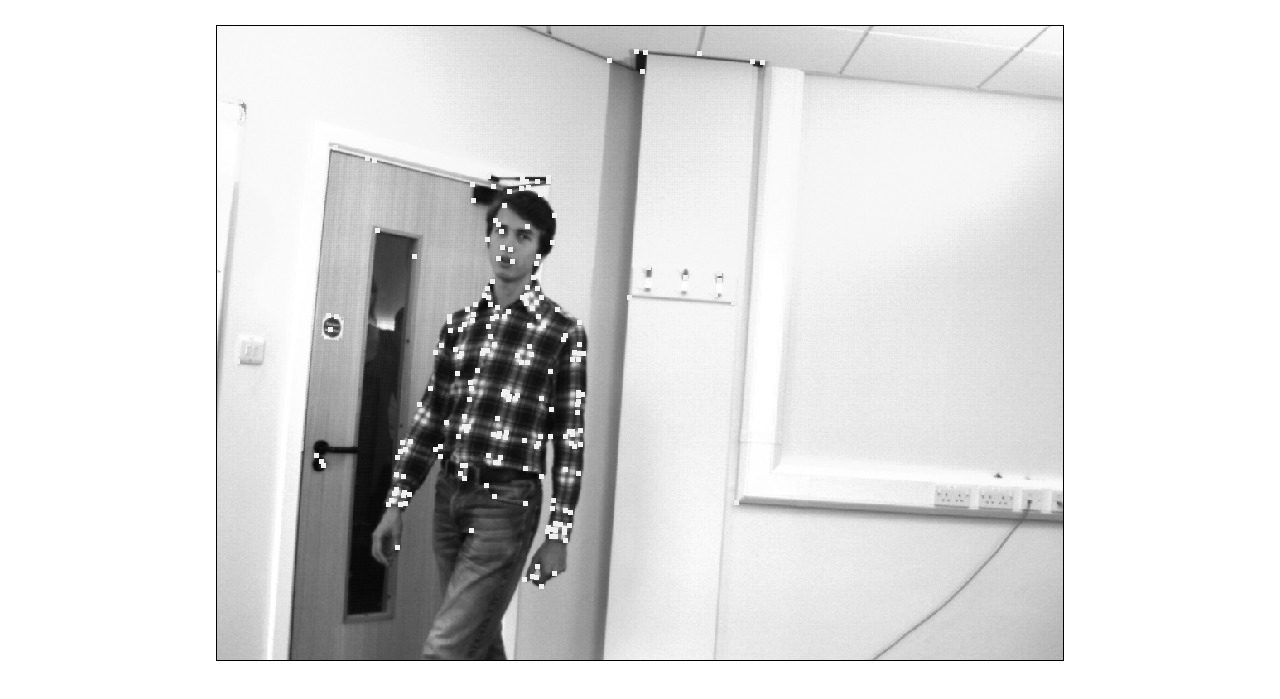}
\includegraphics*[width=0.49\columnwidth,viewport=300 50 730 540,clip]{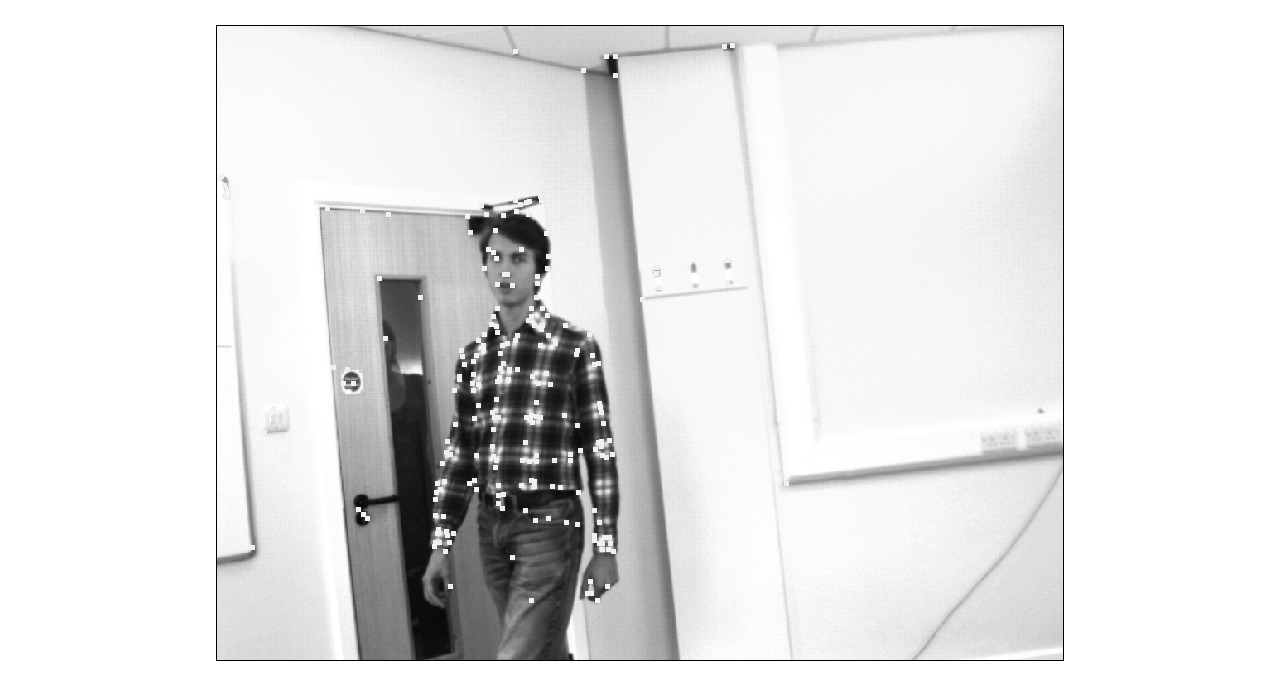}}\\
\subfigure[Second image pair]{\includegraphics*[width=0.49\columnwidth,viewport=300 50 730 540,clip]{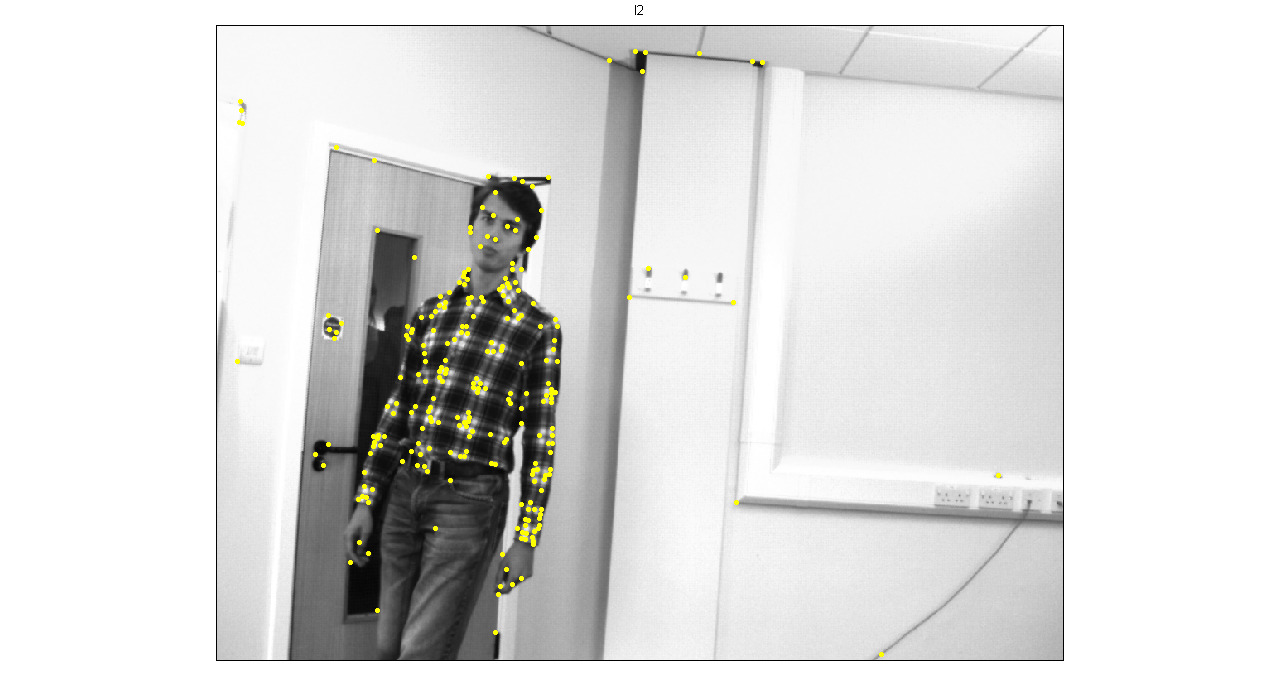}
\includegraphics*[width=0.49\columnwidth,viewport=300 50 730 540,clip]{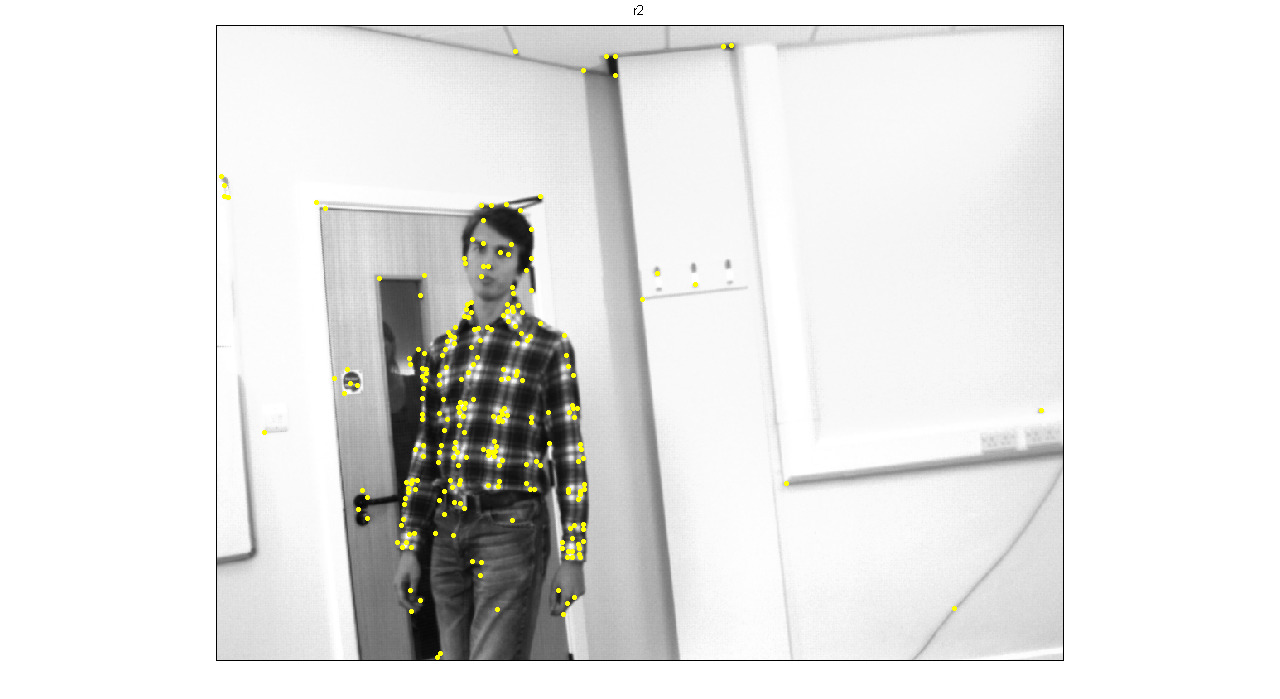}}\\
\subfigure[3D point sets]{\frame{\includegraphics*[height=0.15\textheight]{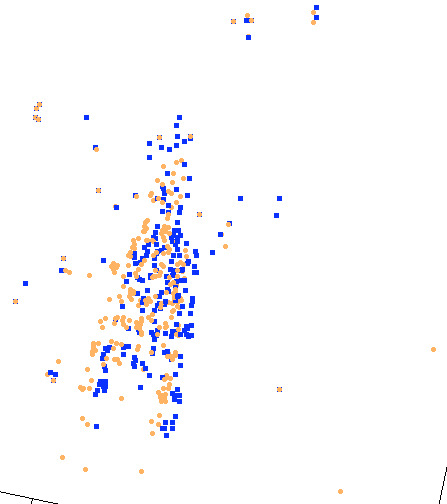}}}
\subfigure[ECMPR-rigid]{\includegraphics*[height=0.15\textheight,viewport=300 50 600 540,clip]{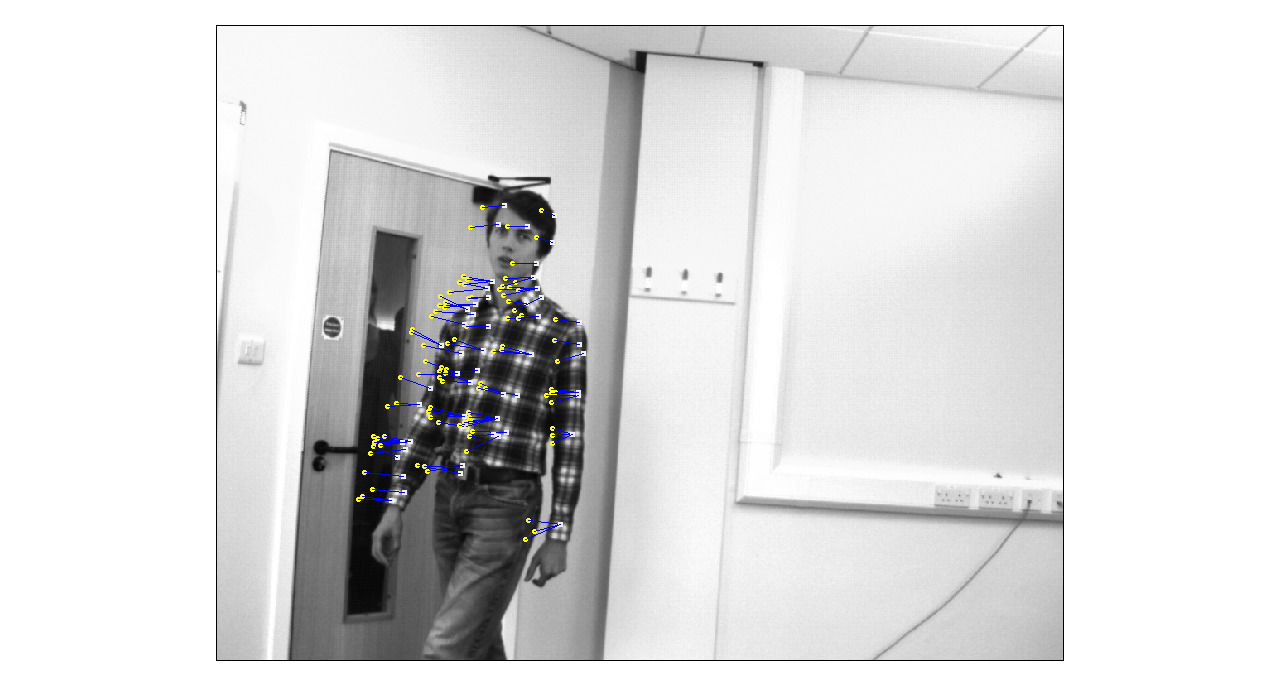}}
\subfigure[ICP]{\includegraphics*[height=0.15\textheight,viewport=300 50 600 540,clip]{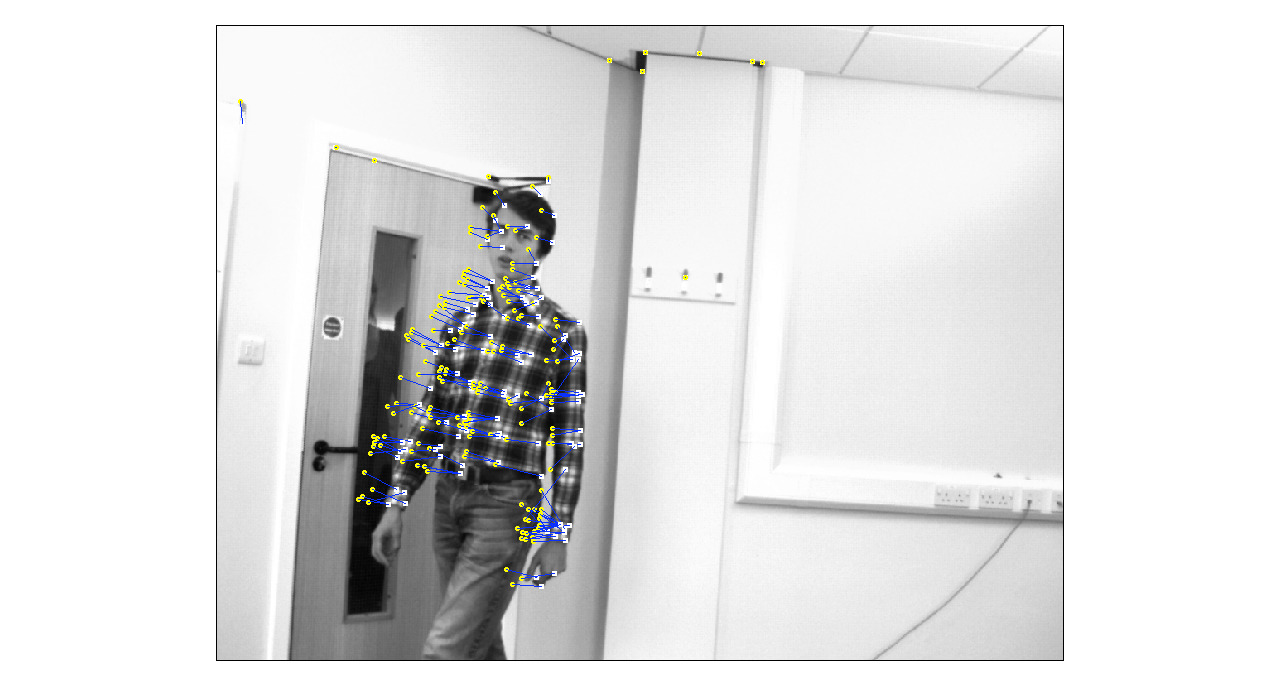}}
\caption{Comparison between ECMPR-rigid and ICP applied to stereo data. (a) The first stereo image
  pair of a walking person. (b) The second stereo pair. The person performed a translational motion of 280 mm towards the camera and from right to left.
  (c) The two sets of 3D points before registration (223 model points and 249 data points).
  The result of matching with (d) ECMPR and with (e) ICP are  shown superimposed onto the left image of the first pair. In this example, ECMPR found 95 inliers while ICP found 177 inliers. Both algorithms estimated the correct rotation. ICP failed to estimate the correct translation (see Table~\ref{table:rigid-compare} for a quantitative comparison).}
\label{fig:stereo-data}
\end{figure}

\begin{table}[htb]
\caption{Comparison between ECMPR-rigid and ICP applied to the stereo data of Fig.~\ref{fig:stereo-data}.}
\begin{center}
\begin{tabular}{|p{0.16\columnwidth}|p{0.14\columnwidth}|p{0.15\columnwidth}|p{0.15\columnwidth}|p{0.16\columnwidth}|}
\hline
Algorithm &  
Number of \newline iterations & 
Number of \newline inliers & 
Translation \newline error (\%) & 
Minimization \newline error (mm) \\
\hline
ECMPR & 12 & 95 & 12.5 & 4.6\\
\hline
ICP (Worst) &4 & 227 &34.6 & 6.0\\
\hline
ICP (Best) &6 &177 &22.4 & 4.9\\
\hline
\end{tabular}
\end{center}
\label{table:rigid-compare}
\end{table}


We also applied both ECMPR-rigid and ICP to real data obtained with a
stereo camera pair as shown on Fig.~\ref{fig:stereo-data} and Table~\ref{table:rigid-compare}: The two stereo image
pairs of a walking person were grabbed at two diffierent time
instances. Two sets of 3D points were
reconstructed from these two image pairs, Fig.~\ref{fig:stereo-data}-(c). The first set has 223 ``model" points and the second set has 249 ``data" points. These 3D
points belong either to the walking person or to the static
background. Fig.~\ref{fig:stereo-data}-(d) shows the matches found by ECMPR-rigid and Fig.~\ref{fig:stereo-data}-(e) shows the matches found by ICP. Table~\ref{table:rigid-compare} summarizes the results. Both algorithms were initialized with $\mm{R}=\mm{I}$ and $\vv{t}=\vv{0}$. The error in translation is computed with $\|\vv{t}-\vv{t}_g\|/\|\vv{t}_g\|$ where $\vv{t}$ is the estimated translation vector and $\vv{t}_g$ is the ground truth. The minimization error is computed with the square root of $1/n_{in} \sum_{i=1}^{n_{in}}\|\vv{Y}_i - \mm{R} \vv{X}_i - \vv{t}\|^2$ where $n_{in}$ is the number of inliers estimated by each algorithm. ICP was run with different threshold values. In all cases (ECMPR and ICP) the rotation matrix is correctly estimated. 

\subsection{Experiments with ECMPR-articulated}

We tested ECMPR-articulated on a hand-tracking task, with both
simulated and real data. 
We note that recent work in this topic uses specific constraints such as skin texture, skin shading\cite{GorceParagiosFleet2008} or skin color \cite{Hamer2009} that are incorporated into the hand model, together with a variational framework \cite{GorceParagiosFleet2008} or a probabilistic graphical \cite{Hamer2009} model that are tuned to the task of hand tracking. We did not attempt to devise such a special-purpose hand tracker from our general-purpose articulated registration algorithm.
\begin{figure}[h!]
\centering
\includegraphics*[width=0.190\columnwidth,viewport=30 30 400 400,clip]{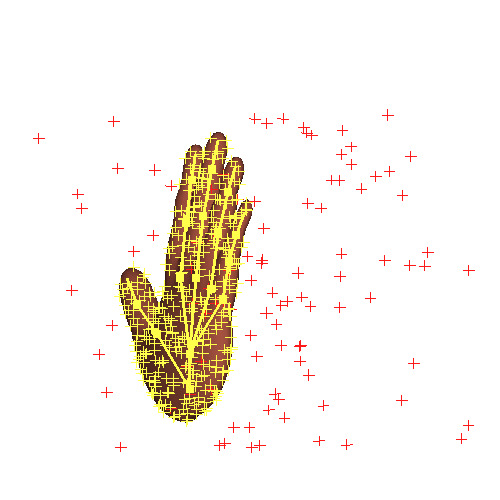}
\includegraphics*[width=0.190\columnwidth,viewport=30 30 400 400,clip]{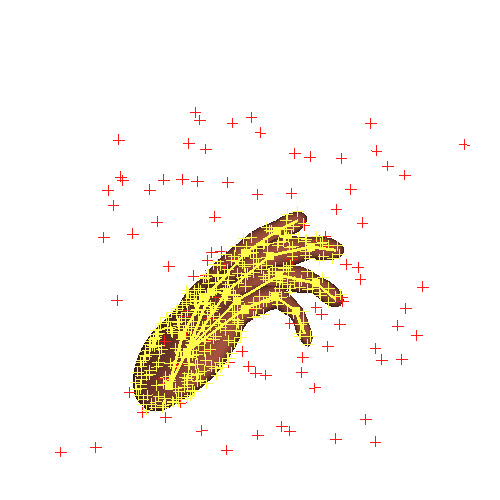}
\includegraphics*[width=0.190\columnwidth,viewport=30 30 400 400,clip]{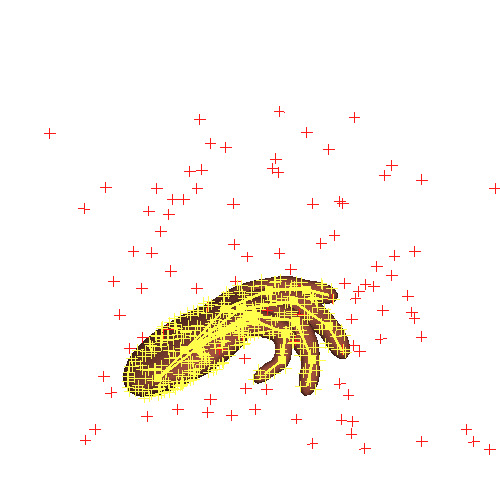}
\includegraphics*[width=0.190\columnwidth,viewport=30 30 400 400,clip]{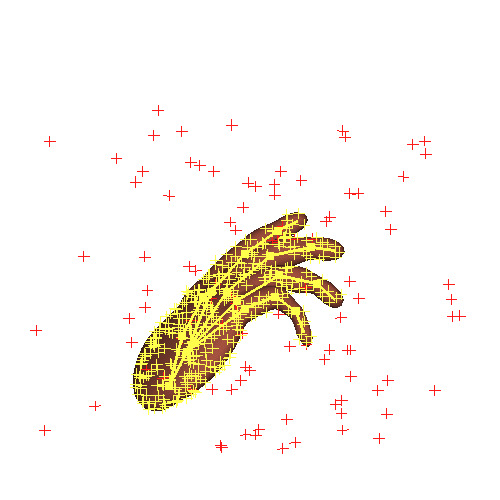}
\includegraphics*[width=0.190\columnwidth,viewport=30 30 400 400,clip]{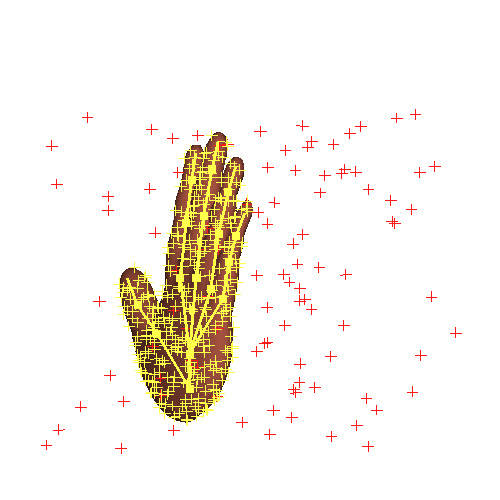}\\
\includegraphics*[width=0.190\columnwidth,viewport=30 30 400 400,clip]{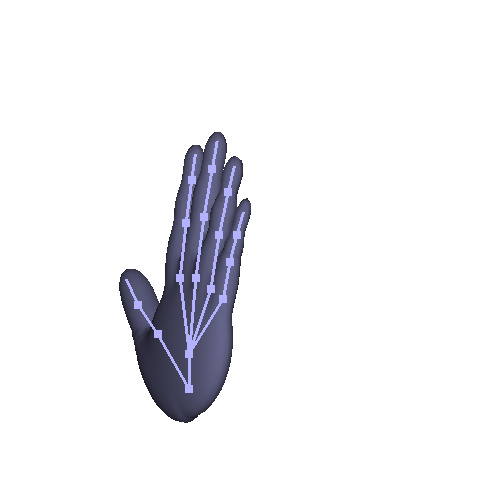}
\includegraphics*[width=0.190\columnwidth,viewport=30 30 400 400,clip]{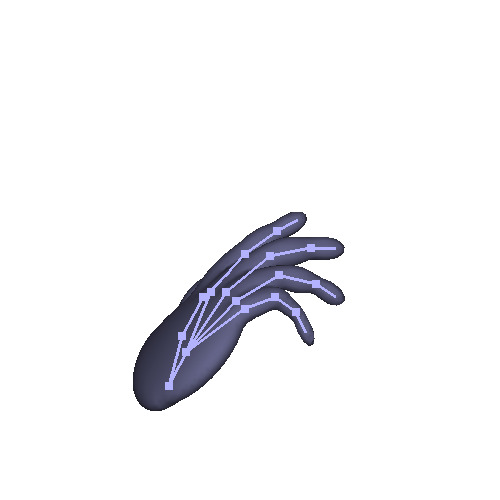}
\includegraphics*[width=0.190\columnwidth,viewport=30 30 400 400,clip]{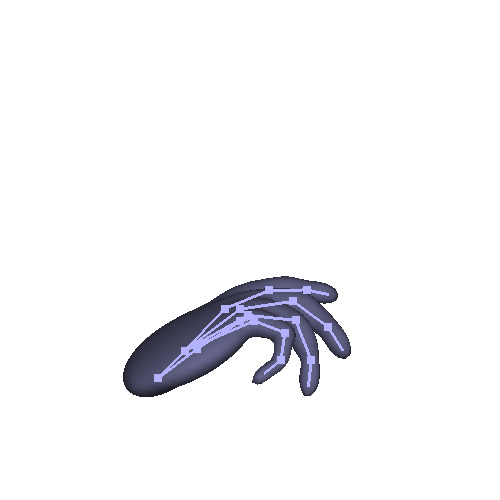}
\includegraphics*[width=0.190\columnwidth,viewport=30 30 400 400,clip]{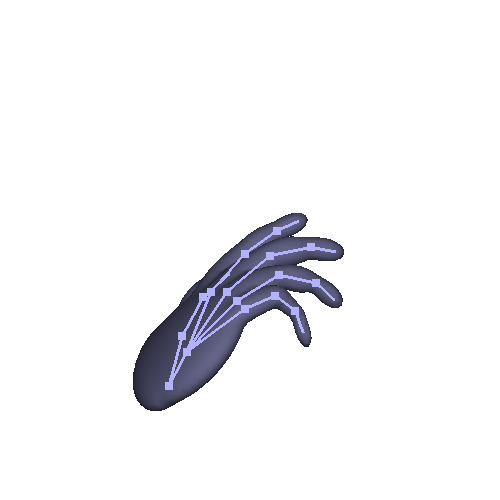}
\includegraphics*[width=0.190\columnwidth,viewport=30 30 400 400,clip]{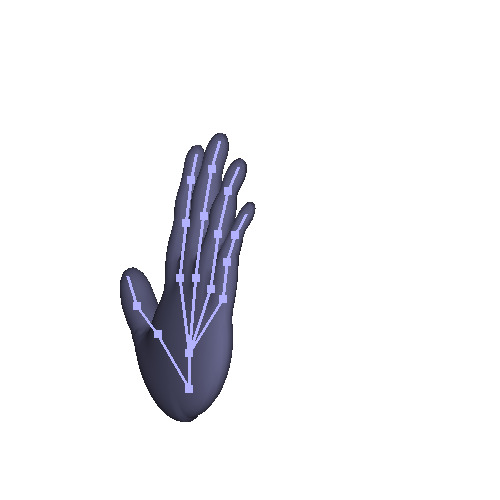}\\
\includegraphics*[width=0.190\columnwidth,viewport=30 30 400 400,clip]{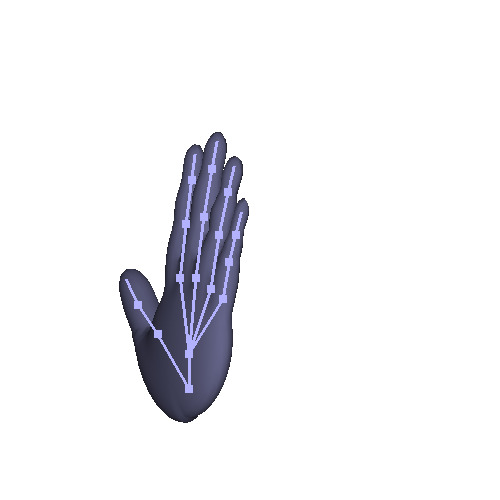}
\includegraphics*[width=0.190\columnwidth,viewport=30 30 400 400,clip]{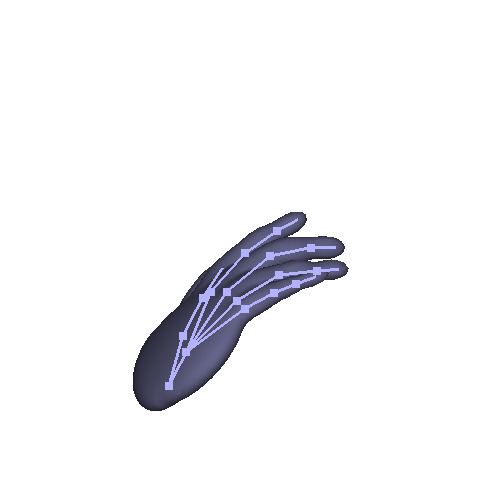}
\includegraphics*[width=0.190\columnwidth,viewport=30 30 400 400,clip]{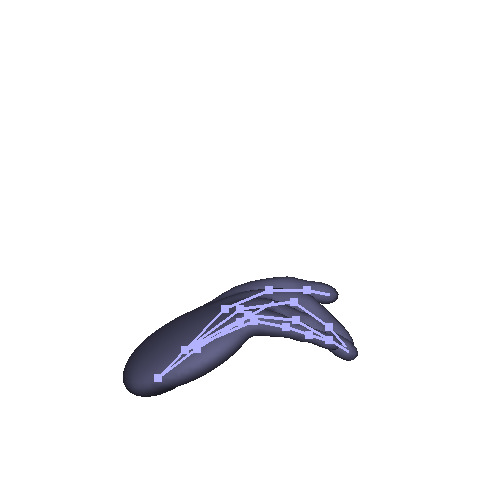}
\includegraphics*[width=0.190\columnwidth,viewport=30 30 400 400,clip]{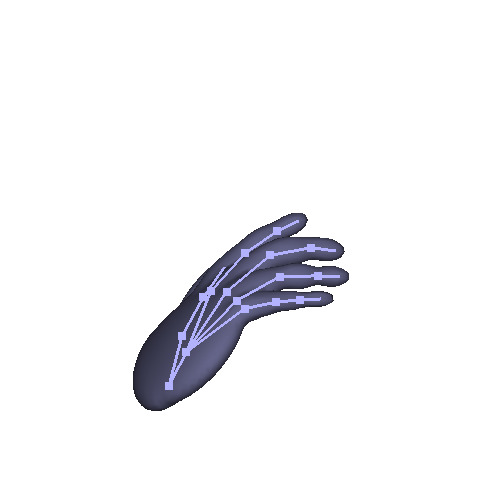}
\includegraphics*[width=0.190\columnwidth,viewport=30 30 400 400,clip]{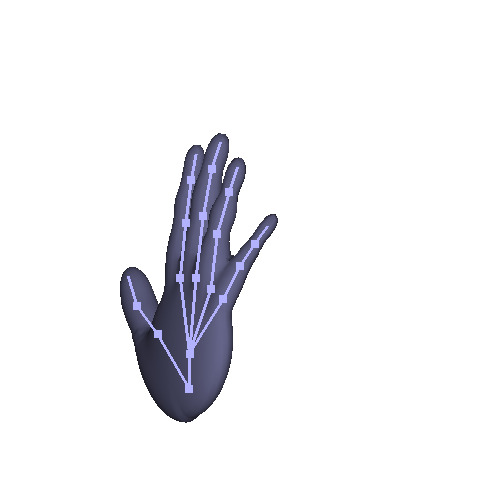}
\caption{Top row: the ground-truth of the simulated
  poses and the
  simulated data (inliers and outliers). Milddle row: a correct registration result.
  Bottom row: ECMPR failed to correctly estimate all the
  kinematic parameters due
  to an improper initialization of the covariance matrix.}
\label{fig:simulatedhand-1}
\end{figure}
\begin{figure}[h!]
\centering
\includegraphics*[width=0.190\columnwidth,viewport=30 30 400 400,clip]{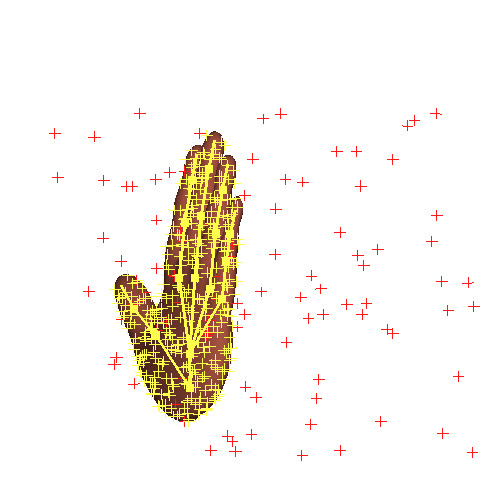}
\includegraphics*[width=0.190\columnwidth,viewport=30 30 400 400,clip]{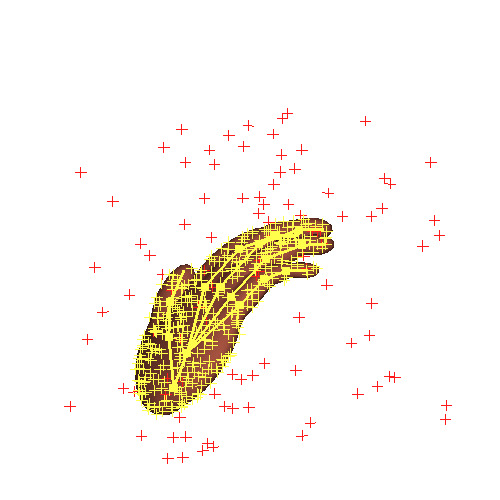}
\includegraphics*[width=0.190\columnwidth,viewport=30 30 400 400,clip]{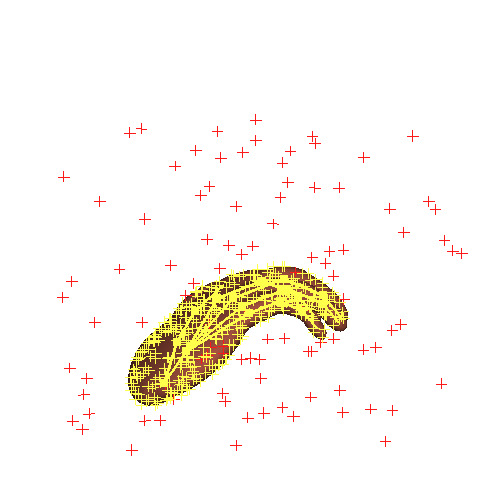}
\includegraphics*[width=0.190\columnwidth,viewport=30 30 400 400,clip]{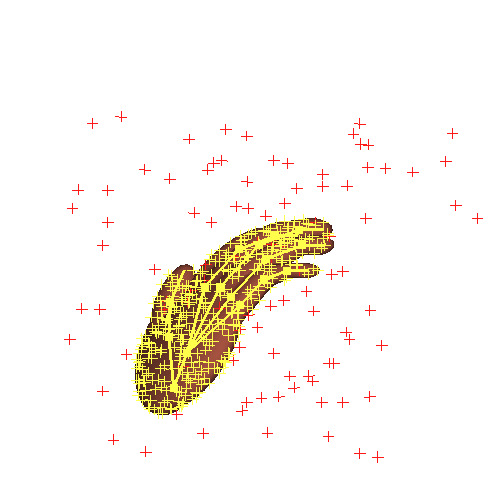}
\includegraphics*[width=0.190\columnwidth,viewport=30 30 400 400,clip]{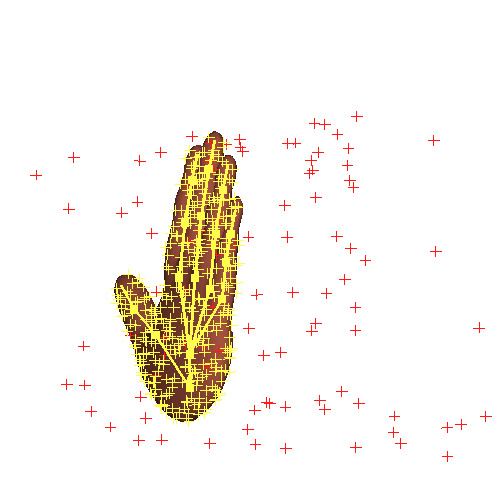}\\
\includegraphics*[width=0.190\columnwidth,viewport=30 30 400 400,clip]{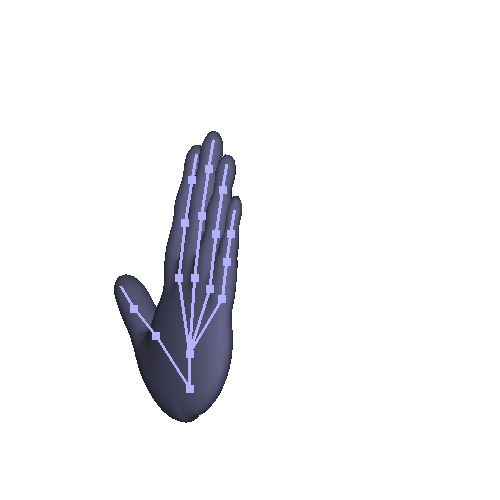}
\includegraphics*[width=0.190\columnwidth,viewport=30 30 400 400,clip]{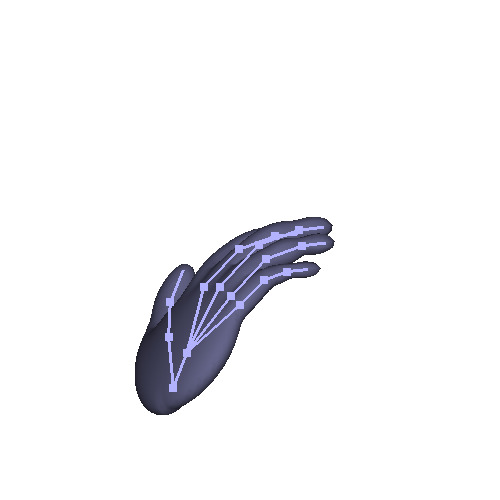}
\includegraphics*[width=0.190\columnwidth,viewport=30 30 400 400,clip]{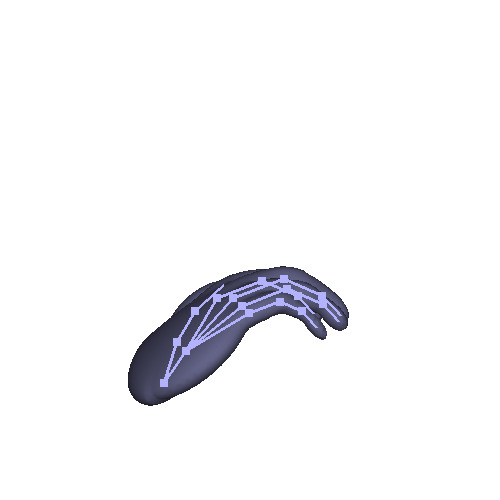}
\includegraphics*[width=0.190\columnwidth,viewport=30 30 400 400,clip]{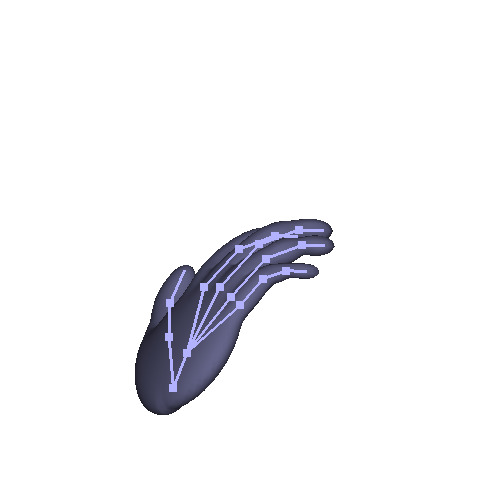}
\includegraphics*[width=0.190\columnwidth,viewport=30 30 400 400,clip]{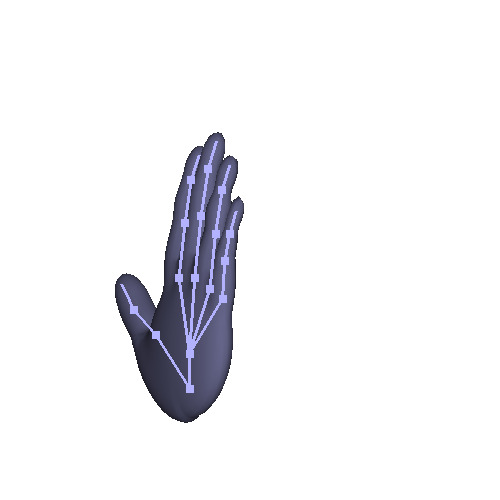}
\caption{Another simulated sequence and the result of ECMPR.}
\label{fig:simulatedhand-3}
\end{figure}

The hand model used in all our
experiments consists in five kinematic chains that share a common
root part -- the palm. Each kinematic chain is composed of four
rigid parts, one part for the palm and three other parts for the
phalanges composing each finger.
Altogether, the kinematic hand model has 16 rigid parts and 21
rotational degrees of freedom (5 rotations for the thumb and 4
rotations for the other fingers). With the additional six degrees
of freedom (three rotations and three translations) associated
with the free motion of the palm, the hand has a total of 27
degrees of freedom. Each hand-part is modeled with an ellipsoid
with fixed dimensions. Model points are obtained by uniformly
sampling the surface of each one of these ellipsoids. This
representation also allows to define an \textit{articulated
implicit surface} over the set of ellipsoids
\cite{PlankersFua2003,DDH04,DDHF06,HNDB09}. Here we only use this
implicit surface representation for visualization purposes.
\begin{figure}[ht!bp]
\centering
\subfigure[Ground-truth parameters]{\includegraphics[width=0.45\columnwidth,viewport=20 0 180 126,clip]{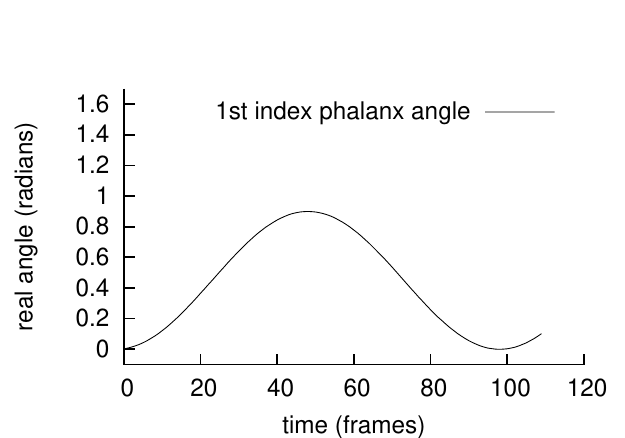}
\includegraphics[width=0.45\columnwidth,viewport=20 0 180 126,clip]{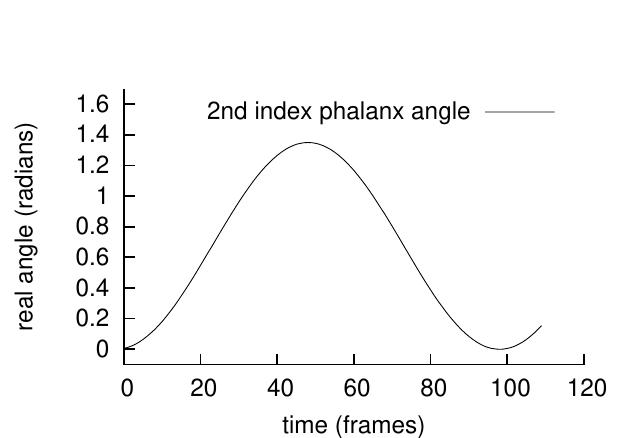}}\\
\subfigure[Correct parameter estimation]{\includegraphics[width=0.45\columnwidth,viewport=20 0 180 126,clip]{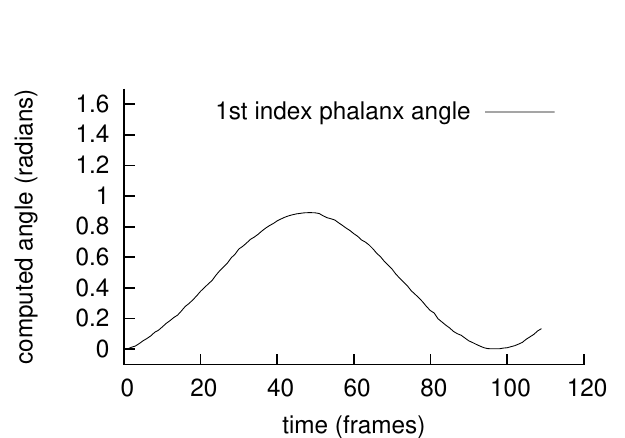}
\includegraphics[width=0.45\columnwidth,viewport=20 0 180 126,clip]{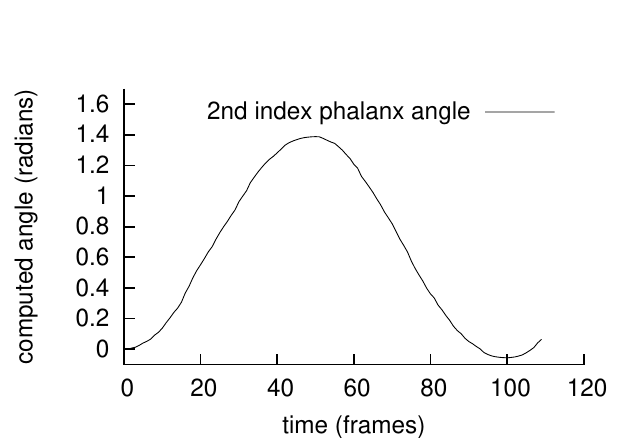}}\\
\subfigure[Incorrect parameter estimation]{\includegraphics[width=0.45\columnwidth,viewport=20 0 180 126,clip]{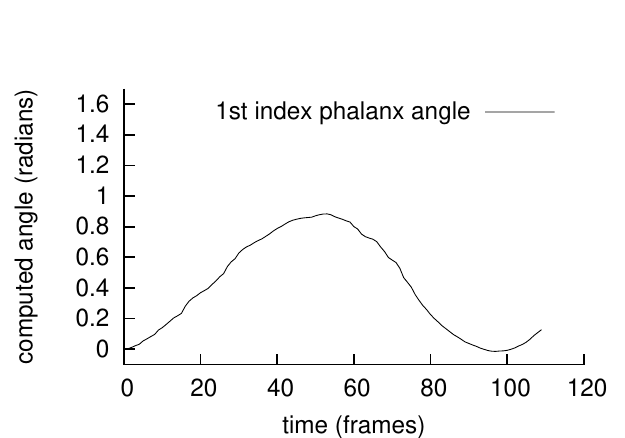}
\includegraphics[width=0.45\columnwidth,viewport=20 0 180 126,clip]{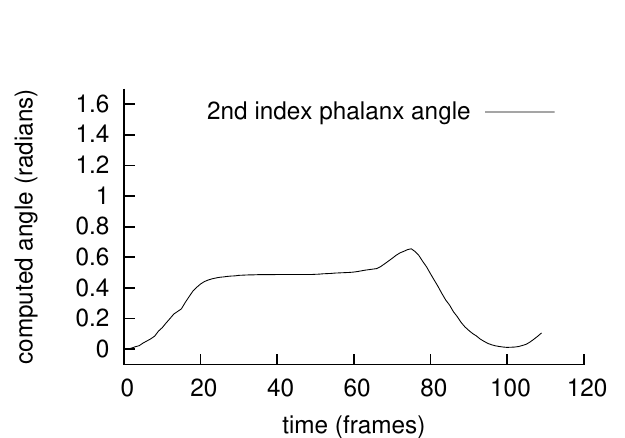}}
\caption{(a) Simulated trajectories of two angular joints associated with
  the first and second phalanges of the index finger. (b) Good
  estimation of these angle values. (c) Bad estimation of the angle
  values due to improper initialisation of the covariance matrix in
  ECMPR. These trajectories correspond to the examples shown in
  Fig.~\ref{fig:simulatedhand-1}.}
\label{figure:res1sim-good}
\end{figure}

A commonly used strategy, in almost every articulated object tracking
algorithm, is to specify joint-limit constraints thus preventing
impossible kinematic poses. It is straightforward to impose such
linear constraints into our convex optimization framework, i.e.,
section~\ref{subsection:anisotropic-covariance}. Indeed, inequality
constraints can be incorporated into
(\ref{eq:minimize-rigid-4}) without affecting the convexity nature of
the problem. In practice we did not implement joint-limit constraints
and hence the solutions found in the examples described below
correspond exactly to (\ref{eq:minimize-rigid-4}).


In the case of simulated data, we animated the hand model just
described in order to
produce realistic articulated motions and to generate sets of
model points, one set for each pose of the model. In practice, all
the experiments described below used 15 model points for each hand
part which corresponds to a total of 240 model points namely
$\vv{X}_i^{(p)}$ with  $1\leq i \leq 15$ and $0 \leq p \leq 15$.

In order to simulate realistic observations we added Gaussian
noise to the surface points. The standard deviation of the noise
was 10\% of the size of the bounding box of the data set. We also
added outliers drawn from a uniform distribution defined over the
volume occupied by the working space of the hand. In all these
simulations the data sets contain 30\% of outliers, i.e., there
are 240 model points, 240 inliers and 72 outliers.

Fig.~\ref{fig:simulatedhand-1}
and Fig.~\ref{fig:simulatedhand-3} show two experiments performed
with simulated hand motions. Each one of these simulated data (top
rows) contains a sequence of 120 articulated poses. We applied our
registration method to these sequences, we estimated the kinematic
parameters, and we compared them with the ground truth.
ECMPR-articulated is applied in parallel to the five kinematic
chains. First, ECMPR-rigid registers the root part (the hand palm)
common to all the chains. Second, ECMPR-rigid is applied to the
first phalanx of the index, middle, ring, and baby fingers. Third,
it is applied to the second phalanx, etc.

\begin{figure}[t!]
\centering
\includegraphics[width=\imgwfour]{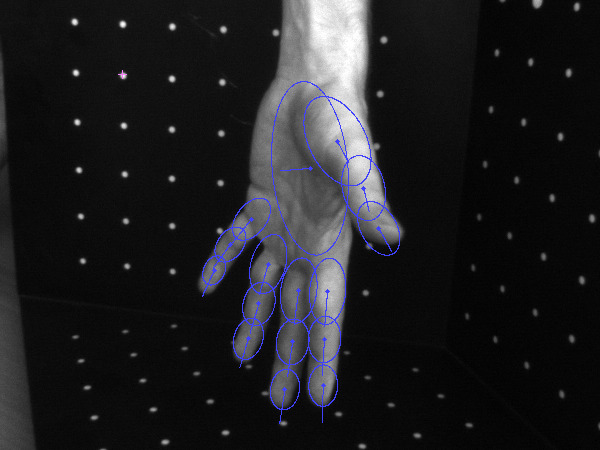}
\includegraphics[width=\imgwfour]{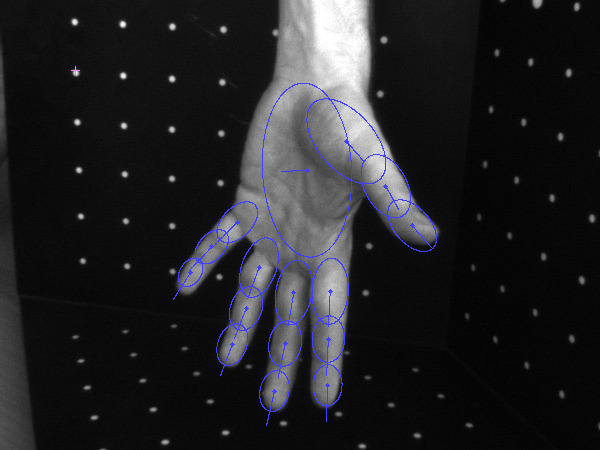}
\includegraphics[width=\imgwfour]{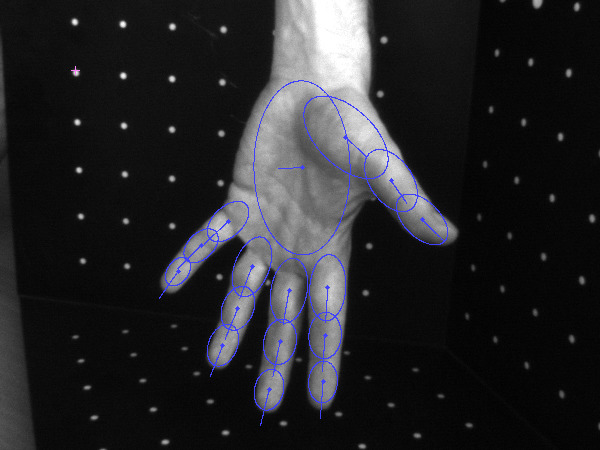}
\includegraphics[width=\imgwfour]{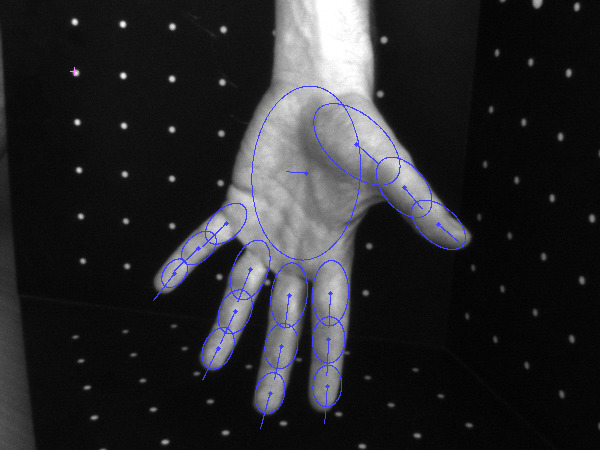}\\
\includegraphics[width=\imgwfour]{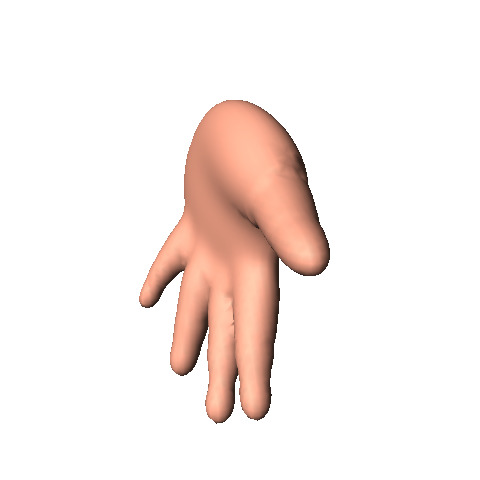}
\includegraphics[width=\imgwfour]{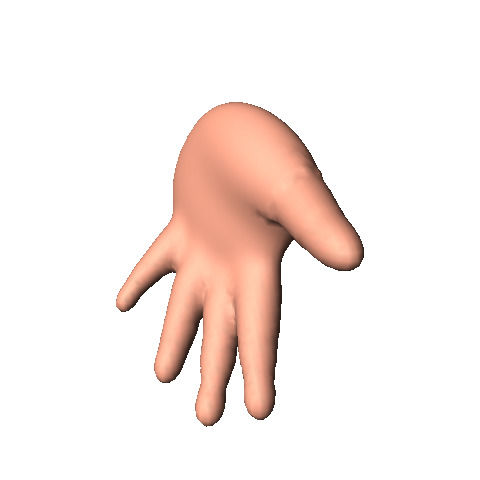}
\includegraphics[width=\imgwfour]{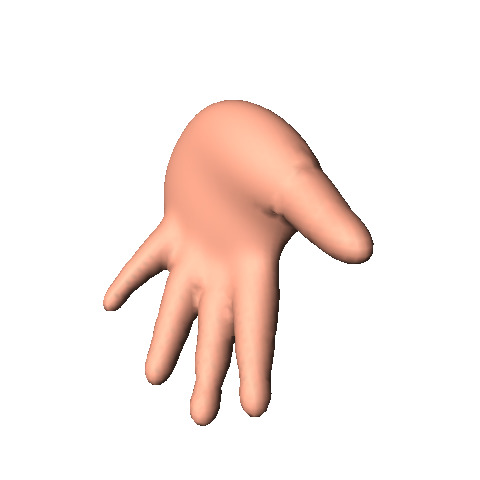}
\includegraphics[width=\imgwfour]{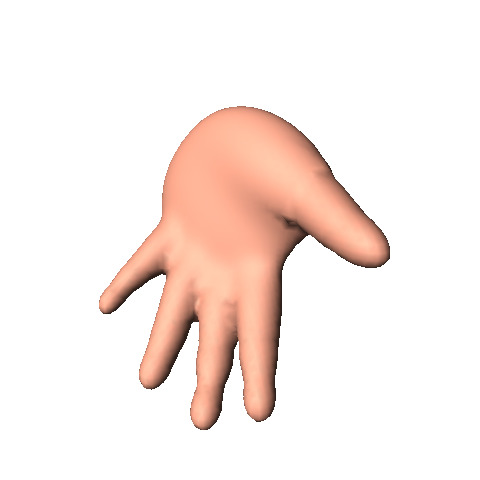}\\
\includegraphics[width=\imgwfour]{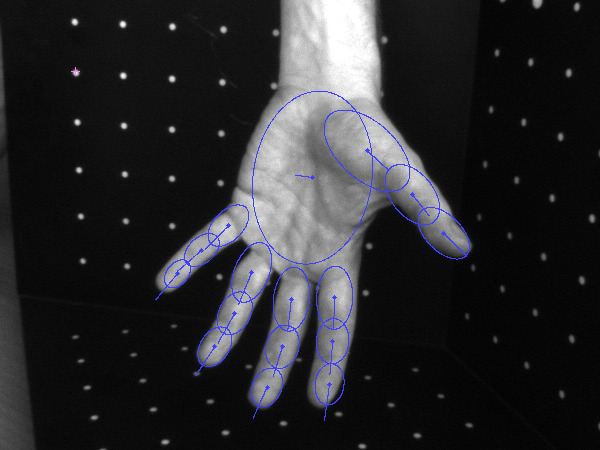}
\includegraphics[width=\imgwfour]{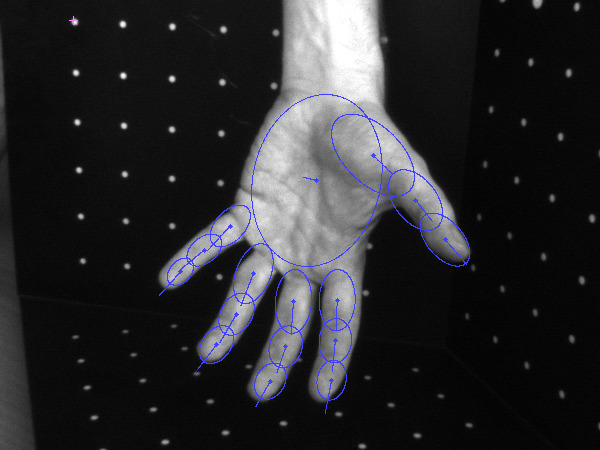}
\includegraphics[width=\imgwfour]{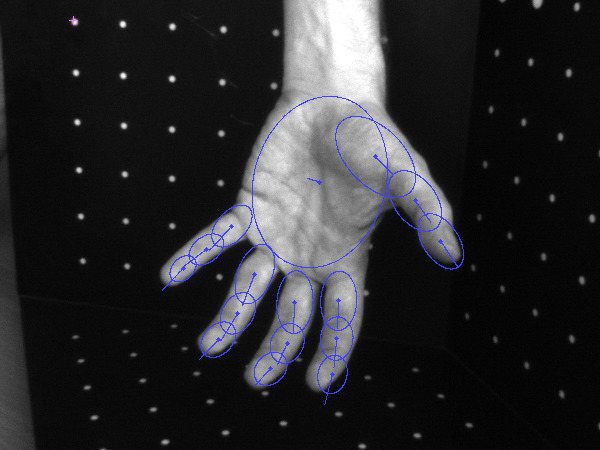}
\includegraphics[width=\imgwfour]{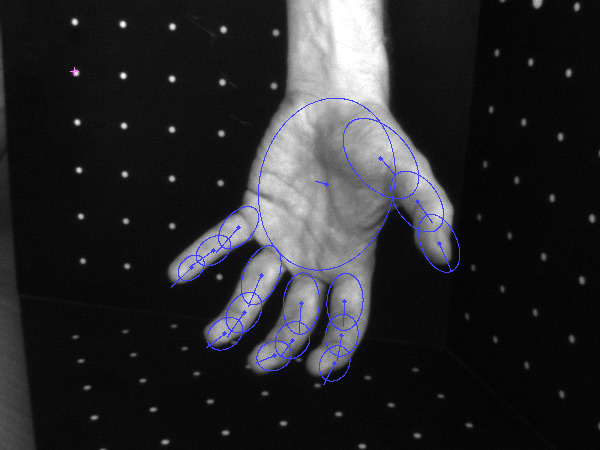}\\
\includegraphics[width=\imgwfour]{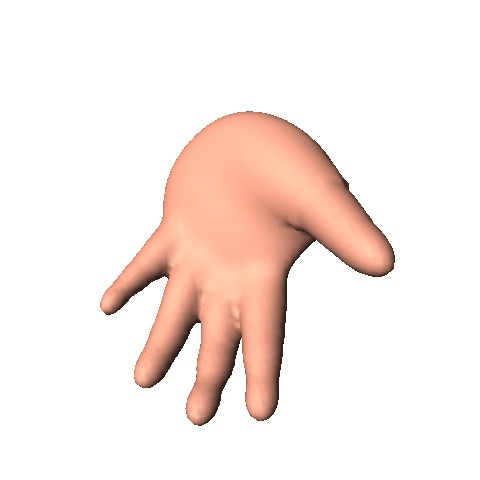}
\includegraphics[width=\imgwfour]{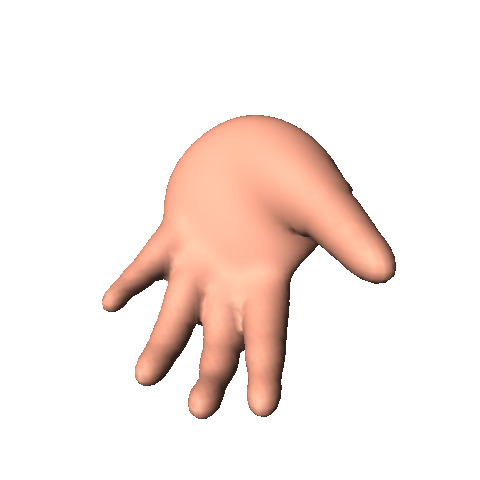}
\includegraphics[width=\imgwfour]{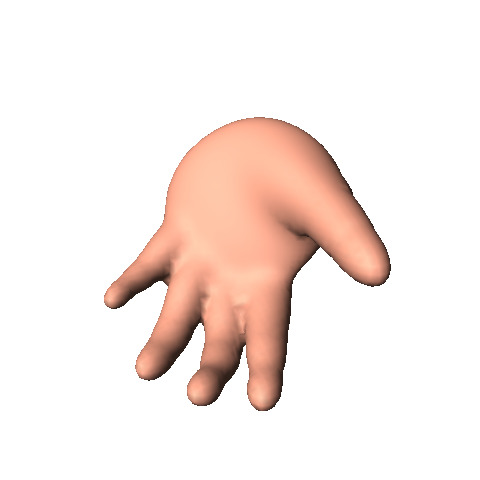}
\includegraphics[width=\imgwfour]{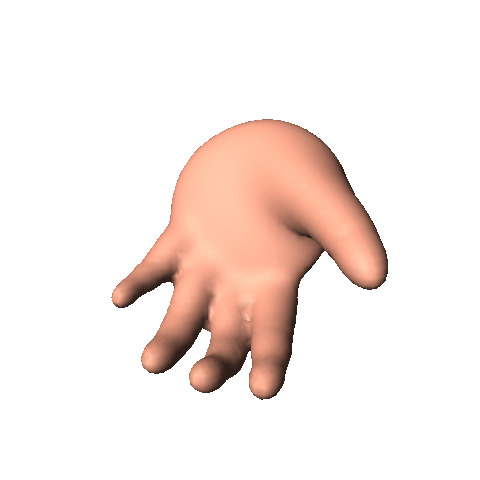}\\
\includegraphics[width=\imgwfour]{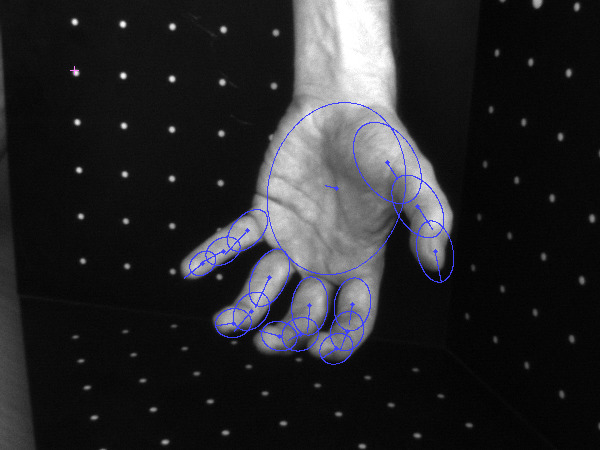}
\includegraphics[width=\imgwfour]{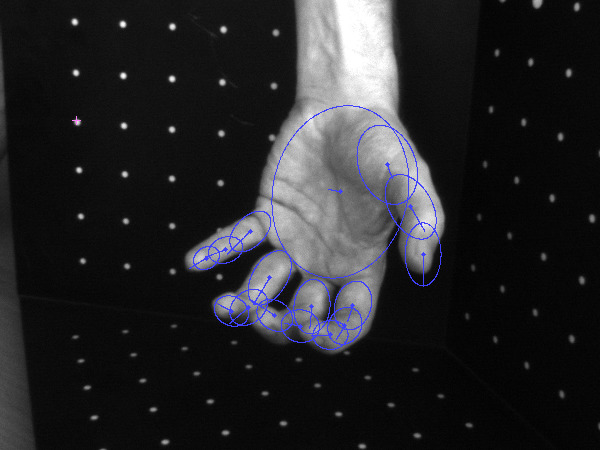}
\includegraphics[width=\imgwfour]{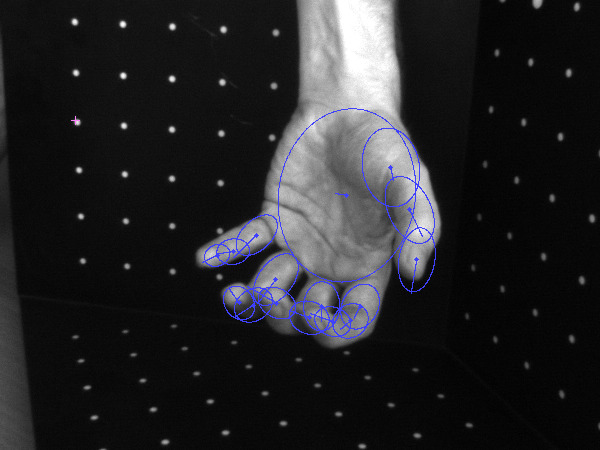}
\includegraphics[width=\imgwfour]{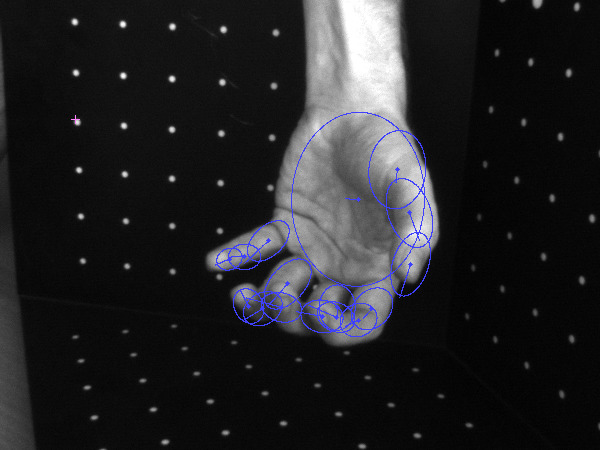}\\
\includegraphics[width=\imgwfour]{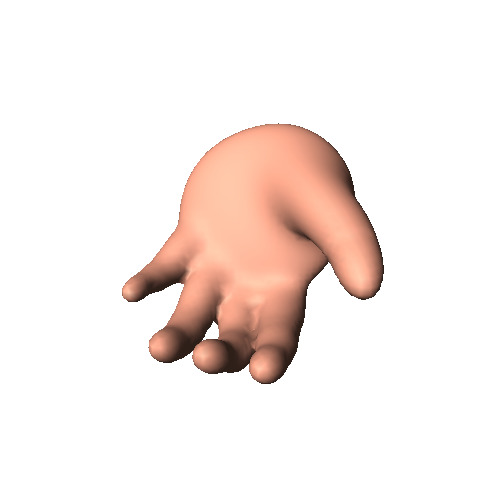}
\includegraphics[width=\imgwfour]{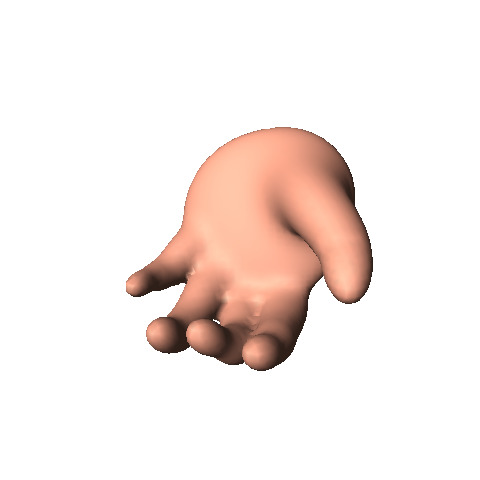}
\includegraphics[width=\imgwfour]{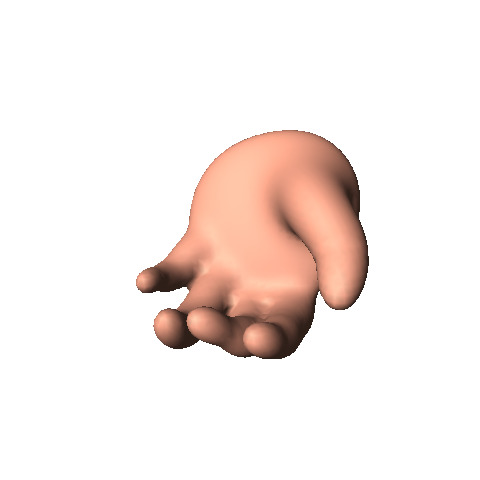}
\includegraphics[width=\imgwfour]{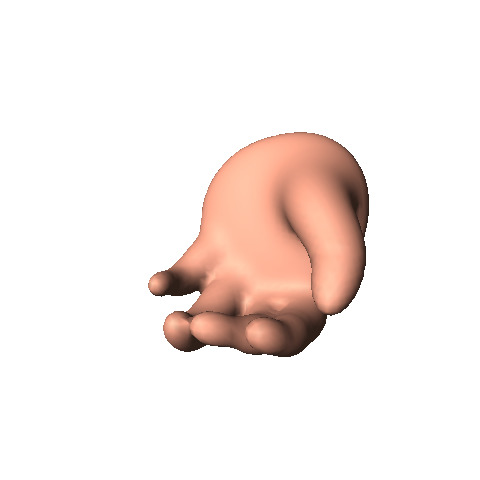}\\
\caption{The image of a hand and the result of tracking for a grasping
movement.}
\label{fig:example-one}
\end{figure}

Fig.~\ref{fig:simulatedhand-1}
shows a sequence of simulated poses (top row) and the results
obtained with our algorithm (middle and bottom rows). When
starting with a large covariance, ECMPR correctly estimated the
articulated poses of the simulated hand (middle row). Starting
with small covariances is equivalent to consider the data points
that are in the neighbourhood of the model points and to disregard
data points that are farther away from the current model point
positions. In this case the trajectory of the thumb has been
correctly estimated but the other four fingers failed to bend
(bottom row). Notice, however, that in both cases the tracker has
been able to ``catch up'' with these finger motions and to reduce
the discrepancy between the estimated trajectories and the ground
truth. The simulated trajectories and the estimated trajectories
of the first and second phalanges of the index finger are shown on
Fig.~\ref{figure:res1sim-good}. Fig.~\ref{fig:simulatedhand-3}
shows another experiment on a different simulated sequence.


These experiments yielded very good results. As expected,
the percentage of outliers barely affected the registration results.
These experiments
confirmed the importance of using an anisotropic covariance model as well
as the fact that covariance initialization is crucial.
All the instances of the ECMPR-rigid algorithm (embedded in
ECMPR-articulated) are
initialized with large spherical covariances. While this increases
the number of EM iterations, it allows the algorithm to escape from
local minima.


\begin{figure}[t!]
\centering
\includegraphics[width=\imgwfour]{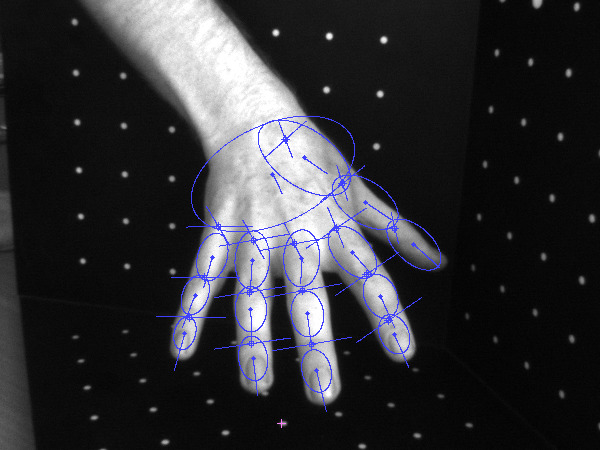}
\includegraphics[width=\imgwfour]{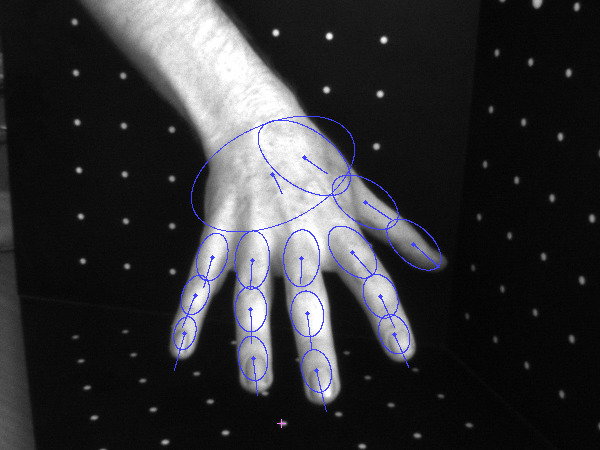}
\includegraphics[width=\imgwfour]{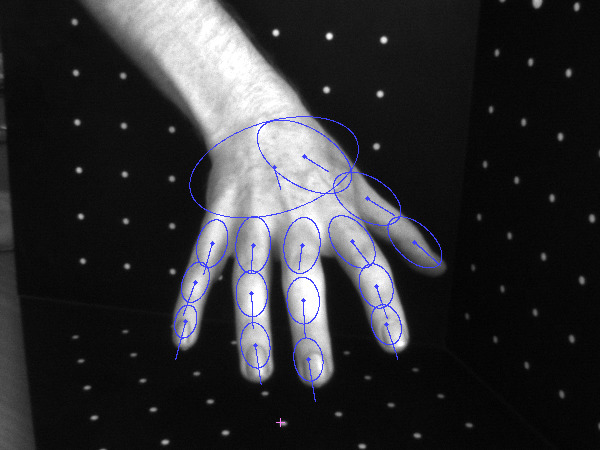}
\includegraphics[width=\imgwfour]{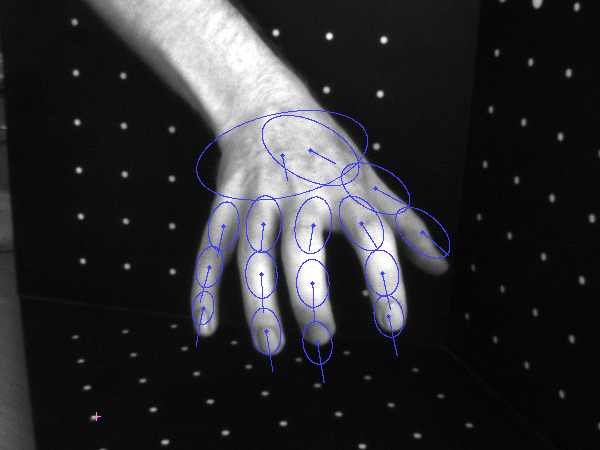}\\
\includegraphics[width=\imgwfour]{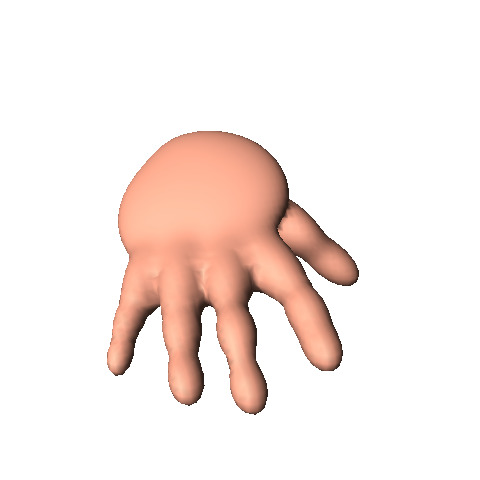}
\includegraphics[width=\imgwfour]{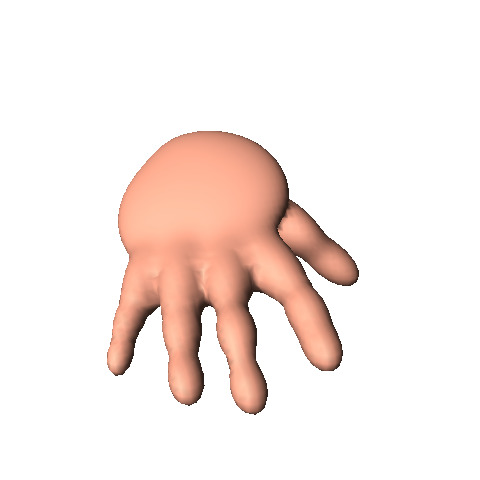}
\includegraphics[width=\imgwfour]{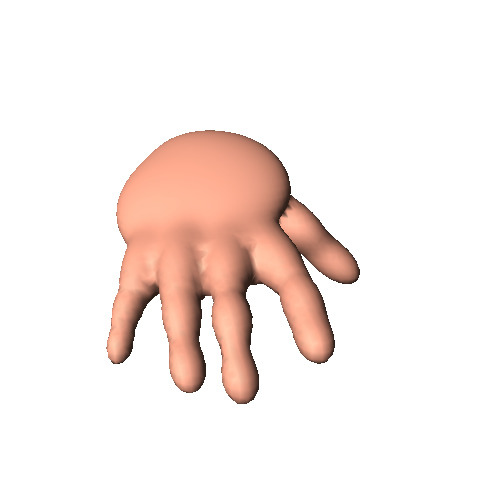}
\includegraphics[width=\imgwfour]{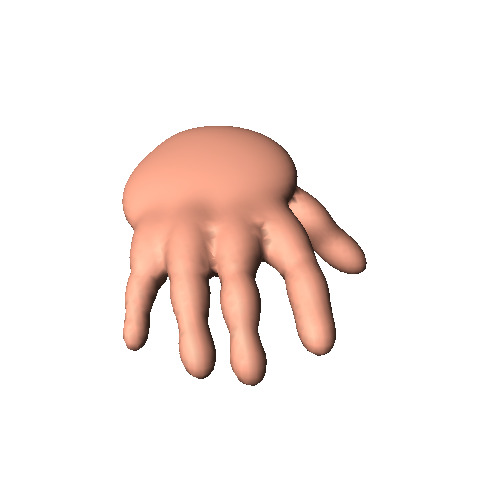}\\
\includegraphics[width=\imgwfour]{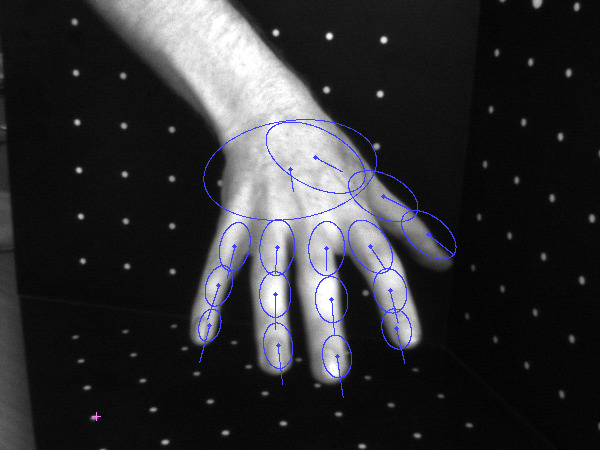}
\includegraphics[width=\imgwfour]{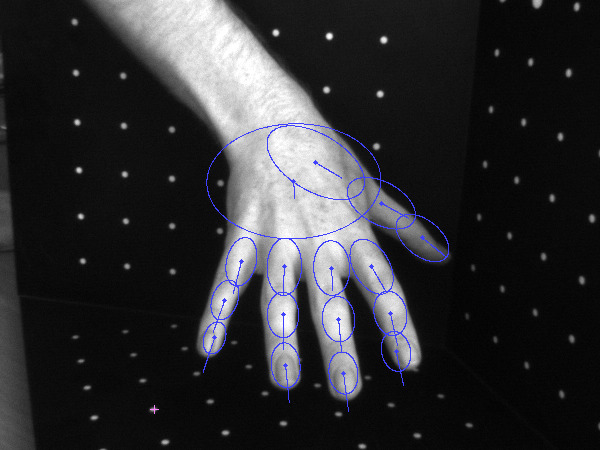}
\includegraphics[width=\imgwfour]{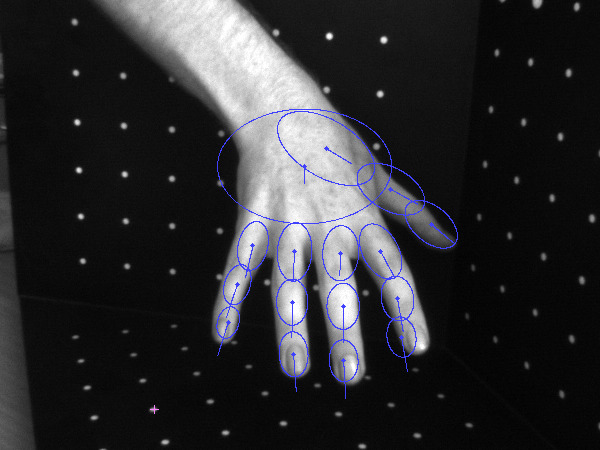}
\includegraphics[width=\imgwfour]{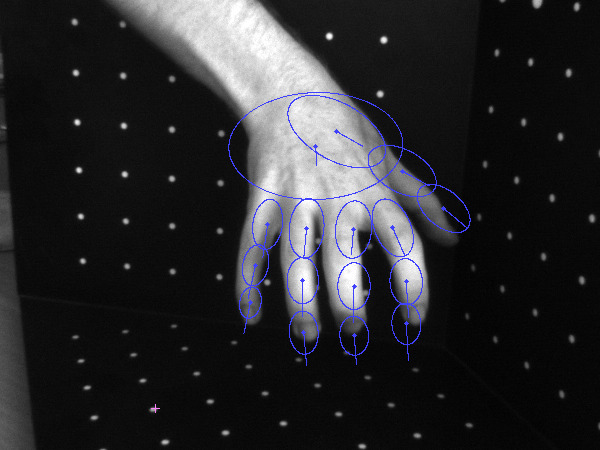}\\
\includegraphics[width=\imgwfour]{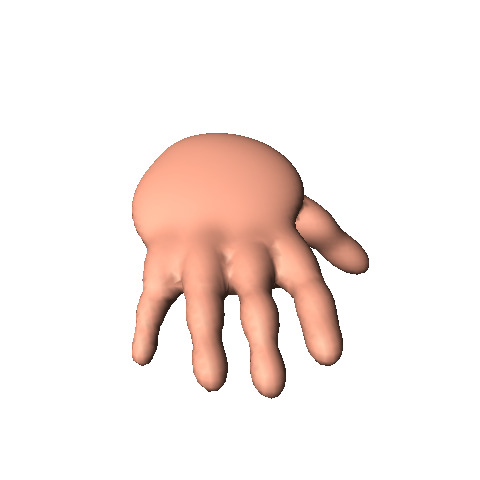}
\includegraphics[width=\imgwfour]{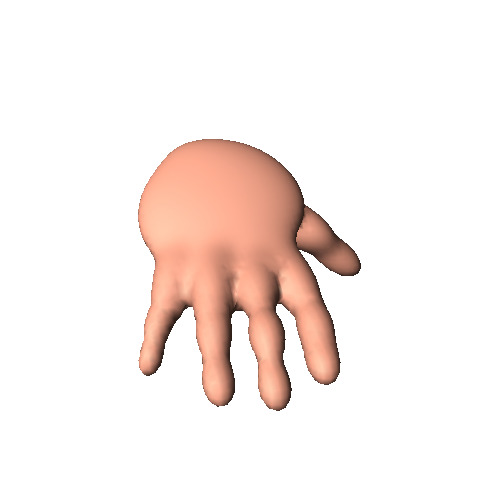}
\includegraphics[width=\imgwfour]{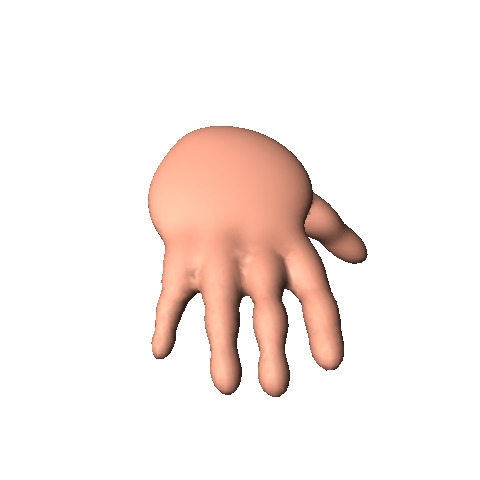}
\includegraphics[width=\imgwfour]{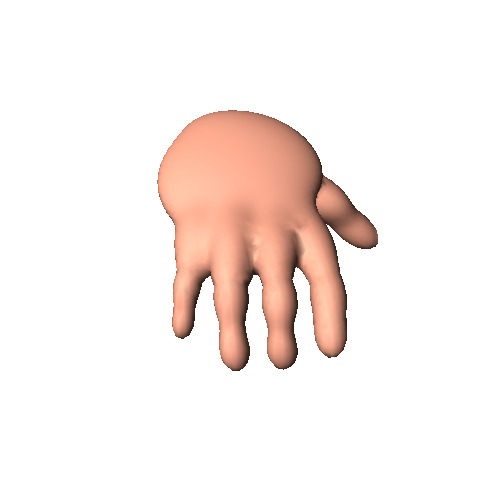}\\
\includegraphics[width=\imgwfour]{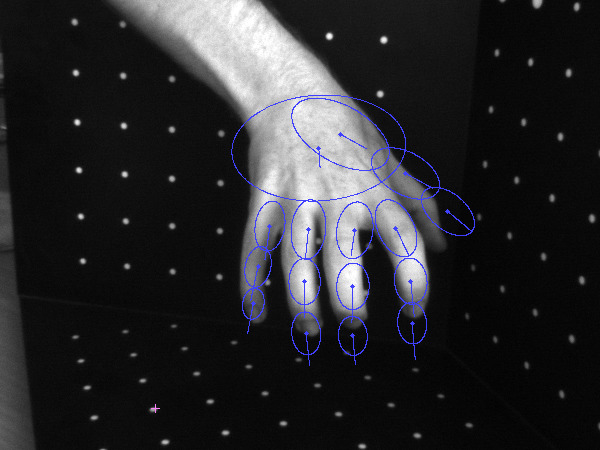}
\includegraphics[width=\imgwfour]{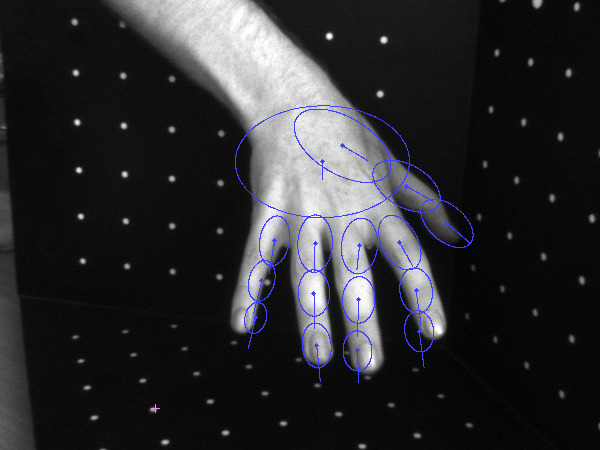}
\includegraphics[width=\imgwfour]{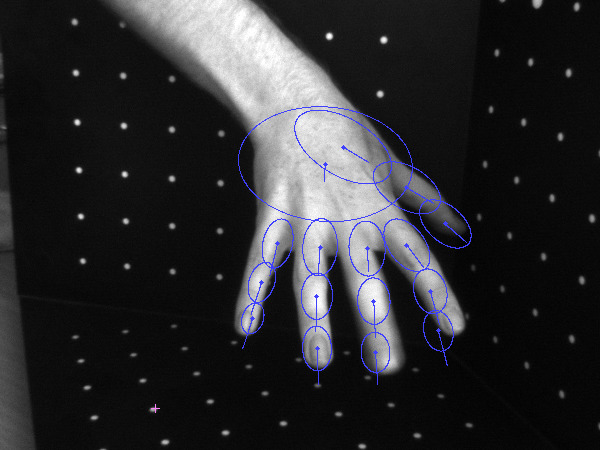}
\includegraphics[width=\imgwfour]{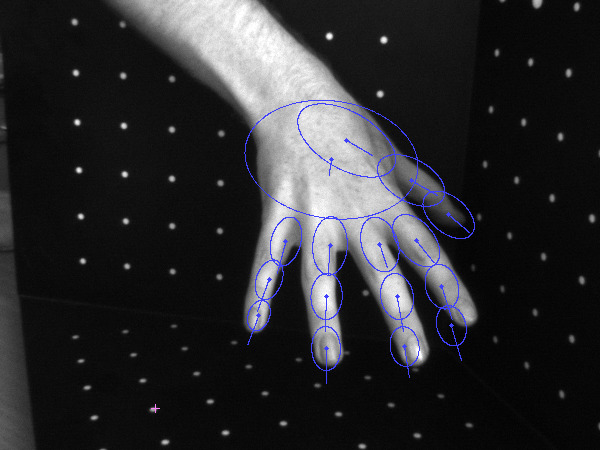}\\
\includegraphics[width=\imgwfour]{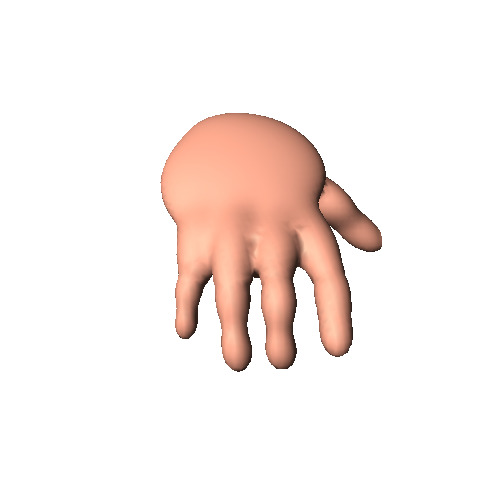}
\includegraphics[width=\imgwfour]{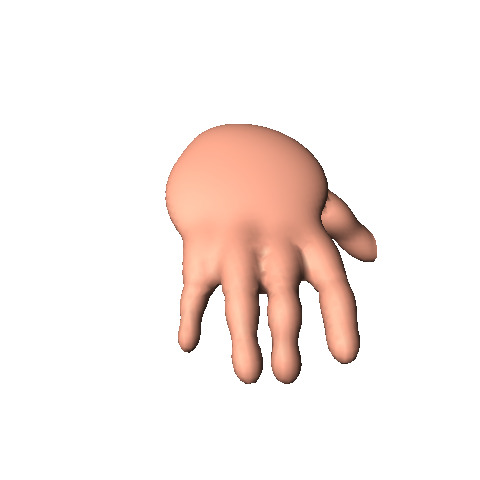}
\includegraphics[width=\imgwfour]{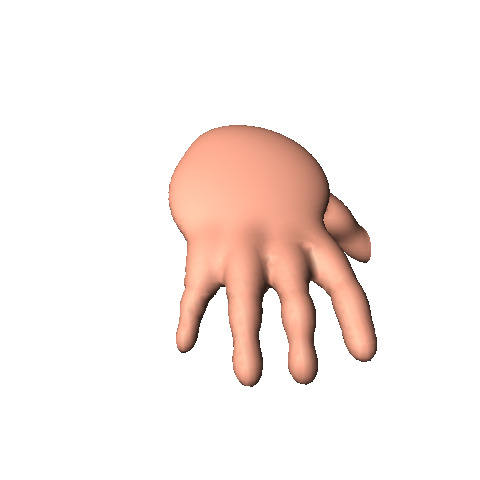}
\includegraphics[width=\imgwfour]{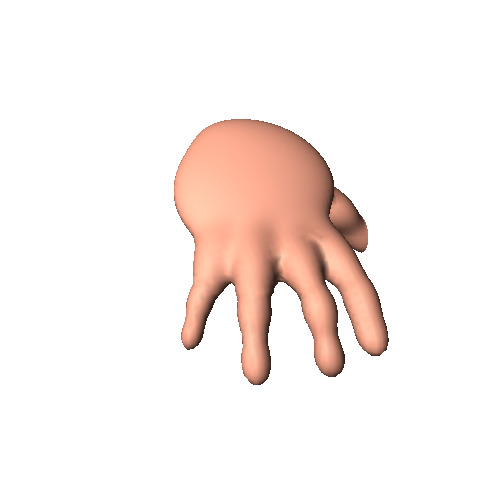}\\
\caption{A similar grasping movement but the hand is viewed from above.}
\label{fig:example-two}
\end{figure}

\begin{figure}[h!]
\centering
\includegraphics[width=\imgwfour]{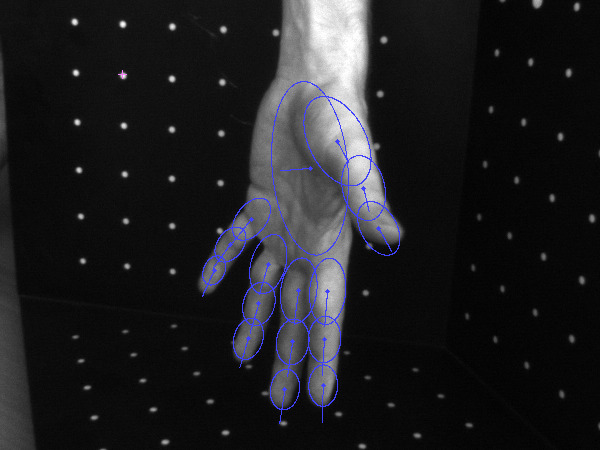}
\includegraphics[width=\imgwfour]{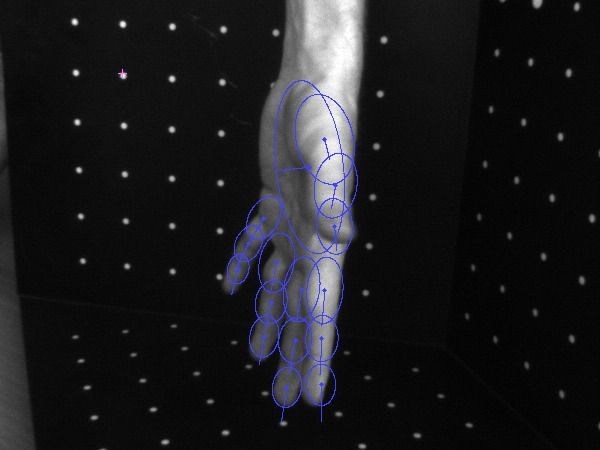}
\includegraphics[width=\imgwfour]{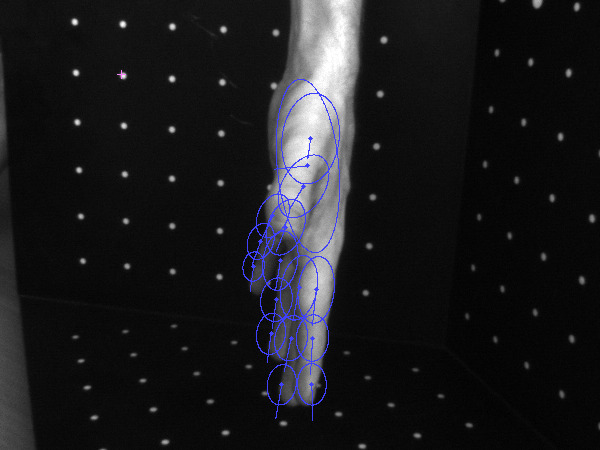}
\includegraphics[width=\imgwfour]{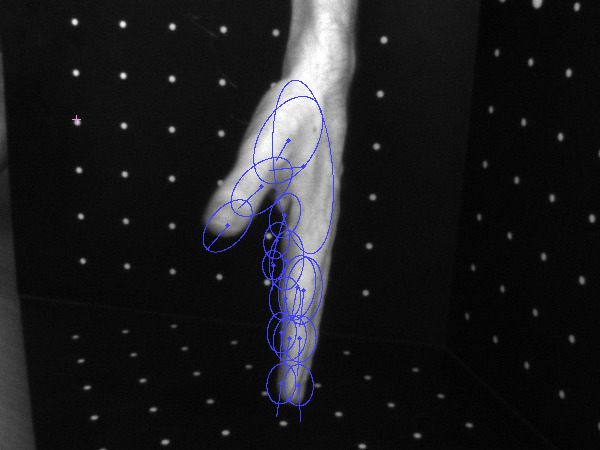}\\
\includegraphics[width=\imgwfour]{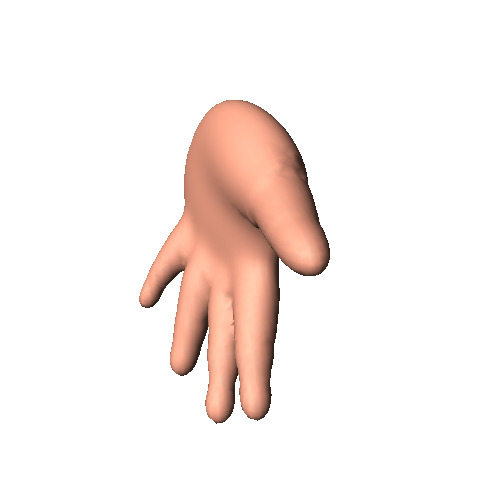}
\includegraphics[width=\imgwfour]{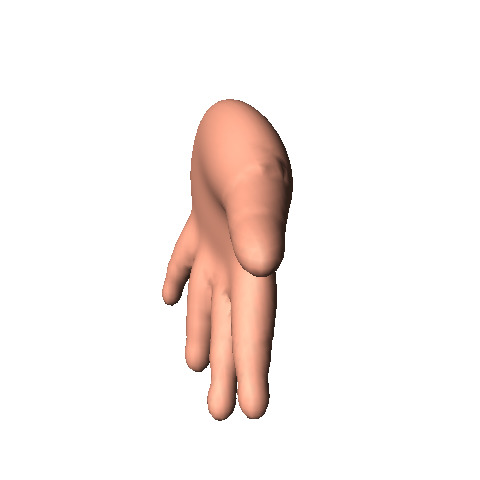}
\includegraphics[width=\imgwfour]{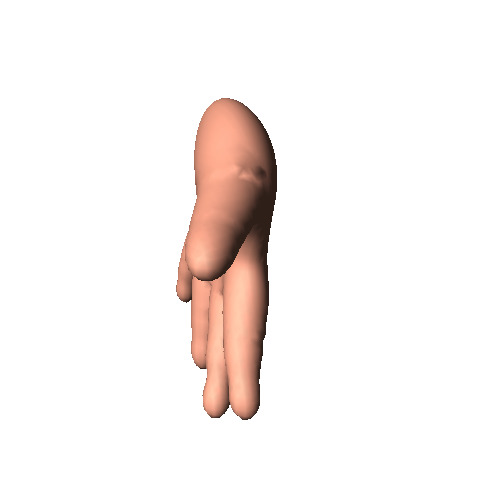}
\includegraphics[width=\imgwfour]{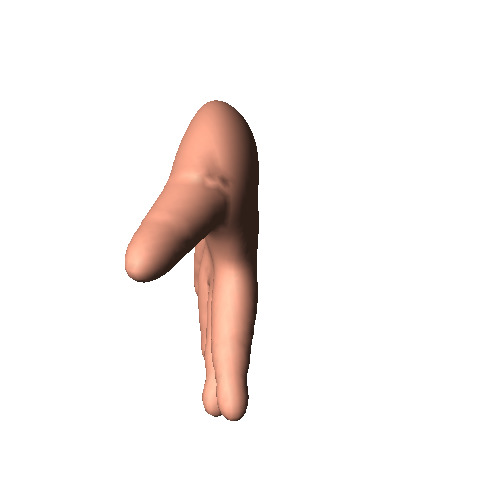}\\
\includegraphics[width=\imgwfour]{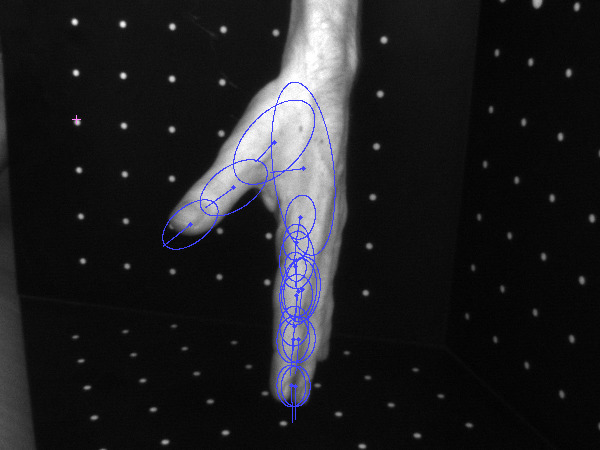}
\includegraphics[width=\imgwfour]{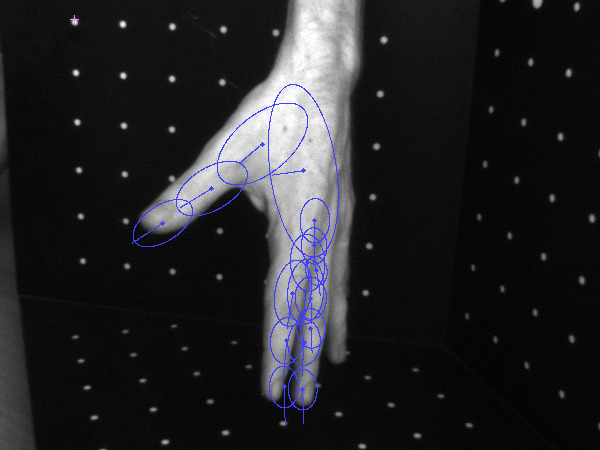}
\includegraphics[width=\imgwfour]{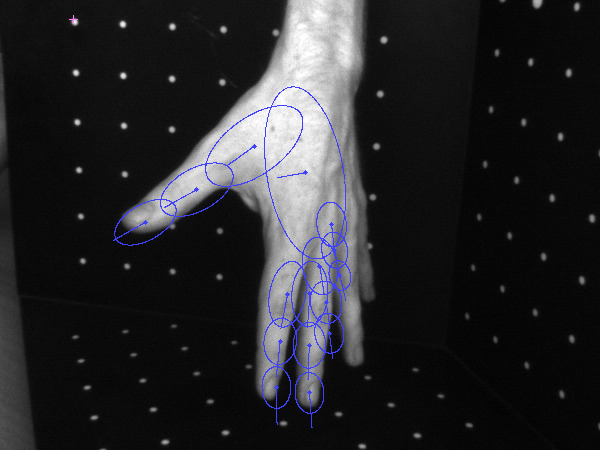}
\includegraphics[width=\imgwfour]{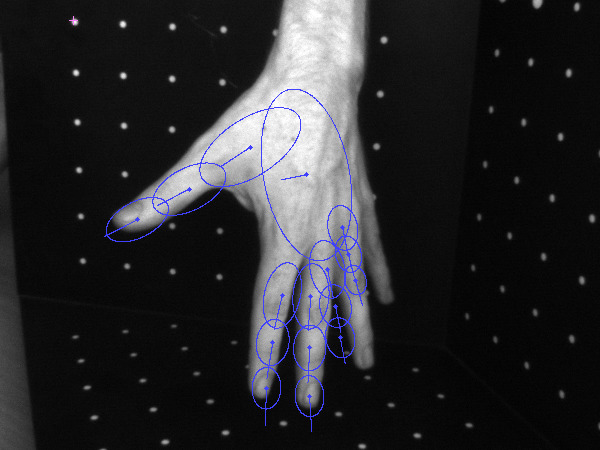}\\
\includegraphics[width=\imgwfour]{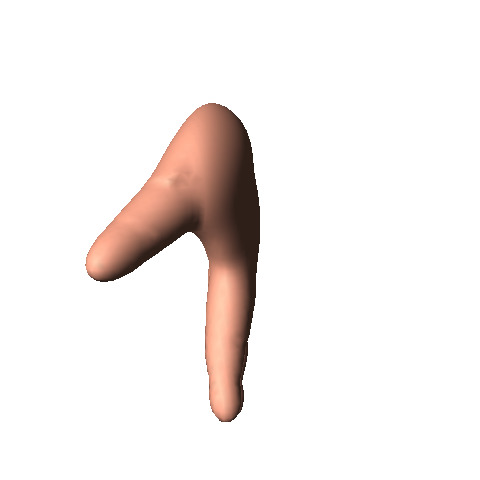}
\includegraphics[width=\imgwfour]{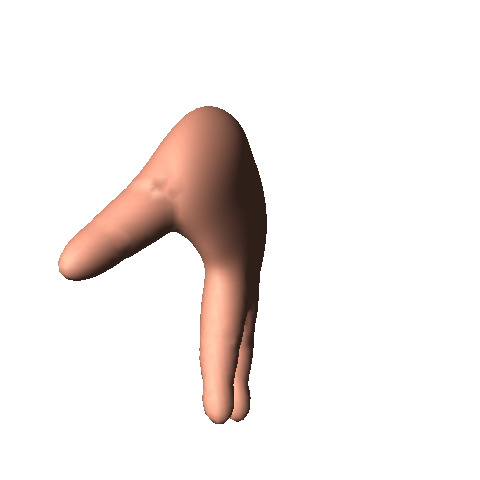}
\includegraphics[width=\imgwfour]{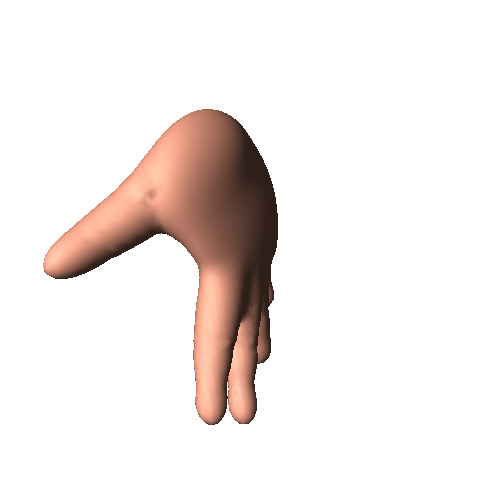}
\includegraphics[width=\imgwfour]{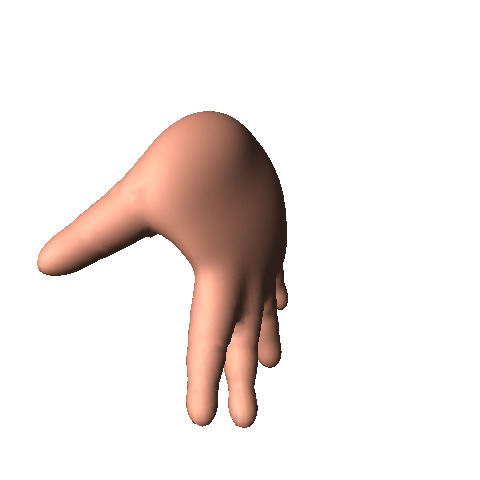}\\
\includegraphics[width=\imgwfour]{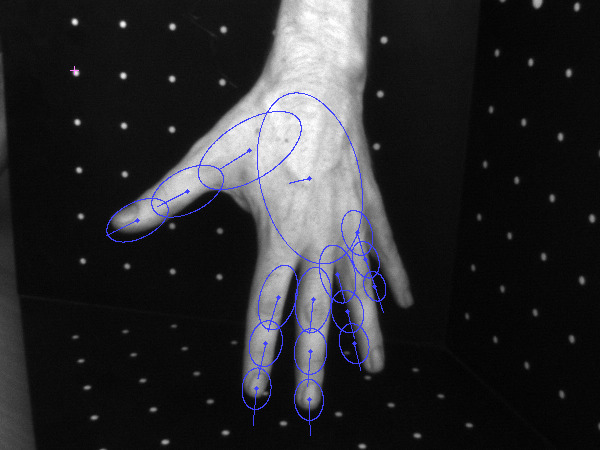}
\includegraphics[width=\imgwfour]{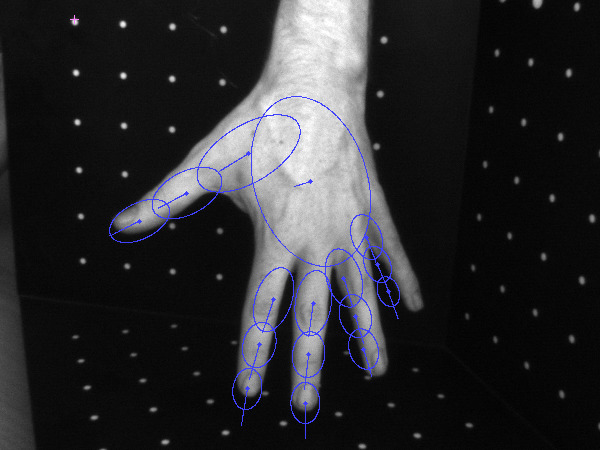}
\includegraphics[width=\imgwfour]{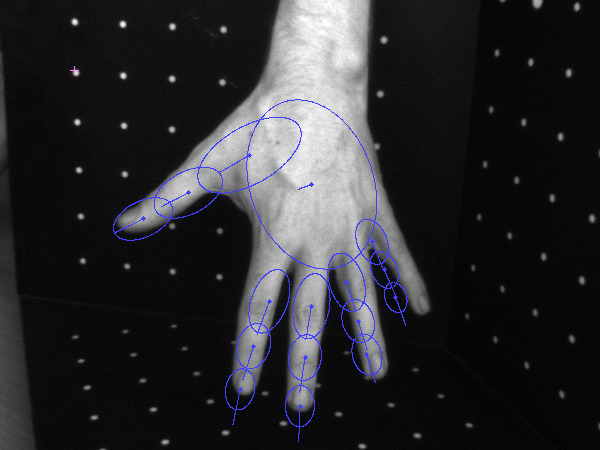}
\includegraphics[width=\imgwfour]{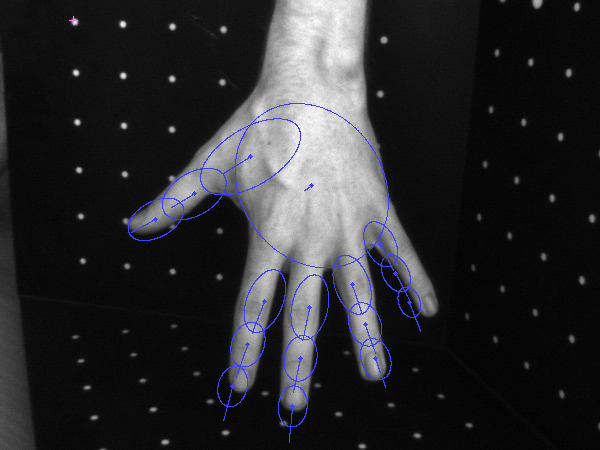}\\
\includegraphics[width=\imgwfour]{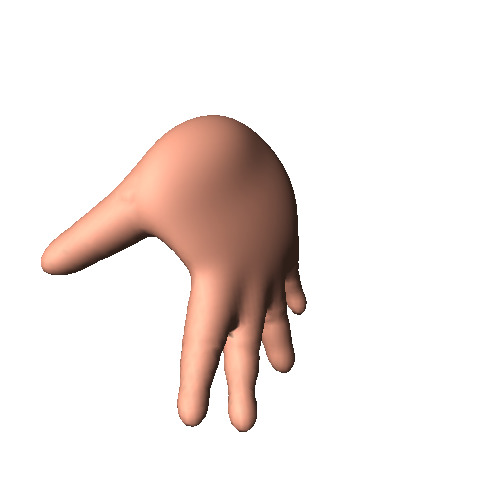}
\includegraphics[width=\imgwfour]{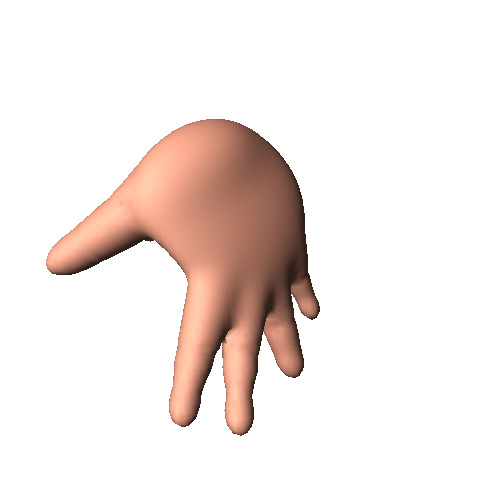}
\includegraphics[width=\imgwfour]{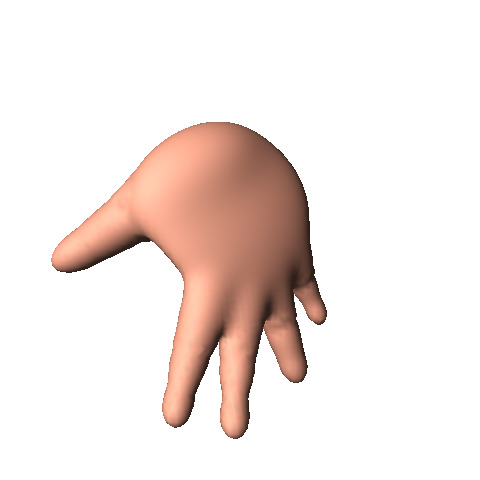}
\includegraphics[width=\imgwfour]{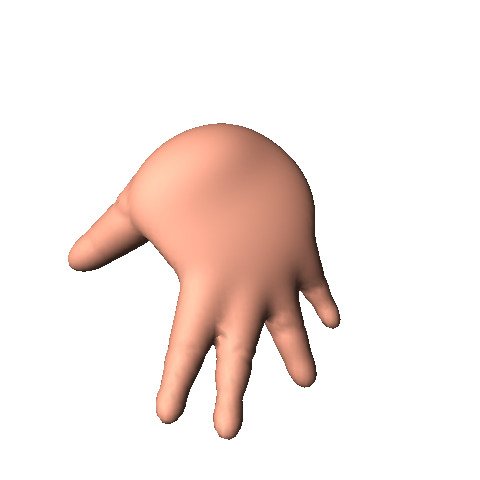}\\
\caption{A rotational movement of the hand around an axis parallel to
  the image plane causes the fingers to disappear
  from the left-hand side of the image and appear again onto the right-hand
  side. These occlusions have as a result a very coarse initialization
  of the current pose. In spite of this problem the tracker performs
  quite well due to re-initialization of the covariance matrix at each
  time step of the tracker.}
\label{fig:example-three}
\end{figure}

We then tested our method with real data consisting in several
hand motions observed with a stereoscopic camera system,
Fig~\ref{fig:stereo-pair-and-result}. Each data sequence that we
used contains 100 image pairs gathered at 20 frames per second. We
run a standard stereo algorithm to estimate 3-D points. This
yielded 500 to 1000 reconstructed points at each time step. The
noise associated with these stereo data is inherently anisotropic
because of the inaccuracy in depth. Moreover, there are many
outliers that correspond either to data points which do not lie on
the hand or to stereo mismatches.

The results of applying ECMPR to these data sets are illustrated
on Figs.~\ref{fig:example-one}, \ref{fig:example-two} and \ref{fig:example-three}. In the
first and second examples the hand performs a grasping movement. In the third
example the hand rotates around an axis roughly parallel
to the image plane. In all these cases the algorithm selected, on
average, 250 inliers per frame; This number roughly corresponds to
the number of model points being considered (240). All the other
data points were assigned to the outlier class. Notice that the
number of data points vary a lot (500 to 1000 observations at each
frame) and that the outlier rejection mechanism that we propose in
this paper does not need to know in advance the percentage of
outliers.

Note that along these motion sequences the hand flips from one
side to another side while the positions and orientations of the
fingers vary considerably. This means that it is often the case
that almost all the model points that were currently registered,
may suddenly disappear while other model points suddenly appear. This is
one of the main difficulties associated with registering
articulated objects. Therefore, during the tracking, the algorithm
must perform some form of bootstrapping, i.e., it must establish
data-point-to-model-point assignments \textit{from scratch}.
Re-initialization of the covariance matrix at each time step,
along the lines described above, is crucial to the success of the
registration/tracking algorithm.

\section{Conclusions}

In this paper we addressed the problem of matching rigid and
articulated shapes through robust point registration. The proposed
approach has its roots in model-based clustering \cite{FraleyRaftery2002}. More
specifically, the point registration problem is cast into the
framework of maximum likelihood with hidden variables \cite{DempsterLairdRubin77,RednerWalker84,McLachlanKrishnan97}. We formally derived a
variant of the EM algorithm which \textit{maximizes} the
\textit{expected} complete-data log-likelihood. This guarantees
maximization of the observed-data log-likelihood. We
showed that it is convenient to replace the standard M-step by three
conditional maximization steps, or CM-steps, while preserving the
convergence properties of EM.

Our approach
differs significantly from existing methods for point registration,
namely ICP and its variants \cite{BeslMcKay92,Zhang1994,RusinkiewiczLevoy2001,Fitzgibbon2003,ChetverikovStepanovKrsek2005,SharpLeeWehe2008,Demirdjian2004,MundermanCorazzaAndriacchi2007}, soft assignment
methods  \cite{RCB:IPMI97,CR:CVIU03,Liu2005}, as well as
various EM implementations \cite{Wells97,ChuiRangarajan2000,GrangerPennec2002,MyronenkoSongCarreira-Perpinan2006,SofkaYangStewart2007,JianVemuri2005}:
The ECMPR-rigid and -articulated algorithms that we proposed fit a set
of model points to a set of data points where each model point is the
center of a Gaussian component in a mixture model. Each component in
the mixture may have its own anisotropic covariance. Our method treats
the data
points and the model points in a non-symmetric way, which has
several advantages: It allows to deal with a varying number of
observations, either larger or smaller than the number of model points, it
performs robust parameter estimation in the presence of data corrupted with noise and
outliers, and it is based on a principled probabilistic approach.

More specifically, the method guarantees robustness via a uniform component added to the
Gaussian mixture model. This built-in outlier rejection mechanism differs
from existing outliers detection/rejection strategies used in conjunction
with point registration, such as methods based on non-linear loss
functions that can be trapped in local minima, or methods based on
random sampling which are time-consuming and that can only deal with a
limited number of outlying data.

In particular we put emphasis on a general model that uses anistropic
covariance matrices, in which case the
rotation associated with rigid alignment cannot be found in
closed-form. This led us to approximate the associated non-convex optimization
problem with a convex one. Namely, we showed how to transform the
non-linear problem into a constrained quadratic optimization one and
how to use semi-definite positive relaxation to solve it in practice.

We provided in detail the ECMPR-rigid algorithm. We showed how this
algorithm can be incrementally applied to articulated registration
using a novel kinematic representation that is well suited in the case
of point registration.

In general, ECMPR performs better than ICP. In particular it is less sensitive to initialization and it is more robust to outliers. In the future we plan to investigate various ways of implementing our algorithm more efficiently. Promising approaches are based on modifying the standard E-step.  A fast but suboptimal ``winner take all" variant is \textit{Classification EM}, or CEM, which consists in forcing the posterior probabilities to either 0 or 1 after each E-step \cite{CeleuxGovaert92}. We plan to study CEM in the particular context of point registration and, possibly, derive a more efficient implementation of ECMPR.
This may also lead to a probabilistic interpretation of ICP, and hence to a better understanding of the links existing between probabilistic and deterministic registration methods. Other efficient variants of the E-step are based on structuring the data using either block-like organizations \cite{ThiessonMeekHeckerman2001}, or  KD-trees \cite{VerbeekNunninkVlassis2006}. We also plan to implement KD-trees in order to increase the efficiency of ECMPR.



\appendix[Expansion of $\mm{A}$ and $\vv{b}$ in
eq. (\ref{eq:minimize-rigid-2})]
\label{app:expansion-of-Ab}

By expanding (\ref{eq:minimize-rigid-1}),
substituting the optimal
translation with
(\ref{eq:optimal-t})
and rearranging terms, one
obtains the following expressions for the 9$\times$9 matrix  $\mm{A}$
and the 9$\times$1 vector $\vv{b}$:
\begin{equation}
\mm{A} = \mm{N} - \mm{M}\tp\mm{K}\mm{M}
\end{equation}
\begin{equation}
\vv{b} =  \mm{M}\tp \vv{p} - \vv{q}
\end{equation}
with:
\begin{eqnarray*}
\mm{N}_{9\times 9} &=& \sum_{i=1}^{n} \lambda_i \vv{X}_i \vv{X}_i\tp \otimes
\mathbf{\Sigma}_i\inverse \\
\mm{M}_{3\times 9} &=& \sum_{i=1}^{n} \lambda_i \vv{X}_i\tp \otimes
\mathbf{\Sigma}_i\inverse \\
\mm{K}_{3\times 3} &=& \left( \sum_{i=1}^{n} \lambda_i \mathbf{\Sigma}_i\inverse
\right)\inverse \\
\vv{p}_{3\times 1} &=& \mm{K} \left( \sum_{i=1}^{n} \lambda_i
  \mathbf{\Sigma}_i\inverse \vv{W}_i \right) \\
\vv{q}_{9\times 1}&=& \text{vec}\left( \sum_{i=1}^{n} \lambda_i
  \mathbf{\Sigma}_i\inverse \vv{W}_i \vv{X}_i\tp \right)
\end{eqnarray*}
The Kronecker product between the $m\times n$ matrix/vector $\mm{A}$ and the
$p\times q$ matrix/vector $\mm{B}$ is the $mp\times nq$ matrix/vector defined by:
\[
\mm{A} \otimes \mm{B} = \left[
\begin{array}{ccc}
A_{11} \mm{B} & \ldots & A_{1n}  \mm{B} \\
\vdots        &        & \vdots \\
A_{m1} \mm{B} & \ldots & A_{mn}  \mm{B}
\end{array}
\right]
\]
Moreover, $\text{vec}(\mm{A})$ returns the $mn\times 1$ vector:
\[ \text{vec}(\mm{A}) = (A_{11} \ldots A_{mn})\tp
\]

\bibliographystyle{IEEEtran}

\begin{thebibliography}{10}
\providecommand{\url}[1]{#1}
\csname url@samestyle\endcsname
\providecommand{\newblock}{\relax}
\providecommand{\bibinfo}[2]{#2}
\providecommand{\BIBentrySTDinterwordspacing}{\spaceskip=0pt\relax}
\providecommand{\BIBentryALTinterwordstretchfactor}{4}
\providecommand{\BIBentryALTinterwordspacing}{\spaceskip=\fontdimen2\font plus
\BIBentryALTinterwordstretchfactor\fontdimen3\font minus
  \fontdimen4\font\relax}
\providecommand{\BIBforeignlanguage}[2]{{%
\expandafter\ifx\csname l@#1\endcsname\relax
\typeout{** WARNING: IEEEtran.bst: No hyphenation pattern has been}%
\typeout{** loaded for the language `#1'. Using the pattern for}%
\typeout{** the default language instead.}%
\else
\language=\csname l@#1\endcsname
\fi
#2}}
\providecommand{\BIBdecl}{\relax}
\BIBdecl

\bibitem{BeslMcKay92}
P.~J. Besl and N.~D. McKay, ``A method for registration of 3-{D} shapes,''
  \emph{IEEE Transactions on Pattern Analysis and Machine Intelligence},
  vol.~14, pp. 239--256, February 1992.

\bibitem{Zhang1994}
Z.~Zhang, ``Iterative point matching for registration of free-form curves and
  surfaces,'' \emph{International Journal of Computer Vision}, vol.~13, pp.
  119--152, 1994.

\bibitem{RusinkiewiczLevoy2001}
S.~Rusinkiewicz and M.~Levoy, ``Efficient variants of the {ICP} algorithm,'' in
  \emph{IEEE Proc. 3rd International Conference on 3D Digital Imaging and
  Modeling}, Quebec, Canada, May-June 2001.

\bibitem{Fitzgibbon2003}
A.~W. Fitzgibbon, ``Robust registration of {2D} and {3D} point sets,''
  \emph{Image and Vision Computing}, vol.~21, no.~12, pp. 1145--1153, December
  2001.

\bibitem{ChetverikovStepanovKrsek2005}
D.~Chetverikov, D.~Stepanov, and P.~Krsek, ``{Robust Euclidean alignment of 3D
  point sets: the trimmed iterative closest point algorithm},'' \emph{Image and
  Vision Computing}, vol.~23, no.~3, pp. 299--309, March 2005.

\bibitem{SharpLeeWehe2008}
G.~C. Sharp, S.~W. Lee, and D.~K. Wehe, ``Maximum-likelihood registration of
  range images with missing data,'' \emph{IEEE Transactions on Pattern Analysis
  and Machine Intelligence}, vol.~30, no.~1, pp. 120--130, January 2008.

\bibitem{Demirdjian2004}
D.~Demirdjian, ``Combining geometric- and view-based approaches for articulated
  pose estimation,'' in \emph{European Conference on Computer Vision}, vol.
  III, 2004, pp. 183--194.

\bibitem{MundermanCorazzaAndriacchi2007}
L.~Munderman, S.~Corazza, and T.~P. Andriacchi, ``Accurately measuring human
  movement using articulated {ICP} with soft-joint constraints and a repository
  of articulated models,'' in \emph{IEEE Proc. of the Eleventh International
  Conference on Computer Vision}, Rio de Janeiro, Brazil, November 2007.

\bibitem{RCB:IPMI97}
A.~Rangarajan, H.~Chui, and F.~L. Bookstein, ``The softassign procrustes
  matching algorithm,'' in \emph{Information Processing in Medical Imaging
  {(IPMI)}}, 1997, pp. 29--42.

\bibitem{CR:CVIU03}
H.~Chui and A.~Rangarajan, ``A new point matching algorithm for non-rigid
  registration,'' \emph{Computer Vision and Image Understanding}, vol.~89, no.
  2-3, pp. 114--141, February 2003.

\bibitem{Liu2005}
Y.~Liu, ``Automatic {3D} free form shape matching using the graduated
  assignment algorithm,'' \emph{Pattern Recognition}, vol.~38, pp. 1615--1631,
  2005.

\bibitem{Liu2007}
------, ``A mean field annealing approach to accurate free form shape
  matching,'' \emph{Pattern Recognition}, vol.~40, pp. 2418--2436, 2007.

\bibitem{Wells97}
W.~Wells~III, ``Statistical approaches to feature-based object recognition,''
  \emph{International Journal of Computer Vision}, vol.~28, no. 1/2, pp.
  63--98, 1997.

\bibitem{ChuiRangarajan2000}
H.~Chui and A.~Rangarajan, ``A feature registration framework using mixture
  models,'' in \emph{Proc. of IEEE Workshop on Mathematical Methods in
  Biomedical Image Analysis}, 2000, pp. 190--197.

\bibitem{GrangerPennec2002}
S.~Granger and X.~Pennec, ``Multi-scale {EM-ICP}: A fast and robust approach
  for surface registration,'' in \emph{European Conference on Computer Vision},
  vol.~IV, 2002, pp. 418--432.

\bibitem{MyronenkoSongCarreira-Perpinan2006}
A.~Myronenko, X.~Song, and M.~A. Carreira-Perpinan, ``Non-rigid point set
  registration: Coherent point drift,'' in \emph{Proc. of Advances in Neural
  Information Processing Systems}, December 2006, pp. 1009--1016.

\bibitem{SofkaYangStewart2007}
M.~Sofka, G.~Yang, and C.~V. Stewart, ``Simultaneous covariance driven
  correspondence {(CDC)} and transformation estimation in the expectation
  maximization framework,'' in \emph{Proceedings of the IEEE Conference on
  Computer Vision and Pattern Recognition}, June 2007.

\bibitem{JianVemuri2005}
B.~Jian and B.~C. Vemuri, ``A robust algorithm for point set registration using
  mixture of {G}aussians,'' in \emph{IEEE Proc. of the Tenth International
  Conference on Computer Vision}, Beijing, Chian, 2005.

\bibitem{Meer2004}
P.~Meer, ``Robust techniques for computer vision,'' in \emph{Emerging Topics in
  Computer Vision}.\hskip 1em plus 0.5em minus 0.4em\relax Prentice Hall, 2004.

\bibitem{Sinkhorn64}
R.~Sinkhorn, ``A relationship between arbitrary positive matrices and doubly
  stochastic matrices,'' \emph{Annals of Mathematical Statistics}, vol.~35, pp.
  876--879, 1964.

\bibitem{DempsterLairdRubin77}
A.~P. Dempster, N.~M. Laird, and D.~B. Rubin, ``Maximum likelihood estimation
  from incomplete data via the {EM} algorithm (with discussion),''
  \emph{Journal of the Royal Statistical Society, Series B}, vol.~39, pp.
  1--38, 1977.

\bibitem{FraleyRaftery2002}
C.~Fraley and A.~E. Raftery, ``Model-based clustering, discriminant analysis,
  and density estimation,'' \emph{Journal of the American Statistical
  Association}, vol.~97, pp. 611--631, 2002.

\bibitem{ArunHuangBlostein87}
K.~S. Arun, T.~S. Huang, and S.~D. Blostein, ``Least-squares fitting of two
  {3-D} point sets,'' \emph{IEEE Trans. on Pattern Analysis and Machine
  Intelligence}, vol.~9, no.~5, pp. 698--700, September 1987.

\bibitem{Horn87-ortho}
B.~Horn, ``Closed-form solution of absolute orientation using orhtonormal
  matrices,'' \emph{J. Opt. Soc. Amer. A.}, vol.~5, no.~7, pp. 1127--1135,
  1987.

\bibitem{Horn87-quat}
------, ``Closed-form solution of absolute orientation using unit
  quaternions,'' \emph{J. Opt. Soc. Amer. A.}, vol.~4, no.~4, pp. 629--642,
  1987.

\bibitem{Umeyama91}
S.~Umeyama, ``Least-squares estimation of transformation parameters between two
  point patterns,'' \emph{IEEE Transactions on Pattern Analysis and Machine
  Intelligence}, vol.~13, no.~4, pp. 376--380, April 1991.

\bibitem{WilliamsBennamoun2000}
J.~Williams and M.~Bennamoun, ``A multiple view 3{D} registration algorithm
  with statistical error modeling,'' \emph{IEICE Transactions on Information
  and Systems}, vol. E83-D, no.~8, pp. 1662--1670, August 2000.

\bibitem{YuilleStolorzUtans94}
A.~I. Yuille, P.~Stolorz, and J.~Utans, ``Statistical physics, mixture of
  distributions, and the {EM} algorithm,'' \emph{Neural Computation}, vol.~6,
  pp. 334--340, 1994.

\bibitem{LuoHancock2003}
B.~Luo and E.~Hancock, ``A unified framework for alignment and
  correspondence,'' \emph{Computer Vision and Image Understanding}, vol.~92,
  no.~1, pp. 26--55, October 2003.

\bibitem{TsinKanade2004}
Y.~Tsin and T.~Kanade, ``A correlation-based approach to robust point set
  registration,'' in \emph{Proceedings of the Eighth European Conference on
  Computer Vision}, Prague, Czeh Republic, May 2004.

\bibitem{Wang2006}
F.~Wang, B.~Vemuri, A.~Rangarajan, I.~Schmalfuss, and S.~Eisenschenk,
  ``Simultaneous nonrigid registration of multiple point sets and atlas
  construction,'' in \emph{Proceedings of the Ninth European Conference on
  Computer Vision}, Graz, Austria, May 2006.

\bibitem{MengRubin93}
X.-L. Meng and D.~B. Rubin, ``Maximum likelihood estimation via the {ECM}
  algorithm: a general framework,'' \emph{Biometrika}, vol.~80, pp. 267--278,
  1993.

\bibitem{LemarechalOustry2001}
C.~Lemarechal and F.~Oustry, ``{SDP} relaxations in combinatorial optimization
  from a {L}agrangian viewpoint,'' in \emph{Advances in Convex Analysis and
  Global Optimization}, Hadjisavvas and Panos, Eds.\hskip 1em plus 0.5em minus
  0.4em\relax Kluwer Academic Publishers, 2001, ch.~6, pp. 119--134.

\bibitem{KakadiarisMetaxas2000}
I.~Kakadiaris and D.~Metaxas, ``Model-based estimation of 3d human motion,''
  \emph{IEEE Transactions on Pattern Analysis and Machine Intelligence},
  vol.~22, no.~12, pp. 1453--1459, 2000.

\bibitem{PlankersFua2003}
R.~Plaenkers and P.~Fua, ``Articulated soft objects for multi-view shape and
  motion capture,'' \emph{IEEE Transactions on Pattern Analysis and Machine
  Intelligence}, vol.~25, no.~9, pp. 1182--1187, 2003.

\bibitem{BMP04}
C.~Bregler, J.~Malik, and K.~Pullen, ``Twist based acquisition and tracking of
  animal and human kinematics,'' \emph{International Journal of Computer
  Vision}, vol.~56, no.~3, pp. 179--194, February - March 2004.

\bibitem{KRH08}
D.~Knossow, R.~Ronfard, and R.~Horaud, ``Human motion tracking with a kinematic
  parameterization of extremal contours,'' \emph{International Journal of
  Computer Vision}, vol.~79, no.~2, pp. 247--269, September 2008.

\bibitem{HNDB09}
R.~Horaud, M.~Niskanen, G.~Dewaele, and E.~Boyer, ``Human motion tracking by
  registering an articulated surface to 3-{D} points and normals,'' \emph{IEEE
  Transactions on Pattern Analysis and Machine Intelligence}, vol.~31, no.~1,
  pp. 158--164, January 2009.

\bibitem{PellegriniSchindlerNardi2008}
S.~Pellegrini, K.~Schindler, and D.~Nardi, ``A generalization of the {ICP}
  algorithm for articulated bodies,'' in \emph{Proc. of the British Machine
  Vision Conference}, September 2008.

\bibitem{BanfieldRaftery93}
J.~D. Banfield and A.~E. Raftery, ``Model-based {Gaussian and non-Gaussian}
  clustering,'' \emph{Biometrics}, vol.~49, no.~3, pp. 803--821, September
  2002.

\bibitem{Hennig2004}
C.~Hennig, ``Breakdown points for maximum likelihood estimators of
  location-scale mixtures,'' \emph{The Annals of Statistics}, vol.~32, no.~4,
  pp. 1313--1340, 2004.

\bibitem{HennigCoretto2007}
C.~Hennig and P.~Coretto, ``The noise component in model-based cluster
  analysis,'' in \emph{Proceedings of the 31st Annual Conference of the German
  Classification Society on Data Analysis, Machine Learning, and Applications},
  Freiburg, Germany, March 2007.

\bibitem{RednerWalker84}
R.~A. Redner and H.~F. Walker, ``Mixture densities, maximum likelihood and the
  {EM} algorithm,'' \emph{SIAM Review}, vol.~26, no.~2, pp. 195--239, April
  1984.

\bibitem{McLachlanKrishnan97}
G.~J. McLachlan and T.~Krishnan, \emph{The {EM} Algorithm and
  Extensions}.\hskip 1em plus 0.5em minus 0.4em\relax New-York: Wiley, 1997.

\bibitem{Rousseeuw84}
P.~J. Rousseeuw, ``Least median of squares regression,'' \emph{Journal of the
  American Statistical Association}, vol.~79, pp. 871--880, 1984.

\bibitem{RousseeuwVanAelst99}
P.~J. Rousseeuw and S.~Van~Aelst, ``Positive-breakdown robust methods in
  computer vision,'' in \emph{Computing Science and Statistics}, Berk and
  Pourahmadi, Eds.\hskip 1em plus 0.5em minus 0.4em\relax Interface Foundation
  of North America, 1999, vol.~31, pp. 451--460.

\bibitem{Bishop2006}
C.~Bishop, \emph{Pattern Recognition and Machine Learning}.\hskip 1em plus
  0.5em minus 0.4em\relax Springer, 2006.

\bibitem{IngrassiaRocci07}
S.~Ingrassia and R.~Rocci, ``Constrained monotone {EM} algorithms for finite
  mixture of multivariate {G}aussians,'' \emph{Computational Statistics and
  Data Analysis}, vol.~51, pp. 5339--5351, 2007.

\bibitem{Hathaway85}
R.~J. Hathaway, ``A constrained formulation of maximum-likelihood estimation
  for normal mixture distributions,'' \emph{Annals of Statistics}, vol.~13, pp.
  795--?800, 1985.

\bibitem{LemarechalOustry99}
C.~Lemarechal and F.~Oustry, ``Semidefinite relaxations and {L}agrangian
  duality with application to combinatorial optimization,'' INRIA, Tech. Rep.
  3710, June 1999.

\bibitem{McCarthy90}
J.~M. McCarthy, \emph{Introduction to Theoretical Kinematics}.\hskip 1em plus
  0.5em minus 0.4em\relax Cambridge: MIT Press, 1990.

\bibitem{GorceParagiosFleet2008}
M.~de~la Gorce, N.~Paragios, and D.~Fleet., ``Model-based hand tracking with
  texture, shading and self-occlusions,'' in \emph{Proc. of Computer Vision and
  Pattern Recognition}, June 2008.

\bibitem{Hamer2009}
H.~Hamer, K.~Schindler, E.~Koller-Meier, and L.~Van~Gool, ``Tracking a hand
  manipulating an object,'' in \emph{Proc. of International Conference on
  Computer Vision}, October 2009.

\bibitem{DDH04}
G.~Dewaele, F.~Devernay, and R.~Horaud, ``Hand motion from 3{D} point
  trajectories and a smooth surface model,'' in \emph{Proc. of the Eighth
  European Conference on Computer Vision}, ser. LNCS 3021, vol.~I.\hskip 1em
  plus 0.5em minus 0.4em\relax Springer, May 2004, pp. 495--507.

\bibitem{DDHF06}
G.~Dewaele, F.~Devernay, R.~Horaud, and F.~Forbes, ``The alignment between
  3-{D} data and articulated shapes with bending surfaces,'' in \emph{Proc. of
  the Ninth European Conference on Computer Vision}, ser. LNCS 3953, vol.
  III.\hskip 1em plus 0.5em minus 0.4em\relax Springer, May 2006, pp. 578--591.

\bibitem{CeleuxGovaert92}
G.~Celeux and G.~Govaert, ``A classification {EM} algorithm for clustering and
  two stochastic versions,'' \emph{Computational statistics and data analysis},
  vol.~14, no.~3, pp. 315--332, 1992.

\bibitem{ThiessonMeekHeckerman2001}
B.~Thiesson, C.~Meek, and D.~Heckerman, ``Accelerating {EM} for large
  databases,'' \emph{Machine Learning}, vol.~45, no.~3, pp. 279--299, 2001.

\bibitem{VerbeekNunninkVlassis2006}
J.~Verbeek, J.~Nunnink, and N.~Vlassis, ``Accelerated em-based clustering of
  large data sets,'' \emph{Data Mining and Knowledge Discovery}, vol.~13,
  no.~3, pp. 291--307, November 2006.

\end{thebibliography}

\begin{biography}[{\includegraphics[width=1in,
height=1.25in,clip,
keepaspectratio]{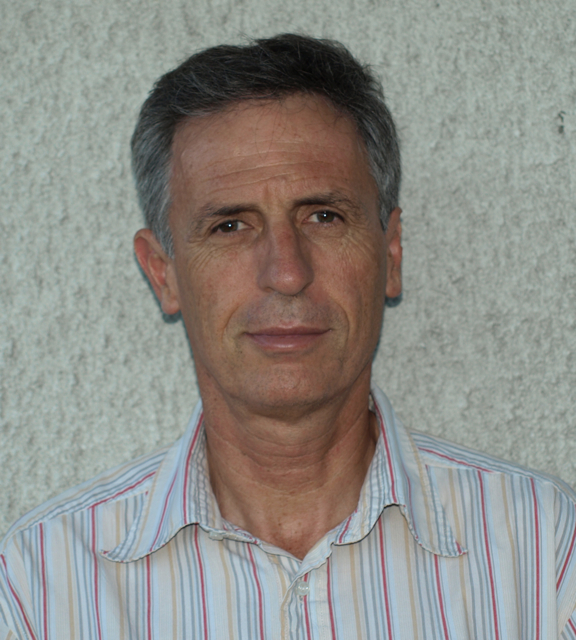}}]{Radu Horaud} 
received the B.Sc. degree in electrical engineering, the M.Sc. degree
in control engineering, and the Ph.D. degree in computer science from
the Institut National Polytechnique de Grenoble, Grenoble, France. 

He holds a position of Director of Research with the Institut National de Recherche en Informatique et
Automatique (INRIA), Grenoble Rh\^one-Alpes, Montbonnot, France, where
he is the head of the PERCEPTION team since 2006. His
research interests include computer vision, machine learning,
multisensory fusion, and robotics. He is an Area Editor of the
\textit{Elsevier Computer Vision and Image Understanding}, a member of
the advisory board of the \textit{Sage International Journal of Robotics
  Research}, and a member of the editorial board of the
\textit{Kluwer International Journal of Computer Vision}. He was a
Program Cochair of the Eighth IEEE International Conference on
Computer Vision (ICCV 2001).
\end{biography}
\vspace*{-1cm}
\begin{biography}[{\includegraphics[width=1in,
height=1.25in,clip,
keepaspectratio]{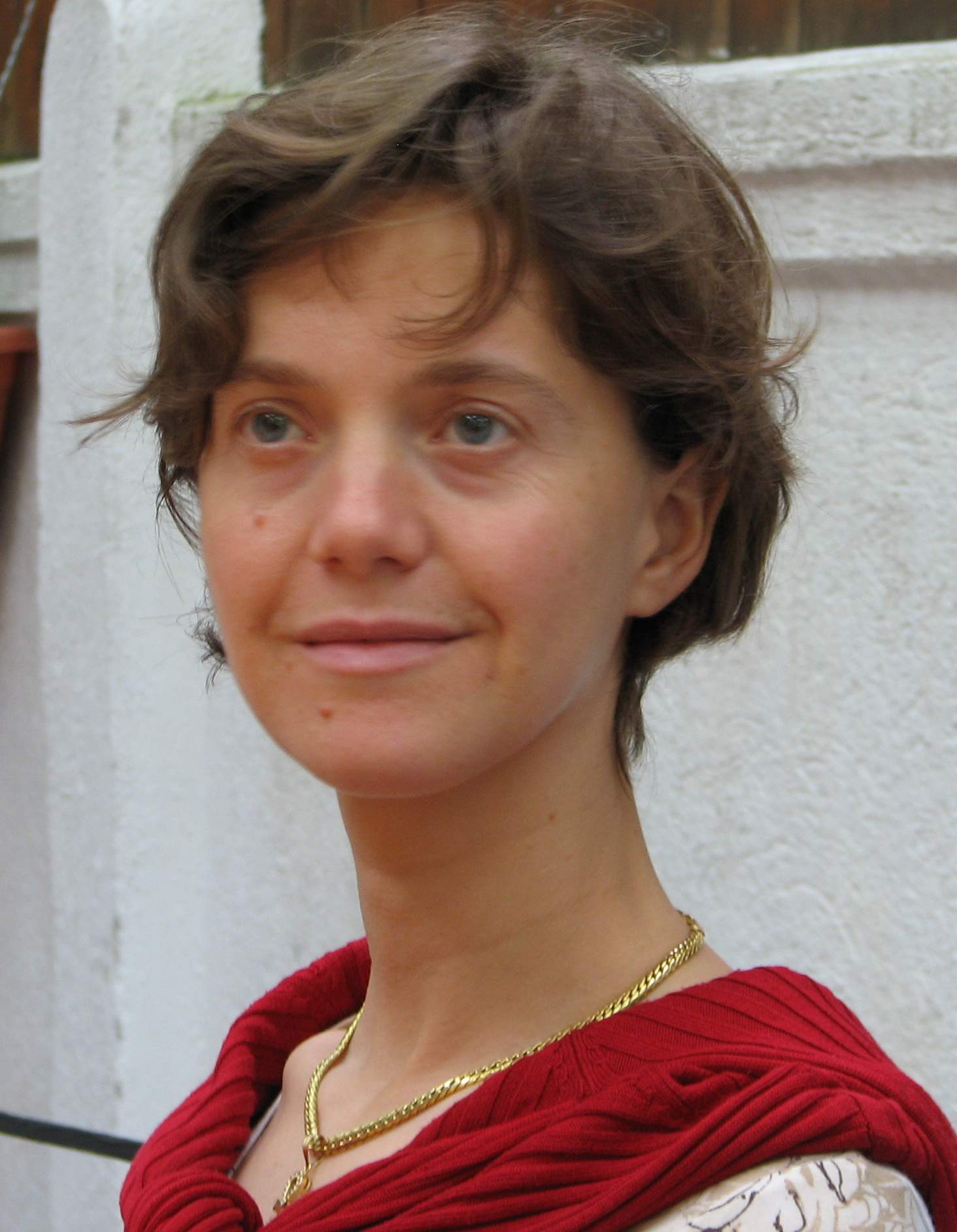}}]{Florence Forbes}
was born in Monaco in 1970.
She received both the B.Sc. and M.Sc. degrees in computer science and applied
mathematics from the Ecole Nationale Sup\'erieure d'Informatique et
Ma\-th\'e\-ma\-ti\-ques Appliqu\'ees de Grenoble, Grenoble, France,
and the Ph.D. degree in applied probabilities from
University Joseph Fourier, Grenoble, France. 

Since 1998 she has been a research
scientist with the Institut National de Recherche en Informatique et
Automatique (INRIA), Grenoble Rh\^one-Alpes, Montbonnot, France, where
she is the head of the MISTIS team since
2003. Her research activities include Bayesian image analysis,
Markov models and hidden structure models.
\end{biography}
\vspace*{-1cm}
\begin{biography}
[{\includegraphics[width=1in,height=1.25in,clip,keepaspectratio]{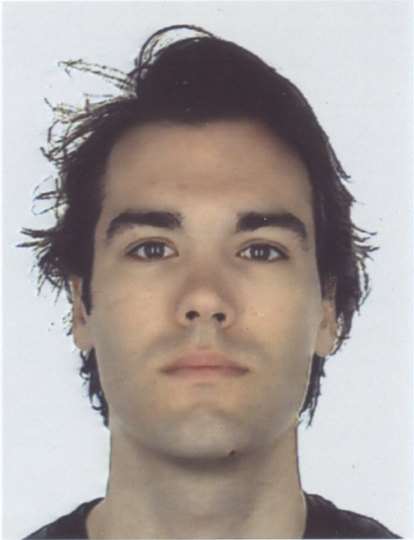}}]
{Manuel Yguel} received both the B.Sc. and M.Sc. degrees in computer science and applied
mathematics from the Ecole Nationale Sup\'erieure d'Informatique et
Ma\-th\'e\-ma\-ti\-ques Appliqu\'ees de Grenoble, Grenoble, France,
and the Ph.D. degree in computer science from the Institut National
Polytechnique de Grenoble, Grenoble, France. 
His research interests include autonomous robot navigation, multisensory fusion, and probabilistic SLAM.
\end{biography}
\vspace*{-1cm}
\begin{biography}
[{\includegraphics[width=1in,height=1.25in,clip,keepaspectratio]{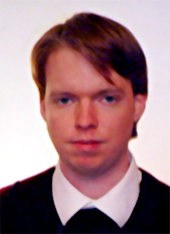}}]
{Guillaume Dewaele} received both the B.Sc. and M.Sc. degrees in Physics from Ecole Normale Sup\'erieure, Lyon, France, and the Ph.D. degree in computer science from the Institut National Polytechnique de Grenoble, Grenoble, France. His research interests include modeling of deformable and articulated objects for computer animation and computer vision.
\end{biography}
\vspace*{-1cm}
\begin{biography}[{\includegraphics[width=1in,
height=1.25in,clip,
keepaspectratio]{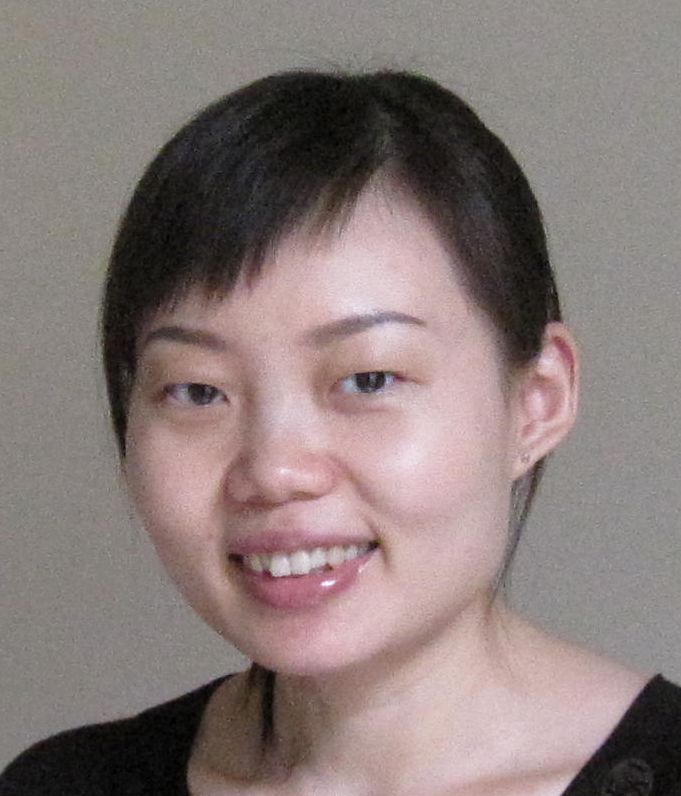}}]{Jian Zhang}
is a Ph.D. student in the Department of Electrical and
Electronic Engineering at the University of Hong Kong, Pokfulam, Hong Kong. In 2009 she spent 4 months at
INRIA Grenoble Rh\^one-Alpes, Montbonnot, France, in the PERCEPTION team. Her research interestes include
feature matching, motion tracking, and 3D reconstruction.

\end{biography}

\end{document}